\documentclass{article}

 \usepackage[main, final]{neurips_2026}

\usepackage[utf8]{inputenc} 
\usepackage[T1]{fontenc}    
\usepackage{hyperref}       
\usepackage{url}            
\usepackage{booktabs}       
\usepackage{amsfonts}       
\usepackage{nicefrac}       
\usepackage{microtype}      
\usepackage{xcolor}         

\usepackage{graphicx}
\usepackage{amsmath, amssymb, amsthm}
\usepackage{algorithm}
\usepackage{algorithmic}
\usepackage{multirow}
\usepackage{subfigure}
\usepackage{caption}
\usepackage{rotating}
\usepackage{subcaption}
\usepackage{float}
\usepackage{cleveref}
\usepackage{dsfont}
\newcommand{\mathbbm}[1]{\mathds{#1}}

\newtheorem{theorem}{Theorem}[section]
\newtheorem{proposition}[theorem]{Proposition}
\newtheorem{lemma}[theorem]{Lemma}

\newtheorem{definition}[theorem]{Definition}
\newtheorem{assumption}[theorem]{Assumption}

\newcommand{\wx}[1]{#1}
\newcommand{\wxcmd}[1]{\textcolor{teal}{Wenkai: #1}}
\newcommand{\lm}[1]{#1}

\newcommand{\POSTNIPS}[1]{#1}

\newcommand{\scorename}{Multivariate Kernel Score }
\newcommand{\scoreacc}{MKS}
\newcommand{\scoreaccsp}{MKS }

\title{A Kernel Nonconformity Score for Multivariate Conformal Prediction}

\author{%
  Louis Meyer \\
  University of Warwick\\
  \texttt{louis.meyer@warwick.ac.uk}
  \And
  Wenkai Xu \\
  University of Warwick\\
  \texttt{wenkai.xu@warwick.ac.uk}
}

\begin{document}

\maketitle

\begin{abstract}
Multivariate conformal prediction requires nonconformity scores that compress residual vectors into scalars while preserving certain implicit geometric structure of the residual distribution. We introduce a 
\scorename (\scoreacc) that produces prediction regions that  explicitly adapt to this geometry. We show that the proposed score 
\wx{resembles} the Gaussian process posterior variance, unifying Bayesian uncertainty quantification with the coverage guarantees \wx{of  frequentist-type}. \wx{Moreover, the \scoreacc} can be decomposed into an anisotropic Maximum Mean Discrepancy (MMD) that interpolates between kernel density estimation and covariance-weighted distance. We prove finite-sample coverage guarantees and establish convergence rates that depend on the effective rank of the 
\wx{kernel-based} covariance operator rather than the ambient dimension, enabling dimension-free adaptation. On regression tasks, the \scoreaccsp reduces the volume of prediction regions significantly, compared 
to ellipsoidal baselines while maintaining nominal coverage, with larger gains at higher dimensions and tighter coverage levels.
\end{abstract}

\section{Introduction}
Conformal prediction constructs prediction regions with finite-sample coverage guarantees under minimal assumptions on the data distribution \citep{shafer2008tutorial, vovk2005algorithmic}. Given a point predictor $\hat{f}$ and a confidence level $1-\alpha$, the framework produces a region $C(X)$ satisfying $\mathbb{P}(Y \in C(X)) \geq 1-\alpha$ for any model class and any distribution \citep{angelopoulos2022gentle}. The univariate case is well-understood, conformalized quantile regression \citep{romano2019conformalized} and standardized residual methods \citep{lei2018distribution} provide simple and effective prediction intervals.  Exact conditional coverage $\mathbb{P}(Y \in C(X) \mid X) \geq 1-\alpha$ cannot be achieved distribution-free without structural assumptions \citep{foygel2021limits, vovk2012conditional}. The efficiency of the score, measured by the size of its prediction region, depends on the choice of nonconformity score and the order it induces.

For multivariate responses $Y \in \mathbb{R}^d$, the nonconformity score determines the \wx{underlying ``shape''} of the prediction region, \wx{where} the efficiency of the region depends on how faithfully this 
shape 
reflects the geometry of the residual distributions. Designing scores that yield \wx{data}-adaptive regions, without requiring/learning a generative model or imposing a fixed parametric shape, remains an open problem. The difficulty is intrinsic to the multivariate setting. In one dimension, residuals \wx{reside in an ordered set $\mathbb{R}$}, and admit a natural total order \wx{where} the 
quantile is 
\wx{well-}defined. In $\mathbb{R}^d$ where no canonical ordering exists, the score function $S: \mathbb{R}^d \to \mathbb{R}$ imposes a total order by compressing a $d$-dimensional residual into a scalar, and this compression 
\wx{unavoidably} discards information. Different choices of $S$ discard different structural properties of the residual distribution, and the efficiency of the prediction region depends 
\wx{predominantly} on \wx{the information} preserved.

Existing approaches split into two families\wx{: ellipsoidal approaches and density-based methods}. Ellipsoidal approaches score residuals by their Mahalanobis distance under the sample covariance, producing regions aligned with the principal axes of the residual distribution \citep{johnstone2021conformal, xu2024conformal}. While \wx{still being} efficient when the residuals are approximately Gaussian, the ellipsoid must expand into low-density areas to maintain coverage when the residual geometry departs from ellipticity, whether from model misspecification, multimodality, or nonlinear dependencies between output coordinates, inflating the region unnecessarily. Density-based methods instead threshold an estimated density, of the residuals or of the response given the covariates, producing level sets that can take arbitrary shape \citep{dheur2025unified, izbicki2022cd}. In practice, nonparametric density estimation deteriorates as the output dimension grows,
and even a perfect density estimate ranks residuals by a single scalar that is invariant to directions: two residuals at equal density but in structurally different orientations relative to the calibration data receive identical scores. Ellipsoidal scores capture directional covariance but are restricted to a fixed geometry, while density-based  scores capture inherent geometry without preserving the directional information. 

We introduce a kernel-based nonconformity score that combines \wx{desirable properties from both}. The score embeds calibration residuals into a reproducing kernel Hilbert space (\wx{RKHS}) and measures the conformity of a test residual through a regularized quadratic form involving the empirical covariance operator \citep{fukumizu2004dimensionality} via \textit{its feature space}. We prove that \wx{the proposed} score \wx{resembles}
the posterior variance of a Gaussian process \wx{(GP)} \citep{williams2006gaussian} 
evaluated at the calibration residual locations. {Importantly, the GP model  does not need to be correctly specified.} The posterior variance serves as a scoring function whose geometric properties we inherit from Bayesian nonparametrics; the coverage guarantee rests on conformal calibration, which requires only exchangeability of the scores. Our approach resembles
Bayesian geometry for the design of the prediction region 
\wx{while achieving frequentist-type} 
validity.

The \scoreaccsp identity yields a decomposition into two interpretable components from the kernel literature. The first is a Maximum Mean Discrepancy \wx{(MMD)} term \citep{gretton2012kernel} that measures the discrepancy between the \wx{empirical} test residual and the calibration distribution, acting as a one-sample typicality test through a kernel density score. The second is a kernel \wx{principal component analysis (KPCA)} \cite{scholkopf1998nonlinear} correction term that accounts for the anisotropy of the residual distribution. The prediction regions that result are neither density level sets nor ellipsoids, but kernel-adapted regions shaped by both the density and the directional covariance of the residuals. When the kernel is linear, the score reduces to the regularized Mahalanobis distance,
so the framework, \wx{in principle,} 
generalizes the ellipsoidal methods 
of Johnstone et al.~\cite{johnstone2021conformal} 
and Xu et al.~\cite{xu2024conformal}. 

Our contributions are 
\wx{threefold}. (1) We propose \scoreacc: a kernel\wx{-based} nonconformity score for multivariate conformal prediction and prove that it equals \wx{a} GP posterior variance (Proposition~\ref{prop:gp_connection}). We show that the score decomposes into an MMD term and a KPCA correction (Proposition~\ref{prop:kmd_mmd}); and that the linear kernel recovers the Mahalanobis distance (Proposition~\ref{prop:linear_kernel}), establishing that our method strictly generalizes ellipsoidal conformal prediction.
(2) We provide finite-sample coverage guarantees under exchangeability, with convergence rates depending on the effective rank of the RKHS covariance operator rather than the ambient dimension $d$ (Theorem~\ref{thm:coverage}); we extend these guarantees to stationary $\alpha$-mixing sequences (Theorem~\ref{thm:convergence_mixing} in Appendix~\ref{sec:mixing}), following the framework of Xu et al.~\cite{xu2024conformal}.
(3) We demonstrate on synthetic and real data that the proposed kernel-based score produces smaller prediction regions than Mahalanobis, density-only and optimal transport baselines while maintaining nominal coverage, with the largest gains at higher coverage levels and higher output dimensions.
The code of our \scoreaccsp is available at \url{https://github.com/meyerlouis/mks-conformal}.



\section{Background}
\paragraph{Split conformal prediction.}

We consider a regression setting with inputs $X \in \mathcal{X}$ and multivariate responses $Y \in \mathbb{R}^d$. Given a dataset $\{(X_i, Y_i)\}_{i=1}^n$ drawn from a distribution $\mathbb{P}_{X,Y}$, the goal is to construct a prediction region $C(X_{\mathrm{new}})$ satisfying
\begin{equation}
\mathbb{P}\bigl(Y_{\mathrm{new}} \in C(X_{\mathrm{new}})\bigr)
\geq 1 - \alpha
\label{eq:marginal_coverage}
\end{equation}
for a user-specified level $\alpha \in (0,1)$.
Following the split conformal framework \citep{lei2018distribution, papadopoulos2002inductive}, the data is randomly partitioned into a training set $\mathcal{D}_{\mathrm{train}}$ of size $n_{\mathrm{train}}$ and a calibration set $\mathcal{D}_{\mathrm{cal}} = \{(X_i, Y_i)\}_{i=1}^T$ of size $T$. A model $\hat{f}: \mathcal{X} \to \mathbb{R}^d$ is fitted on $\mathcal{D}_{\mathrm{train}}$, and the residuals $\varepsilon_i = Y_i - \hat{f}(X_i) \in \mathbb{R}^d$ are computed on the calibration set \wx{$\mathcal{D}_{\mathrm{cal}}$}. A nonconformity score function $S: \mathbb{R}^d \to \mathbb{R}$ maps each residual to a scalar measuring its degree of nonconformity, with larger values indicating less typical residuals. The conformal threshold $\hat{q}$ is defined as the $\lceil(1-\alpha)(T+1)\rceil / T$ quantile of the calibration scores $\{S(\varepsilon_1), \ldots, S(\varepsilon_T)\}$, yielding the prediction region
\begin{equation}
C(X) = \bigl\{y \in \mathbb{R}^d :
S\bigl(y - \hat{f}(X)\bigr) \leq \hat{q}\bigr\}
\label{eq:prediction_region_bg}
\end{equation}
If the calibration scores and the test score are exchangeable, then $C(X)$ satisfies~\eqref{eq:marginal_coverage} \citep{vovk2005algorithmic}. This guarantee is distribution-free, i.e. the result holds for any data distribution, any model $\hat{f}$, and any score function $S$. The choice of $S$ does not affect coverage but determines the shape and efficiency of the prediction region, \wx{e.g. the length of the prediction interval in one dimension}.

\paragraph{Exchangeability.}

The coverage guarantee~\eqref{eq:marginal_coverage} requires the calibration residuals $\varepsilon_1, \ldots, \varepsilon_T$ and the test residual $\varepsilon_{T+1}$ to be exchangeable. Under the split conformal setup with i.i.d.\ data, this holds by construction. When the data exhibits temporal or distributional dependence, exchangeability generally fails. Barber et al.~\cite{barber2023conformal} provide a general framework for conformal prediction beyond exchangeability, bounding the coverage gap in terms of the total variation distance between the true joint distribution and the nearest exchangeable one. Xu et al.~\cite{xu2023conformal} derive finite-sample coverage guarantees under $\alpha$-mixing dependence,
showing that coverage deviations vanish as the mixing coefficients decay. Other forms of non-exchangeability, including covariate shift \cite{tibshirani2019conformal}, missing data \cite{zaffran2023conformal}, and contamination \cite{clarkson2024split}, have also been studied.


\paragraph{Multivariate nonconformity scores.}

The choice of $S$ determines the geometry of the prediction region~\eqref{eq:prediction_region_bg}, since the region is a sublevel set of $S$. 
An intuitive approach applies univariate conformal prediction independently to each coordinate at level $\alpha/d$, producing hyper-rectangular regions that ignore all the dependencies between outputs. Copula-based methods \cite{messoudi2021copula, sun2024copula} calibrate the marginals jointly but still yield rectangular regions.

The Mahalanobis distance $S(\varepsilon) = (\varepsilon - \bar{\varepsilon})^\top (\widehat{\Sigma} + \lambda I)^{-1} (\varepsilon - \bar{\varepsilon})$ produces ellipsoidal regions aligned with the principal axes of the residual distribution \citep{johnstone2021conformal}. Xu et al.~\cite{xu2024conformal} use this score with a global covariance estimate for multivariate time series, providing coverage guarantees under $\alpha$-mixing
dependence. 
Braun et al.~\cite{braun2025multivariate} learn an input-dependent covariance $\Sigma(X)$, producing locally adaptive ellipsoids with improved conditional coverages. Messoudi et al.~\cite{messoudi2022ellipsoidal} estimate the local covariance 
structures using methods of nearest neighbors.

Density-based approaches use an estimated density as the nonconformity score, producing regions that are its level sets \citep{izbicki2022cd, sampson2025flexibleconformalhighestpredictive}. Methods such as HPD-split \citep{izbicki2022cd} estimate the conditional density $\hat{p}(y \mid x)$, so the shape of the region varies with the covariates and targets conditional coverage when the residual distribution depends on $X$. In practice, nonparametric density estimation becomes unreliable as the output dimension grows, and scalable alternatives require fitting conditional density estimators or generative models
\citep{plassier2024probabilistic, dheur2025unified}. Optimal transport approaches \citep{thurin2025optimal, klein2025multivariate} and adaptive norm-based formulations \citep{braun2025minimum} offer more flexible region shapes at the cost of solving expensive optimization problems or training dedicated networks. Dheur et al.~\cite{dheur2025unified} provide a comprehensive comparison of multivariate conformal methods within a unified framework. 

{These methods also differ in whether the prediction region adapts to the covariates. Locally adaptive methods \citep{braun2025multivariate, izbicki2022cd, messoudi2022ellipsoidal} learn a covariate-dependent structure, such as a local covariance or a conditional density, so that the shape of the region changes with $X$. This targets conditional coverage when the residual distribution depends on the input, at the cost of learning an additional function of $X$. Our score belongs to the class of methods that produce a single region translated by the point prediction, like the Mahalanobis ellipsoid \citep{johnstone2021conformal} and the optimal transport score \citep{klein2025multivariate}. Within this class, it adapts to the global geometry of the residual distribution, including multimodality, curvature and anisotropy, without learning any covariate-dependent structure.}


\section{\scorename}
\label{sec:kmd:def}

We 
\wx{introduce}
a kernel-based nonconformity score that extends the 
Mahalanobis distance  to nonlinear settings. Let $\{\varepsilon_i\}_{i=1}^T \subset \mathbb{R}^d$ be calibration residuals obtained from a split conformal procedure, with empirical mean $\bar{\varepsilon} = \frac{1}{T}\sum_{i=1}^T \varepsilon_i$ and sample covariance $\widehat{\Sigma} = \frac{1}{T-1}\sum_{i=1}^T (\varepsilon_i - \bar{\varepsilon})(\varepsilon_i - \bar{\varepsilon})^\top$.

\begin{definition}[Regularized Mahalanobis Distance]
\label{def:mahalanobis}
The regularized Mahalanobis distance nonconformity score is
\begin{equation}
\mathrm{MD}(\varepsilon) = (\varepsilon - \bar{\varepsilon})^\top \big(\widehat{\Sigma} + \lambda I_d\big)^{-1} (\varepsilon - \bar{\varepsilon})
\label{eq:mahalanobis}
\end{equation}
where $\lambda > 0$ is a regularization parameter.
\end{definition}

Let $k: \mathbb{R}^d \times \mathbb{R}^d \to \mathbb{R}$ be a positive definite kernel with feature map $\phi: \mathbb{R}^d \to \mathcal{H}$, where $\mathcal{H}$ is a reproducing kernel Hilbert space satisfying $k(\mathbf{x}, \mathbf{x}') = \langle \phi(\mathbf{x}), \phi(\mathbf{x}') \rangle_{\mathcal{H}}$.

\begin{definition}[\scorename (\scoreacc)]
\label{def:kmd}
The \scorename is the regularized Mahalanobis distance computed in the feature space $\mathcal{H}$:
\begin{equation}
\hat{e}_k(\varepsilon) = \langle \tilde{\phi}(\varepsilon), \; \big(\widehat{C}_\phi + \lambda I_{\mathcal{H}}\big)^{-1} \tilde{\phi}(\varepsilon) \rangle_{\mathcal{H}}
\label{eq:kmd_rkhs}
\end{equation}
where $\tilde{\phi}(\varepsilon) = \phi(\varepsilon) - \hat{\mu}_\phi$ is the centered feature map, $\hat{\mu}_\phi = \frac{1}{T}\sum_{i=1}^T \phi(\varepsilon_i)$ is the kernel mean embedding, and $\widehat{C}_\phi = \frac{1}{T-1}\sum_{i=1}^T \tilde{\phi}(\varepsilon_i) \otimes \tilde{\phi}(\varepsilon_i)$ is the sample covariance operator on $\mathcal{H}$ \cite{fukumizu2004dimensionality}.
\end{definition}

The Tikhonov regularization ensures the score is well-defined. In an RKHS with infinite-dimensional feature space,
the covariance operator has eigenvalues decaying to $0$ and $C_\phi^{-1}$ is unbounded, but $(C_\phi + \lambda I_\mathcal{H})^{-1}$ has operator norm at most $1/\lambda$. In the Gaussian Process interpretation developed in Section~\ref{sec:kmd:interp}, $\lambda$ corresponds to the \wx{variance of the} observation noise,
controlling the balance between interpolating the calibration data and smoothing over it. Using the kernel trick and the Woodbury matrix identity (proof in Appendix~\ref{proof:kernel_md}), the score can be computed entirely in terms of kernel evaluations.
The procedure is summarized in Algorithm~\ref{alg:kernel_cp}. 

\begin{algorithm}[t]
\caption{Kernel Conformal Prediction}
\label{alg:kernel_cp}
\begin{algorithmic}[1]
\REQUIRE Calibration residuals $\{\varepsilon_i\}_{i=1}^T$, kernel $k$, regularization parameter $\gamma$, level $\alpha$
\STATE Split $\{1,\ldots,T\}$ into disjoint folds $\mathcal{I}_{\mathrm{fit}}$ and $\mathcal{I}_{\mathrm{cal}}$ of equal size, with $T_{\mathrm{fit}} = |\mathcal{I}_{\mathrm{fit}}|$ and $T_{\mathrm{cal}} = |\mathcal{I}_{\mathrm{cal}}|$
\STATE Precompute $(\tilde{K} + \gamma I_{T_{\mathrm{fit}}})^{-1}$ with centered Gram matrix $\tilde{K}$ built on the fit fold,
$$\tilde{K}_{ij} = k(\varepsilon_i, \varepsilon_j) - \tfrac{1}{T_{\mathrm{fit}}}\sum_{l} k(\varepsilon_i, \varepsilon_l) - \tfrac{1}{T_{\mathrm{fit}}}\sum_{l} k(\varepsilon_l, \varepsilon_j) + \tfrac{1}{T_{\mathrm{fit}}^2}\sum_{l,m} k(\varepsilon_l, \varepsilon_m), \quad i,j,l,m \in \mathcal{I}_{\mathrm{fit}}$$
\FOR{$i \in \mathcal{I}_{\mathrm{cal}}$}
    \STATE $\hat{e}_k(\varepsilon_i) \leftarrow \tilde{k}(\varepsilon_i, \varepsilon_i) - \tilde{\mathbf{k}}_*(\varepsilon_i)^\top (\tilde{K} + \gamma I_{T_{\mathrm{fit}}})^{-1} \tilde{\mathbf{k}}_*(\varepsilon_i) \qquad$ (as in \textit{Eq.~\eqref{eq:kmd_computable}})
\ENDFOR
\STATE $\hat{q} \leftarrow \lceil(1-\alpha)(T_{\mathrm{cal}}+1)\rceil/T_{\mathrm{cal}}$ quantile of $\{\hat{e}_k(\varepsilon_i)\}_{i \in \mathcal{I}_{\mathrm{cal}}}$
\ENSURE Prediction region $C(X) = \{y : \hat{e}_k(y - \hat{f}(X)) \leq \hat{q}\}$
\end{algorithmic}
\end{algorithm}

{The score requires a single factorization of $\tilde K + \gamma I$ on the fit fold, performed once at $O(T_{\mathrm{fit}}^3)$ time and $O(T_{\mathrm{fit}}^2)$ memory. Each subsequent evaluation then costs $O(T_{\mathrm{fit}}^2)$, and the output dimension only enters through the $O(d)$ cost of each kernel evaluation. For larger calibration sets, the Nystr\"om method approximates $\tilde K$ from a subset of the fit fold. Sampling points according to their ridge leverage scores allows the required sample size to scale with the effective dimension of \(\tilde K\), rather than \(T_{\mathrm{fit}}\) \citep{alaoui2015fast}. By Proposition~\ref{prop:leverage}, the ridge leverage score of a fit fold point is its score divided by $\gamma$, so the points that matter most for the approximation are those the score ranks as most atypical.}

\medskip
\begin{proposition}[Computable Form]
\label{prop:kmd_computable}
Let $\tilde{K} \in \mathbb{R}^{T \times T}$ be the centered Gram matrix with entries $\tilde{K}_{ij} = \langle \tilde{\phi}(\varepsilon_i), \tilde{\phi}(\varepsilon_j) \rangle_{\mathcal{H}}$, and let $\tilde{\mathbf{k}}_*(\varepsilon) \in \mathbb{R}^T$ be the centered kernel vector with $[\tilde{\mathbf{k}}_*(\varepsilon)]_i = \langle \tilde{\phi}(\varepsilon), \tilde{\phi}(\varepsilon_i) \rangle_{\mathcal{H}}$. Setting $\gamma = \lambda(T-1)$, the \scorename is
\begin{equation}
\hat{e}_k(\varepsilon)
= \tilde{k}(\varepsilon, \varepsilon)
- \tilde{\mathbf{k}}_*(\varepsilon)^\top
(\tilde{K} + \gamma I_T)^{-1}
\tilde{\mathbf{k}}_*(\varepsilon)
\label{eq:kmd_computable}
\end{equation}
where $\tilde{k}(\varepsilon,\varepsilon) = \|\tilde{\phi}(\varepsilon)\|_{\mathcal{H}}^2$.
\end{proposition}

Given a target miscoverage level $\alpha \in (0,1)$, we define the conformal threshold $\hat{q}$ as the $\lceil (1-\alpha)(T_{\mathrm{cal}}+1) \rceil / T_{\mathrm{cal}}$ quantile of the scores on the quantile fold, $\{\hat{e}_k(\varepsilon_i)\}_{i \in \mathcal{I}_{\mathrm{cal}}}$. The prediction region for a new input $X_{T+1}$ with point prediction $\hat{f}(X_{T+1})$ is then
\begin{equation}
C(X_{T+1}) = \bigl\{ y \in \mathbb{R}^d : \hat{e}_k\bigl(y - \hat{f}(X_{T+1})\bigr) \leq \hat{q} \bigr\}
\label{eq:prediction_region}
\end{equation}

\medskip

\begin{theorem}[Coverage Guarantee]
\label{thm:coverage}
Let the calibration residuals be divided into disjoint folds $\mathcal{I}_{\mathrm{fit}}$ and $\mathcal{I}_{\mathrm{cal}}$, with the score constructed from $\mathcal{I}_{\mathrm{fit}}$ and the threshold taken as the $\lceil(1-\alpha)(|\mathcal{I}_{\mathrm{cal}}|+1)\rceil$-th smallest score on $\mathcal{I}_{\mathrm{cal}}$. If the residuals $\{\varepsilon_i\}_{i \in \mathcal{I}_{\mathrm{cal}}} \cup \{\varepsilon_{T+1}\}$ are exchangeable conditionally on $\mathcal{I}_{\mathrm{fit}}$, then the prediction region~\eqref{eq:prediction_region} satisfies
\begin{equation}
\mathbb{P}\bigl(Y_{T+1} \in C(X_{T+1})\bigr) \geq 1 - \alpha
\label{eq:coverage}
\end{equation}
exactly, with no distributional assumptions on the residual law. Conditionally on $\mathcal{I}_{\mathrm{fit}}$ the score is a fixed measurable function, so exchangeability of the residuals transfers to the scores and \eqref{eq:coverage} follows from the standard quantile argument.

Under Assumptions~\ref{assumption_iid}--\ref{assumption:trace-class}, which include the standard exogeneity condition $\varepsilon_{T+1} \perp X_{T+1}$, the conditional coverage deviation satisfies
\begin{equation}
\bigl|\mathbb{P}\bigl(Y_{T+1} \in C(X_{T+1}) \mid X_{T+1}\bigr) - (1-\alpha)\bigr| \leq C \cdot T^{-r}
\label{eq:conditional_coverage}
\end{equation}
with high probability, where $r > 0$ depends on the estimation rate $\rho$ and the regularization, and $C$ depends on the effective rank $\mathbf{r}(\Sigma) = \mathrm{tr}(\Sigma)/\|\Sigma\|_{\mathrm{op}}$ of the RKHS covariance operator. Under $\alpha$-mixing we establish marginal coverage guarantees in Appendix~\ref{sec:mixing}.
\end{theorem}

\vspace{-0.2cm}


{\paragraph{Remark.}
For smooth kernels such as the RBF, the eigenvalues of $\Sigma$ decay rapidly and $\mathbf{r}(\Sigma)$ can be much smaller than both $d$ and $T$. For any stationary kernel with $k(0) = 1$, the part of the constant $C$ that depends on the kernel does not depend on $d$ (see the remark following Theorem~\ref{thm:convergence_iid}). The detailed convergence analysis of Theorem~\ref{thm:coverage} is in Appendix~\ref{sec:theory_proofs}.}

\section{Probabilistic and Geometric Structure}
\label{sec:kmd:interp}



\subsection{Gaussian Process Connection}

\POSTNIPS{The \scorename has a direct probabilistic interpretation as the posterior variance of a Gaussian process observed at the calibration residuals.}

\begin{proposition}[Gaussian Process Identity]
\label{prop:gp_connection}
Let $g \sim \mathcal{GP}(0, \tilde{k})$ be a Gaussian process with zero mean and centered kernel $\tilde{k}$. Given observations at locations $\{\varepsilon_i\}_{i=1}^T$ with noise variance $\sigma_n^2 = \gamma$, the posterior variance at a test point $\varepsilon$ is~{\citep[Eq.~2.26]{williams2006gaussian}}:
\begin{equation}
\mathrm{Var}_{\mathrm{GP}}\bigl[g(\varepsilon) \mid \{\varepsilon_i\}_{i=1}^T\bigr] = \tilde{k}(\varepsilon, \varepsilon) - \tilde{\mathbf{k}}_*(\varepsilon)^\top (\tilde{K} + \gamma I_T)^{-1} \tilde{\mathbf{k}}_*(\varepsilon)
\end{equation}
Comparing with Proposition~\ref{prop:kmd_computable}, we have the exact identity
\begin{equation}
\hat{e}_k(\varepsilon) = \mathrm{Var}_{\mathrm{GP}}\bigl[g(\varepsilon) \mid \{\varepsilon_i\}_{i=1}^T\bigr]
\label{eq:gp_identity}
\end{equation}
\end{proposition}

{The process $g$ is only introduced for interpretation. It is unrelated to the regression function $f$ and is never observed or estimated.} In standard GP regression, the posterior variance quantifies uncertainty about a function at unobserved locations, and depends on the observed function values $\mathbf{z}$ through the posterior mean. 
Our score depends only on the locations $\{\varepsilon_i\}_{i=1}^T$ of the calibration residuals, not on any response observed at those locations. The \scoreaccsp is a purely geometric quantity that measures how well a test residual is explained by the spatial configuration of the calibration data. A second distinction is that we use the centered kernel $\tilde{k}$ rather than the raw kernel $k$, introducing density sensitivity. For the RBF kernel, $k(\varepsilon, \varepsilon) = 1$ for any $\varepsilon$, while after centering, $\tilde{k}(\varepsilon, \varepsilon) = 1 + \bar{k} - 2\hat{p}_k(\varepsilon)$ varies with the local density of the calibration distribution. Centering enables the score to distinguish residuals in dense regions from those in the sparse ones.

The regularization parameter $\gamma$ enters the GP interpretation as the observation noise variance, and governs how much of the calibration data's structure the score retains. Small $\gamma$ corresponds to near-interpolation, where the GP posterior variance drops to nearly zero at each calibration point, and the score surface closely tracks the placement of individual residuals. Large $\gamma$ smooths over individual observations, making the posterior variance reflect only the broad distributional structure, and the score approaches the MMD, as shown in Proposition~\ref{prop:kmd_mmd}. 

This interpretation connects Bayesian and frequentist reasoning within a single framework. The GP posterior variance provides geometrically rich machinery for designing the nonconformity score. It captures density, directional covariance, and manifold structure, producing prediction regions that adapt to the shape of the residual distribution. The conformal calibration provides validity, the coverage guarantee~\eqref{eq:coverage} holds regardless of whether the GP model is correctly specified, because it relies on exchangeability of the scores, not on Gaussianity of the process. 

\subsection{MMD-KPCA Decomposition}

\POSTNIPS{The \scorename can also be written in terms of the kernel principal components of the calibration residuals, which separates its density and directional parts.}

{
\begin{proposition}[MMD--KPCA Decomposition]
\label{prop:kmd_mmd}
Let $Q_T = \frac{1}{T}\sum_{i=1}^T \delta_{\varepsilon_i}$ denote the empirical distribution of the calibration residuals, where $\delta_{\varepsilon_i}$ is the Dirac measure at $\varepsilon_i$ (the probability distribution that assigns all its mass to the point $\varepsilon_i$). The \scorename satisfies
\begin{equation}
\hat{e}_k(\varepsilon) = \mathrm{MMD}^2(\delta_\varepsilon,\, Q_T) \;-\; \sum_{j:\,\lambda_j > 0} \frac{\lambda_j}{\lambda_j + \gamma}\, \beta_j(\varepsilon)^2,
\label{eq:kmd_mmd}
\end{equation}
where $\mathrm{MMD}^2(\delta_\varepsilon, Q_T) = \|\tilde{\phi}(\varepsilon)\|_\mathcal{H}^2 = k(\varepsilon, \varepsilon) + \bar{k} - 2\hat{p}_k(\varepsilon)$ is the distance 
between the Dirac measure at $\varepsilon$ and the (empirical) calibration distribution measure in the form of Maximum Mean Discrepancy, with $\hat{p}_k(\varepsilon) = \frac{1}{T}\sum_{i=1}^T k(\varepsilon, \varepsilon_i)$ and $\bar{k} = \frac{1}{T^2}\sum_{i,j} k(\varepsilon_i, \varepsilon_j)$. The quantities $\{\lambda_j\}_{j=1}^T$ are the eigenvalues of $\tilde{K}$, with corresponding eigenvectors $\{\mathbf{v}_j\}_{j=1}^T$, and for each of the $\lambda_j > 0$, $\beta_j(\varepsilon) = \lambda_j^{-1/2}\, \mathbf{v}_j^\top \tilde{\mathbf{k}}_*(\varepsilon)$ is the projection of $\tilde{\phi}(\varepsilon)$ onto the $j$-th normalized kernel principal component of the calibration data.
\end{proposition}}

The proof is in Appendix~\ref{proof:kmd_mmd}. The first term builds on MMD
\cite{gretton2012kernel}, an integral probability metric that measures the distance between two distributions $P$ and $Q$ through their kernel mean embeddings in RKHS, defined as $\mathrm{MMD}^2(P, Q) = \|\mu_P - \mu_Q\|_{\mathcal{H}}^2$. When the kernel is characteristic, $\mathrm{MMD}(P, Q) = 0$ if and only if $P = Q$ \cite{sriperumbudur2010hilbert}. The MMD tests for distributional equality and is widely used as a two-sample test statistic. In our setting, we apply the same principle with $P = \delta_\varepsilon$, a single test residual, and $Q = Q_T$, the empirical calibration distribution, so that the term measures the discrepancy between one point and the calibration set, effectively acting as a one-sample typicality test. For the RBF kernel, $k(\varepsilon, \varepsilon) = 1$ for all $\varepsilon$ and $\bar{k}$ depends only on the calibration set, so the only term that varies with the test point is $\hat{p}_k(\varepsilon) = \frac{1}{T}\sum_i k(\varepsilon, \varepsilon_i)$, a Parzen-Rosenblatt kernel density estimate up to normalization \cite{parzen1962estimation, RosenblattNonParametric}. The MMD term is monotonically decreasing in estimated density, implying that the residuals in high-density regions receive low scores and those in sparse regions receive high scores. Using this term alone as a nonconformity score would produce prediction regions that are level sets of the kernel density estimate.

The KPCA correction introduces anisotropy. It reduces the score for residuals whose feature-space representation aligns with directions of high calibration variance, where spread is expected and deviations are uninformative. The score is large only where the calibration data provides neither density support nor directional explanation, and small where either is present. The prediction region is therefore neither a density level set nor an ellipsoid, but shaped by both the local density and the directional covariance of the calibration residuals.

This decomposition mirrors a familiar progression in $\mathbb{R}^d$. The squared error $\|\varepsilon - \bar{\varepsilon}\|^2$ treats all directions equally, and the Mahalanobis distance adds inverse-covariance weighting to penalize deviations along low-variance directions. The MMD term is the RKHS analogue of the squared error, an isotropic distance from the mean embedding that ignores directional structure. The KPCA correction adds inverse-covariance weighting in $\mathcal{H}$, completing the lift from Mahalanobis in $\mathbb{R}^d$ to an anisotropic distance in the feature space. The parameter $\gamma$ controls how much of the directional structure the score uses. Large $\gamma$ suppresses the correction and recovers isotropic density-only scoring, while small $\gamma$ allows the anisotropic structure to shape the prediction region.


\subsection{Special Cases and Properties}
\label{sec:kmd:special}

\POSTNIPS{The two decompositions above yield several properties of the \scoreacc. The score is non-negative, recovers the Mahalanobis distance under a linear kernel, and reduces to ridge leverage scores on the fit fold residuals.}


\begin{proposition}[Non-negativity]
\label{prop:nonneg}
The \scoreaccsp  satisfies $\hat{e}_k(\varepsilon) \geq 0$ for all $\varepsilon \in \mathbb{R}^d$, with equality if and only if $\tilde{\phi}(\varepsilon) = 0$,  i.e. $\phi(\varepsilon) = \hat{\mu}_\phi$.
\end{proposition}

\begin{proof}
Immediate from the GP identity (Proposition~\ref{prop:gp_connection}): the score equals the posterior variance of a Gaussian process, which is non-negative by construction.
\end{proof}

\begin{proposition}[Linear Kernel Recovery]
\label{prop:linear_kernel}
For the linear kernel $k(\mathbf{x}, \mathbf{x}') = \mathbf{x}^\top \mathbf{x}'$, the \scorename equals $\lambda \cdot \mathrm{MD}(\varepsilon)$, where $\mathrm{MD}(\varepsilon)$ is the regularized Mahalanobis distance of Definition~\ref{def:mahalanobis}. Since the constant factor $\lambda$ does not affect ranking, the two scores produce identical conformal prediction regions.
\end{proposition}

The proof is in Appendix~\ref{proof:linear_kernel}. \scoreaccsp strictly generalizes the Mahalanobis approach~\citep{johnstone2021conformal, xu2024conformal}, as the linear kernel recovers their score exactly, while nonlinear kernels extend the method beyond ellipsoidal regions. In the linear case, the regularization parameter $\lambda$ corresponds to ridge regularization of the sample covariance, and the calibration scores reduce to the diagonal of 
matrix 
$X(X^\top X + \lambda I)^{-1}X^\top$.

Moreover, the \scorename admits a decomposition into a component within the span of the calibration data in $\mathcal{H}$ and a component orthogonal to it.


{
\begin{proposition}[Spectral Decomposition]
\label{prop:spectral}
Let $\{\lambda_j, \mathbf{v}_j\}_{j=1}^T$ be the eigenvalues and eigenvectors of $\tilde{K}$, and for each $\lambda_j > 0$ let $\mathbf{e}_j = \mathbf{w}_j / \|\mathbf{w}_j\|_{\mathcal{H}} = \mathbf{w}_j / \sqrt{\lambda_j}$ be the normalized kernel principal components in $\mathcal{H}$, where $\mathbf{w}_j = \sum_{i=1}^T [\mathbf{v}_j]_i\,\tilde{\phi}(\varepsilon_i)$, so that $\beta_j(\varepsilon) = \langle \tilde{\phi}(\varepsilon), \mathbf{e}_j \rangle_{\mathcal{H}}$ with $\beta_j$ as in Proposition~\ref{prop:kmd_mmd}. Define the orthogonal residual $\tilde{\phi}_\perp(\varepsilon) = \tilde{\phi}(\varepsilon) - \sum_{j:\,\lambda_j > 0} \beta_j(\varepsilon)\,\mathbf{e}_j$. Then
\begin{equation}
\hat{e}_k(\varepsilon) = \sum_{j:\,\lambda_j > 0} \frac{\gamma}{\lambda_j + \gamma}\, \beta_j(\varepsilon)^2 + \|\tilde{\phi}_\perp(\varepsilon)\|_{\mathcal{H}}^2
\label{eq:spectral}
\end{equation}
\end{proposition}}

The proof is in Appendix~\ref{proof:spectral}. 
{The within-span component $\sum_j \gamma\,\beta_j(\varepsilon)^2 / (\lambda_j + \gamma)$ measures atypicality along the principal directions of the calibration data. Each squared projection is weighted by $\gamma/(\lambda_j + \gamma)$, close to $\gamma/\lambda_j$ along directions of large variance}, consistent with the Mahalanobis interpretation.
The orthogonal component $\|\tilde{\phi}_\perp(\varepsilon)\|_{\mathcal{H}}^2$ captures the part of the test residual that lies entirely outside the span of the calibration data in $\mathcal{H}$. It dominates residuals in regions of the output space not explored by the calibration set, where the data provides no structural information and the GP posterior variance reduces to the prior variance.

The parameter $\gamma$ controls the balance between these two components. When $\gamma \to 0$, the within-span weights vanish and the score trusts the calibration structure completely and penalizes only the orthogonal component. When $\gamma \to \infty$, the weights approach $1$, the KPCA correction in~\eqref{eq:kmd_mmd} vanishes, and the score reduces to $\mathrm{MMD}^2(\delta_\varepsilon, Q_T)$, recovering isotropic density-only scoring.
For fit fold residuals, $\tilde{\phi}(\varepsilon_i)$ lies in the span of the fit fold data by construction, so the orthogonal component vanishes and the score admits a closed-form expression in terms of the kernel ridge regression hat matrix.

\begin{proposition}[Leverage Score Identity]
\label{prop:leverage}
Let $H = \tilde{K}(\tilde{K} + \gamma I_T)^{-1}$ be the kernel ridge regression hat matrix, with diagonal entries $h_{ii} = H_{ii}$. For each fit fold residual $\varepsilon_i$, the \scorename satisfies
\begin{equation}
\hat{e}_k(\varepsilon_i) = \gamma \cdot h_{ii}
\label{eq:leverage}
\end{equation}
\end{proposition}

The proof is in Appendix~\ref{proof:leverage}. The diagonal entries $h_{ii}$ are the $\gamma$-ridge leverage scores of the centered Gram matrix~\citep{alaoui2015fast}, which measure the influence of each calibration point on the kernel ridge regression fit. The conformal threshold is a quantile $\hat{q}$ directly determined by the ordering of the leverages $\{h_{ii}\}_{i=1}^T$, where high-leverage calibration residuals occupying unusual positions in feature space receive high nonconformity scores, and low-leverage residuals in well-populated regions receive low scores. For a test point $\varepsilon \notin \{\varepsilon_i\}$, the orthogonal component $\|\tilde{\phi}_\perp(\varepsilon)\|_{\mathcal{H}}^2$ is generally nonzero, and the \scoreaccsp can be understood as an out-of-sample extension of leverage, penalizing test residuals that lie outside the span of the calibration data in $\mathcal{H}$.

{
\paragraph{High-dimensional geometry.} Once its shape is fixed by the covariance estimate, the Mahalanobis ellipsoid can only grow through the conformal threshold $\hat q$, which scales it in all directions at once. At tight coverage levels, $\hat q$ is set by the most extreme calibration residuals, and when the residual distribution is not elliptical these extremes occur in particular directions. Covering them requires enlarging the ellipsoid in every direction, and since its volume grows as $\hat q^{\,d/2}$ (Appendix~\ref{sec:volume_estimation}), this cost grows exponentially with the dimension. The MMD term of the MKS assigns high scores wherever calibration support is absent, so the region grows only where the data does. In the case of higher dimensions, each additional output brings new dependencies that move the residual distribution further from any ellipsoid. The MMD term alone, however, is limited by concentration of norm~\cite[Theorem~6.3.2]{vershynin2018high}. When the residual variance is spread over many directions, pairwise distances between residuals concentrate around a common value. With a lengthscale on the scale of these distances, as the median heuristic gives, the density estimate depends mostly on the distance to the center of the residuals and treats all directions alike. The KPCA correction restores the anisotropy that a covariance captures. Along kernel principal directions of large variance, the correction of Proposition~\ref{prop:kmd_mmd} removes almost all of the squared projection, so the score behaves like a Mahalanobis distance in the feature space along these directions, while the MMD term still follows the support of the residuals elsewhere.}

\section{Experiments}

In all experiments, we use the RBF kernel $k(\varepsilon, \varepsilon') = \exp(-\|\varepsilon - \varepsilon'\|^2 / 2\ell^2)$ for demonstration purposes. The primary contribution of this work is conceptual and theoretical; the choice of kernel and its optimization are important orthogonal questions that we defer to future work. The \scorename (\textbf{\scoreacc}) introduces two hyperparameters, the lengthscale $\ell$ and the regularization $\gamma$. Both are in principle tunable via cross-validation on the calibration scores or through the Gaussian process marginal likelihood implied by Proposition~\ref{prop:gp_connection}, and under split calibration both can be selected on the fit fold at no cost to validity. We use simple heuristics throughout the experiments: $\gamma$ is set to the 90th percentile eigenvalue of the centered Gram matrix $\tilde{K}$ computed on the fit fold, and $\ell$ is set to half the median pairwise distance between fit fold residuals, the standard heuristic \cite{gretton2012kernel}. We compare against the \textbf{Mahalanobis score} (Definition~\ref{def:mahalanobis}), the \textbf{Bonferroni method} (univariate conformal prediction per coordinate at level $\alpha/d$), the \textbf{optimal transport score} of Klein et al. \cite{klein2025multivariate}, and a \textbf{density-only score} using $\mathrm{MMD}^2(\delta_\varepsilon, Q_T)$ alone (the $\gamma \to \infty$ limit of our score). All of these, like our score, produce regions of fixed shape rather than regions that adapt to the covariates. Sensitivity to the lengthscale and the regularization is analyzed in Appendix~\ref{appendix:hyperparameter_grid}, where coverage holds at the nominal level throughout and the default choice sits in the interior of the region where volumes are stable. Both the kernel score and the density-only score share the same lengthscale, so differences are attributable to the KPCA correction term alone.
For each dataset, we fit a \textbf{linear model} and a \textbf{Multi-Layer Perceptron (MLP)} (Appendix~\ref{appendix:synthetic_models}) to obtain the residuals and apply all scores to the resulting residuals (see Appendix~\ref{sec:volume_estimation} for volume estimation details).


\subsection{Synthetic Data}
\label{sec:exp:2d}

We generate $n = 10{,}000$ samples from a bivariate regression task (details in Appendix~\ref{appendix:synthetic_dgp}) with non-Gaussian noise exhibiting nonlinear dependence between the output coordinates.
We fit both a linear regression and a neural network (2-layer MLP), and apply split conformal prediction with a 50/25/25 train/calibration/test split, the calibration set being further divided into two equal folds, a fit fold on which the score is constructed and a quantile fold on which the threshold is computed. Results are averaged over 30 random seeds.

\begin{table}[t]
  \centering
  \resizebox{\textwidth}{!}{%
  \begin{tabular}{ll ccc ccc }
    \toprule
    & & \multicolumn{3}{c}{Linear} & \multicolumn{3}{c}{NN} \\
    \cmidrule(lr){3-5} \cmidrule(lr){6-8}
    $\alpha$ & Method & Coverage & Volume & WSC & Coverage & Volume & WSC \\
    \midrule
    \multirow{5}{*}{0.1} & Bonferroni  & 0.89884\tiny{$\pm$0.01087} & 4.09042\tiny{$\pm$0.09801} & 0.70587\tiny{$\pm$0.04177} & 0.90024\tiny{$\pm$0.01004} & 0.52885\tiny{$\pm$0.01897} & 0.86253\tiny{$\pm$0.02117} \\
    & Mahal  & 0.90008\tiny{$\pm$0.01316} & 2.82873\tiny{$\pm$0.09539} & 0.77820\tiny{$\pm$0.03143} & 0.89967\tiny{$\pm$0.01017} & 0.52723\tiny{$\pm$0.02127} & 0.86703\tiny{$\pm$0.01798} \\
    & OTCP  & 0.89967\tiny{$\pm$0.01139} & 2.94837\tiny{$\pm$0.09241} & 0.77220\tiny{$\pm$0.03784} & 0.89945\tiny{$\pm$0.01123} & 0.52438\tiny{$\pm$0.02037} & 0.85886\tiny{$\pm$0.02320} \\
    & Density  & 0.89749\tiny{$\pm$0.01217} & 3.15115\tiny{$\pm$0.11313} & 0.68213\tiny{$\pm$0.05181} & 0.89937\tiny{$\pm$0.01051} & 0.56598\tiny{$\pm$0.02350} & 0.86246\tiny{$\pm$0.02156} \\
    & \textbf{Kernel}  & 0.90085\tiny{$\pm$0.01311} & \textbf{2.49646\tiny{$\pm$0.08357}} & 0.81608\tiny{$\pm$0.02734} & 0.89932\tiny{$\pm$0.01052} & \textbf{0.47828\tiny{$\pm$0.01605}} & 0.85339\tiny{$\pm$0.02426} \\
    \midrule
    \multirow{5}{*}{0.05} & Bonferroni  & 0.94961\tiny{$\pm$0.00830} & 4.97000\tiny{$\pm$0.15966} & 0.85080\tiny{$\pm$0.03324} & 0.95141\tiny{$\pm$0.00812} & 0.72105\tiny{$\pm$0.03409} & 0.92713\tiny{$\pm$0.01503} \\
    & Mahal  & 0.95021\tiny{$\pm$0.00828} & 3.56900\tiny{$\pm$0.18571} & 0.85280\tiny{$\pm$0.02314} & 0.95020\tiny{$\pm$0.00771} & 0.75634\tiny{$\pm$0.04378} & 0.93218\tiny{$\pm$0.01135} \\
    & OTCP  & 0.95024\tiny{$\pm$0.00922} & 3.60526\tiny{$\pm$0.12250} & 0.88460\tiny{$\pm$0.02570} & 0.95028\tiny{$\pm$0.00877} & 0.72310\tiny{$\pm$0.04076} & 0.92352\tiny{$\pm$0.01806} \\
    & Density  & 0.95005\tiny{$\pm$0.00916} & 3.88299\tiny{$\pm$0.14726} & 0.84880\tiny{$\pm$0.03552} & 0.95053\tiny{$\pm$0.00896} & 0.77974\tiny{$\pm$0.04347} & 0.92633\tiny{$\pm$0.01619} \\
    & \textbf{Kernel}  & 0.95071\tiny{$\pm$0.00714} & \textbf{3.06383\tiny{$\pm$0.10846}} & 0.90239\tiny{$\pm$0.01614} & 0.95119\tiny{$\pm$0.00857} & \textbf{0.64310\tiny{$\pm$0.03299}} & 0.91893\tiny{$\pm$0.02269} \\
    \midrule
    \multirow{5}{*}{0.02} & Bonferroni  & 0.98089\tiny{$\pm$0.00484} & 5.99881\tiny{$\pm$0.24681} & 0.94186\tiny{$\pm$0.01733} & 0.98029\tiny{$\pm$0.00604} & 1.00442\tiny{$\pm$0.08209} & 0.96793\tiny{$\pm$0.01266} \\
    & Mahal  & 0.98051\tiny{$\pm$0.00450} & 4.94703\tiny{$\pm$0.35037} & 0.92787\tiny{$\pm$0.01765} & 0.98076\tiny{$\pm$0.00481} & 1.15531\tiny{$\pm$0.10972} & 0.96973\tiny{$\pm$0.00851} \\
    & OTCP  & 0.97963\tiny{$\pm$0.00483} & 4.39360\tiny{$\pm$0.14539} & 0.95193\tiny{$\pm$0.01369} & 0.97977\tiny{$\pm$0.00410} & 0.99319\tiny{$\pm$0.05870} & 0.96453\tiny{$\pm$0.01110} \\
    & Density  & 0.98093\tiny{$\pm$0.00484} & 4.70010\tiny{$\pm$0.16847} & 0.94566\tiny{$\pm$0.01715} & 0.98000\tiny{$\pm$0.00446} & 1.05032\tiny{$\pm$0.06910} & 0.96633\tiny{$\pm$0.01088} \\
    & \textbf{Kernel}  & 0.97951\tiny{$\pm$0.00643} & \textbf{3.83940\tiny{$\pm$0.19987}} & 0.95633\tiny{$\pm$0.01395} & 0.97996\tiny{$\pm$0.00535} & \textbf{0.85463\tiny{$\pm$0.05454}} & 0.96093\tiny{$\pm$0.01532} \\
    \midrule
    \multirow{5}{*}{0.01} & Bonferroni  & 0.99089\tiny{$\pm$0.00363} & 6.72434\tiny{$\pm$0.29196} & 0.97039\tiny{$\pm$0.01352} & 0.99061\tiny{$\pm$0.00313} & 1.28914\tiny{$\pm$0.12235} & 0.98299\tiny{$\pm$0.00538} \\
    & Mahal  & 0.99059\tiny{$\pm$0.00347} & 6.30128\tiny{$\pm$0.59552} & 0.96333\tiny{$\pm$0.01394} & 0.99092\tiny{$\pm$0.00250} & 1.56025\tiny{$\pm$0.17410} & 0.98392\tiny{$\pm$0.00366} \\
    & OTCP  & 0.99072\tiny{$\pm$0.00367} & 5.04567\tiny{$\pm$0.20328} & 0.97633\tiny{$\pm$0.01096} & 0.99113\tiny{$\pm$0.00358} & 1.25061\tiny{$\pm$0.10896} & 0.98386\tiny{$\pm$0.00682} \\
    & Density  & 0.99060\tiny{$\pm$0.00348} & 5.29902\tiny{$\pm$0.21313} & 0.97406\tiny{$\pm$0.01225} & 0.99129\tiny{$\pm$0.00320} & 1.29962\tiny{$\pm$0.09400} & 0.98420\tiny{$\pm$0.00544} \\
    & \textbf{Kernel}  & 0.98987\tiny{$\pm$0.00384} & \textbf{4.41557\tiny{$\pm$0.24420}} & 0.97740\tiny{$\pm$0.00964} & 0.98988\tiny{$\pm$0.00383} & \textbf{1.04220\tiny{$\pm$0.08911}} & 0.97893\tiny{$\pm$0.00985} \\
    \bottomrule
  \end{tabular}
  }
  \vspace{0.5em}
  \caption{Results on synthetic data across 30 seeds. All values reported as mean $\pm$ std. Bold indicates best volume per model and $\alpha$.}
  \label{tab:main_results_C}
\end{table}

\begin{figure*}[h!]
    \begin{center}
       \subfigure[Linear model residuals]
        {\includegraphics[width=0.5\textwidth]{plots_rebuttal/scatter_linear_SPLIT_fixed.pdf}}
        \subfigure[MLP residuals]
        {\includegraphics[width=0.4855\textwidth]{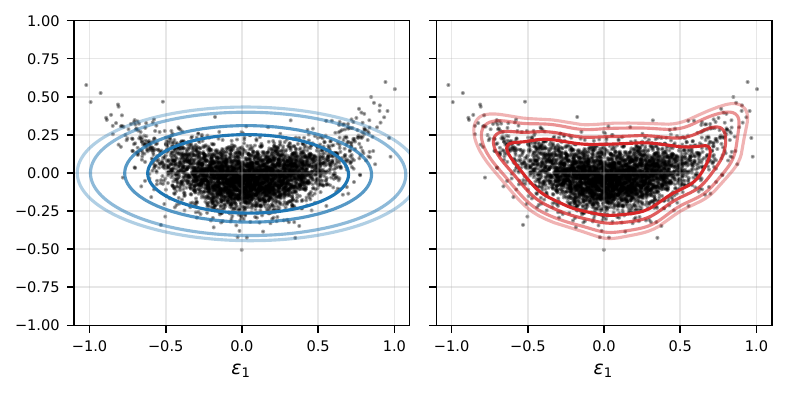}}
       \caption{%
    Prediction regions on synthetic data at coverage levels $1-\alpha \in \{0.90, 0.95, 0.98, 0.99\}$. In each panel, left: Mahalanobis (blue); right: \scoreaccsp (red), overlaid on calibration residuals}
    \label{fig:scatter_2d}
    \end{center}
    \vspace{-1em}
\end{figure*}

Figure~\ref{fig:scatter_2d} displays the prediction regions for both models across four coverage levels. Under the linear model (Figure~\ref{fig:scatter_2d}a), the residuals exhibit clear non-elliptical structure due to model misspecification. The ellipsoids must expand into low-density regions to achieve the target coverage, wasting volume on regions where residuals are unlikely to occur. The \scoreaccsp contour follows the geometry of the calibration data, producing tighter regions that exclude these areas. Under the neural network (Figure~\ref{fig:scatter_2d}b), the residuals are closer to elliptical but the \scoreaccsp retains a volume advantage.

Table~\ref{tab:main_results_C} reports coverage, volume, and worst-slab coverage (WSC) across all methods, models, and coverage levels. 
The \scoreaccsp achieves the smallest volume at every $\alpha$ level under both models while maintaining nominal marginal coverage, with WSC reported as an approximation of conditional coverage\footnote{{WSC searches over linear slabs, structurally favoring methods whose regions align with linear subspaces, e.g. Mahalanobis ellipsoid and isotropic density level sets. The anisotropic KPCA term can produce boundaries less aligned with linear slabs even when the region is tighter. Comparable WSC at smaller volume therefore indicates a genuinely tighter region rather than a loss of conditional coverage.}} (details Appendix~\ref{sec:wsc}). 
The kernel score produces prediction regions whose volumes are $9$ to $33$\% smaller than Mahalanobis and $15$ to $21$\% smaller than the density-only score. The advantage over the ellipsoid grows at tighter coverage levels. As the target coverage increases, the ellipsoid must extend into the tails of the residual distribution where the calibration data provides little support, and the volume cost of this expansion grows faster for the ellipsoid than for the \scoreaccsp contour, which expands along the support of the data. This holds under both the misspecified linear model and the neural network, where the distribution is closer to elliptical. 

The WSC supports this reading. Under the linear model, the \scoreaccsp achieves the best WSC at every $\alpha$ level, so the volume reduction reflects a genuinely tighter fit rather than a sacrifice in conditional coverage. The density-only score produces larger regions under both models and at every level, isolating the KPCA correction as the source of the improvement rather than density estimation alone.

\subsection{Real Data}

We evaluate the method on three regression datasets previously used in Feldman et al.~\cite{feldman2022} and Romano et al.~\cite{romano2019conformalized}, where we augment the original univariate responses with additional variables from the same source, producing response dimensions $d = 2,3,$ and $4$. {On the House dataset we further extend the response to $d = 6, 8$ and $10$. Volumes are estimated via Monte Carlo (Appendix~\ref{sec:volume_estimation}), and the number of samples required grows exponentially with the output dimension, which limits our evaluation to $d \le 10$. We leave the evaluation of higher-dimensional responses to future work.} All details are provided in Appendix~\ref{appendix:datasets}. We report results from linear model residuals. MLP regression results are provided in Appendix~\ref{appendix:full_results}.

Table~\ref{tab:real_data} reports the volume ratio of regions produced using the \scorename relative to the Mahalanobis baseline, computed over 50 random splits. The \scoreaccsp produces smaller prediction regions in every dataset, dimension and coverage level, with the largest gains at higher output dimension and tighter coverage. As the number of output coordinates grows, correlations between outputs create residual distributions that an ellipsoid cannot approximate without enclosing large empty regions, and the \scoreaccsp exploits this structure.

\begin{table}[H]
  \centering
  \resizebox{\textwidth}{!}{%
  \begin{tabular}{ll cccc cccc}
    \toprule
    & & \multicolumn{4}{c}{Volume Ratio (MKS\,/\,Mahalanobis)} & \multicolumn{4}{c}{MKS Coverage} \\
    \cmidrule(lr){3-6} \cmidrule(lr){7-10}
    Dataset & $d$ & $1{-}\alpha{=}.90$ & $.95$ & $.98$ & $.99$ & $.90$ & $.95$ & $.98$ & $.99$ \\
    \midrule
    \multirow{3}{*}{Bio}
    & 2 & 0.55604\tiny{$\pm$0.141} & 0.57973\tiny{$\pm$0.142} & 0.62595\tiny{$\pm$0.147} & 0.74917\tiny{$\pm$0.166} & 0.89959\tiny{$\pm$0.00639} & 0.95005\tiny{$\pm$0.00416} & 0.97995\tiny{$\pm$0.00234} & 0.98993\tiny{$\pm$0.00167} \\
    & 3 & 0.48217\tiny{$\pm$0.123} & 0.47716\tiny{$\pm$0.119} & 0.44091\tiny{$\pm$0.103} & 0.51442\tiny{$\pm$0.125} & 0.89971\tiny{$\pm$0.00564} & 0.95023\tiny{$\pm$0.00383} & 0.97998\tiny{$\pm$0.00227} & 0.99014\tiny{$\pm$0.00175} \\
    & 4 & 0.40794\tiny{$\pm$0.102} & 0.38616\tiny{$\pm$0.096} & 0.42126\tiny{$\pm$0.107} & 0.34347\tiny{$\pm$0.060} & 0.89978\tiny{$\pm$0.00618} & 0.95023\tiny{$\pm$0.00400} & 0.98017\tiny{$\pm$0.00250} & 0.99004\tiny{$\pm$0.00164} \\
    \midrule
    \multirow{3}{*}{Blog}
    & 2 & 0.45632\tiny{$\pm$0.051} & 0.62194\tiny{$\pm$0.068} & 0.75142\tiny{$\pm$0.062} & 0.61701\tiny{$\pm$0.075} & 0.90024\tiny{$\pm$0.00457} & 0.94983\tiny{$\pm$0.00352} & 0.98024\tiny{$\pm$0.00216} & 0.99018\tiny{$\pm$0.00152} \\
    & 3 & 0.52974\tiny{$\pm$0.055} & 0.58295\tiny{$\pm$0.050} & 0.47078\tiny{$\pm$0.053} & 0.32899\tiny{$\pm$0.062} & 0.89961\tiny{$\pm$0.00454} & 0.95007\tiny{$\pm$0.00336} & 0.98014\tiny{$\pm$0.00233} & 0.99015\tiny{$\pm$0.00158} \\
    & 4 & 0.23808\tiny{$\pm$0.080} & 0.28461\tiny{$\pm$0.086} & 0.20771\tiny{$\pm$0.090} & 0.11870\tiny{$\pm$0.045} & 0.89926\tiny{$\pm$0.00418} & 0.94993\tiny{$\pm$0.00354} & 0.98017\tiny{$\pm$0.00219} & 0.99001\tiny{$\pm$0.00176} \\
    \midrule
    \multirow{6}{*}{House}
    & 2 & 0.87402\tiny{$\pm$0.022} & 0.90569\tiny{$\pm$0.028} & 0.84463\tiny{$\pm$0.058} & 0.70139\tiny{$\pm$0.089} & 0.90038\tiny{$\pm$0.00632} & 0.95031\tiny{$\pm$0.00436} & 0.97987\tiny{$\pm$0.00278} & 0.98999\tiny{$\pm$0.00229} \\
    & 3 & 0.88351\tiny{$\pm$0.027} & 0.91212\tiny{$\pm$0.039} & 0.74368\tiny{$\pm$0.060} & 0.63522\tiny{$\pm$0.076} & 0.90041\tiny{$\pm$0.00700} & 0.95067\tiny{$\pm$0.00448} & 0.97996\tiny{$\pm$0.00277} & 0.99043\tiny{$\pm$0.00202} \\
    & 4 & 0.47955\tiny{$\pm$0.068} & 0.51372\tiny{$\pm$0.061} & 0.41236\tiny{$\pm$0.076} & 0.30610\tiny{$\pm$0.097} & 0.90203\tiny{$\pm$0.00793} & 0.95054\tiny{$\pm$0.00540} & 0.98002\tiny{$\pm$0.00383} & 0.99008\tiny{$\pm$0.00268} \\
    & 6 & 0.22313\tiny{$\pm$0.034} & 0.20704\tiny{$\pm$0.032} & 0.18653\tiny{$\pm$0.045} & 0.14300\tiny{$\pm$0.047} & 0.90138\tiny{$\pm$0.00790} & 0.95037\tiny{$\pm$0.00524} & 0.98023\tiny{$\pm$0.00392} & 0.98999\tiny{$\pm$0.00267} \\
    & 8 & 0.13958\tiny{$\pm$0.035} & 0.19108\tiny{$\pm$0.048} & 0.11514\tiny{$\pm$0.048} & 0.07368\tiny{$\pm$0.037} & 0.90165\tiny{$\pm$0.00743} & 0.95123\tiny{$\pm$0.00551} & 0.98041\tiny{$\pm$0.00387} & 0.98981\tiny{$\pm$0.00268} \\
    & 10 & 0.20987\tiny{$\pm$0.052} & 0.29878\tiny{$\pm$0.076} & 0.11135\tiny{$\pm$0.053} & 0.07838\tiny{$\pm$0.048} & 0.90120\tiny{$\pm$0.00741} & 0.95031\tiny{$\pm$0.00611} & 0.98051\tiny{$\pm$0.00370} & 0.98980\tiny{$\pm$0.00276} \\
    \bottomrule
  \end{tabular}}
  \vspace{0.5em}
  \caption{Real-data results under Ridge residuals, 50 seeds, with split
  calibration and $\ell{=}$med$/2$. Left: per-seed volume ratio of MKS to
  Mahalanobis (below~1 = smaller MKS region). Right: MKS empirical coverage
  (target $1{-}\alpha$).}
  \label{tab:real_data}
\end{table}

Table~\ref{tab:real_data} also confirms that coverage is maintained across all configurations. The empirical coverage of the \scoreaccsp tracks the nominal level within sampling variability at every dataset, dimension and coverage level. Coverage values are comparable across all methods, consistent with the distribution-free guarantee of the conformal framework. \wx{Moreover,} all methods 
achieve comparable approximate conditional coverage in higher dimensions and tighter coverage levels, as measured empirically through the WSC 
(Appendix~\ref{appendix:full_results}).

Full results including Bonferroni and density-based nonconformity scores are reported in Appendix~\ref{appendix:full_results}. The Bonferroni method, ignoring correlations and yielding hyper-rectangles, produces the largest volumes across all settings. Under the median heuristic the density-only score is a strong baseline  {for $d \le 4$, improving on Mahalanobis in almost every configuration. In the high-dimensional House settings it falls behind the ellipsoid in most configurations, consistent with the loss of directional information described in Section~\ref{sec:kmd:special}}. The \scoreacc, which adds the KPCA correction to the same density estimate, produces the smallest regions in almost every configuration. Overall, the \scorename maintains nominal coverage while reducing region volumes by $9$ to $93$\% relative to the Mahalanobis baseline, with the largest gains at higher dimensions and tighter coverage levels.

\section{Conclusion}
We introduced the \scorename (\scoreacc) for multivariate conformal prediction that generalizes the Mahalanobis distance to reproducing kernel Hilbert spaces, producing prediction regions adapted to the residual distribution without imposing a parametric shape. The \wx{proposed} score 
\wx{can be perceived via two distinct perspectives: (1)}
the Gaussian process posterior variance;  
\wx{(2) the decomposition} 
into a density term and a directional covariance correction term. \wx{The two-sided interpretation of \scoreaccsp allows us to  connect} Bayesian uncertainty quantification with distribution-free coverage guarantees \wx{of frequentist-type}. By compressing multivariate residuals into scalars through the RKHS geometry, the \scoreaccsp defines a nonparametric center-outward ordering on $\mathbb{R}^d$, with the conformal threshold selecting the quantile level. Finite-sample convergence rates depend on the effective rank of the RKHS covariance operator rather than the ambient dimension. On real data,
the \scoreaccsp significantly reduces prediction region volumes relative to the Mahalanobis 
baseline while maintaining nominal coverage, with the largest gains at higher dimensions and tighter coverage levels. 
Future research directions include adaptation to non-exchangeable settings such as \wx{incorporating} covariate shift, and conformal prediction for function-valued responses.

\begin{ack}
LM is supported by G-Research through the NextGen Scholarship. WX is partly funded by research grant UKRI3926. We thank Michael Gutmann and Futoshi Futami for helpful discussions.
\end{ack}

\bibliography{references}

@book{vovk2005algorithmic,
  title={Algorithmic learning in a random world},
  author={Vovk, Vladimir and Gammerman, Alexander and Shafer, Glenn},
  year={2005},
  publisher={Springer}
}

@article{shafer2008tutorial,
  title={A tutorial on conformal prediction.},
  author={Shafer, Glenn and Vovk, Vladimir},
  journal={Journal of machine learning research},
  volume={9},
  number={3},
  year={2008}
}

@inproceedings{papadopoulos2002inductive,
  title={Inductive confidence machines for regression},
  author={Papadopoulos, Harris and Proedrou, Kostas and Vovk, Volodya and Gammerman, Alex},
  booktitle={European conference on machine learning},
  pages={345--356},
  year={2002},
  organization={Springer}
}

@article{lei2018distribution,
  title={Distribution-free predictive inference for regression},
  author={Lei, Jing and G’Sell, Max and Rinaldo, Alessandro and Tibshirani, Ryan J and Wasserman, Larry},
  journal={Journal of the American Statistical Association},
  volume={113},
  number={523},
  pages={1094--1111},
  year={2018},
  publisher={Taylor \& Francis}
}

@misc{angelopoulos2022gentle,
      title={A Gentle Introduction to Conformal Prediction and Distribution-Free Uncertainty Quantification}, 
      author={Anastasios N. Angelopoulos and Stephen Bates},
      year={2022},
      eprint={2107.07511},
      archivePrefix={arXiv},
      primaryClass={cs.LG},
      url={https://arxiv.org/abs/2107.07511}, 
}

@article{romano2019conformalized,
  title={Conformalized quantile regression},
  author={Romano, Yaniv and Patterson, Evan and Candes, Emmanuel},
  journal={Advances in neural information processing systems},
  volume={32},
  year={2019}
}

@inproceedings{vovk2012conditional,
  title={Conditional validity of inductive conformal predictors},
  author={Vovk, Vladimir},
  booktitle={Asian conference on machine learning},
  pages={475--490},
  year={2012},
  organization={PMLR}
}

@article{foygel2021limits,
  title={The limits of distribution-free conditional predictive inference},
  author={Foygel Barber, Rina and Candes, Emmanuel J and Ramdas, Aaditya and Tibshirani, Ryan J},
  journal={Information and Inference: A Journal of the IMA},
  volume={10},
  number={2},
  pages={455--482},
  year={2021},
  publisher={Oxford University Press}
}

@article{braun2025multivariate,
  title={Multivariate conformal prediction via conformalized gaussian scoring},
  author={Braun, Sacha and Berta, Eug{\`e}ne and Jordan, Michael I and Bach, Francis},
  journal={arXiv preprint arXiv:2507.20941},
  year={2025}
}

@inproceedings{messoudi2022ellipsoidal,
  title={Ellipsoidal conformal inference for multi-target regression},
  author={Messoudi, Soundouss and Destercke, S{\'e}bastien and Rousseau, Sylvain},
  booktitle={Conformal and Probabilistic Prediction with Applications},
  pages={294--306},
  year={2022},
  organization={PMLR}
}

@article{izbicki2022cd,
  title={CD-split and HPD-split: Efficient conformal regions in high dimensions},
  author={Izbicki, Rafael and Shimizu, Gilson and Stern, Rafael B},
  journal={Journal of Machine Learning Research},
  volume={23},
  number={87},
  pages={1--32},
  year={2022}
}

@article{dheur2025unified,
  title={A unified comparative study with generalized conformity scores for multi-output conformal regression},
  author={Dheur, Victor and Fontana, Matteo and Estievenart, Yorick and Desobry, Naomi and Taieb, Souhaib Ben},
  journal={arXiv preprint arXiv:2501.10533},
  year={2025}
}

@article{thurin2025optimal,
  title={Optimal transport-based conformal prediction},
  author={Thurin, Gauthier and Nadjahi, Kimia and Boyer, Claire},
  journal={arXiv preprint arXiv:2501.18991},
  year={2025}
}

@article{klein2025multivariate,
  title={Multivariate conformal prediction using optimal transport},
  author={Klein, Michal and Bethune, Louis and Ndiaye, Eugene and Cuturi, Marco},
  journal={arXiv preprint arXiv:2502.03609},
  year={2025}
}

@article{braun2025minimum,
  title={Minimum volume conformal sets for multivariate regression},
  author={Braun, Sacha and Aolaritei, Liviu and Jordan, Michael I and Bach, Francis},
  journal={arXiv preprint arXiv:2503.19068},
  year={2025}
}

@article{messoudi2021copula,
  title={Copula-based conformal prediction for multi-target regression},
  author={Messoudi, Soundouss and Destercke, S{\'e}bastien and Rousseau, Sylvain},
  journal={Pattern Recognition},
  volume={120},
  pages={108101},
  year={2021},
  publisher={Elsevier}
}

@article{plassier2024probabilistic,
  title={Probabilistic conformal prediction with approximate conditional validity},
  author={Plassier, Vincent and Fishkov, Alexander and Guizani, Mohsen and Panov, Maxim and Moulines, Eric},
  journal={arXiv preprint arXiv:2407.01794},
  year={2024}
}

@book{williams2006gaussian,
  title={Gaussian Processes for Machine Learning},
  author={Rasmussen, Carl Edward and Williams, Christopher K. I.},
  publisher={MIT Press},
  address={Cambridge, MA},
  year={2006},
  isbn={0-262-18253-X}
}

@article{gretton2012kernel,
  title={A kernel two-sample test},
  author={Gretton, Arthur and Borgwardt, Karsten M and Rasch, Malte J and Sch{\"o}lkopf, Bernhard and Smola, Alexander},
  journal={The journal of machine learning research},
  volume={13},
  number={1},
  pages={723--773},
  year={2012},
  publisher={JMLR. org}
}

@article{parzen1962estimation,
  title={On estimation of a probability density function and mode},
  author={Parzen, Emanuel},
  journal={The annals of mathematical statistics},
  volume={33},
  number={3},
  pages={1065--1076},
  year={1962},
  publisher={JSTOR}
}

@article{RosenblattNonParametric,
  author={Murray Rosenblatt},
  title={{Remarks on Some Nonparametric Estimates of a Density Function}},
  volume={27},
  journal={The Annals of Mathematical Statistics},
  number={3},
  publisher={Institute of Mathematical Statistics},
  pages={832--837},
  year={1956}
}

@article{scholkopf1998nonlinear,
  title={Nonlinear component analysis as a kernel eigenvalue problem},
  author={Sch{\"o}lkopf, Bernhard and Smola, Alexander and M{\"u}ller, Klaus-Robert},
  journal={Neural computation},
  volume={10},
  number={5},
  pages={1299--1319},
  year={1998}
}

@misc{xu2024conformal,
  title={Conformal prediction for multi-dimensional time series by ellipsoidal sets}, 
  author={Chen Xu and Hanyang Jiang and Yao Xie},
  year={2024},
  eprint={2403.03850}
}

@book{rio2017asymptotic,
  title={Asymptotic Theory of Weakly Dependent Random Processes},
  author={Rio, Emmanuel},
  year={2017},
  publisher={Springer}
}

@article{koltchinskii2017concentration,
  title={Concentration inequalities and moment bounds for sample covariance operators},
  author={Koltchinskii, Vladimir and Lounici, Karim},
  journal={Bernoulli},
  pages={110--133},
  year={2017}
}

@inproceedings{johnstone2021conformal,
  title={Conformal uncertainty sets for robust optimization},
  author={Johnstone, Chancellor and Cox, Bruce},
  booktitle={Conformal and Probabilistic Prediction and Applications},
  pages={72--90},
  year={2021},
  organization={PMLR}
}

@misc{sampson2025flexibleconformalhighestpredictive,
      title={Flexible Conformal Highest Predictive Conditional Density Sets}, 
      author={Max Sampson and Kung-Sik Chan},
      year={2025},
      eprint={2406.18052}
}

@inproceedings{zaffran2023conformal,
  title={Conformal prediction with missing values},
  author={Zaffran, Margaux and Dieuleveut, Aymeric and Josse, Julie and Romano, Yaniv},
  booktitle={International Conference on Machine Learning},
  pages={40578--40604},
  year={2023}
}

@article{tibshirani2019conformal,
  title={Conformal prediction under covariate shift},
  author={Tibshirani, Ryan J and Foygel Barber, Rina and Candes, Emmanuel and Ramdas, Aaditya},
  journal={Advances in neural information processing systems},
  volume={32},
  year={2019}
}

@article{xu2023conformal,
  title={Conformal prediction for time series},
  author={Xu, Chen and Xie, Yao},
  journal={IEEE transactions on pattern analysis and machine intelligence},
  volume={45},
  number={10},
  pages={11575--11587},
  year={2023}
}

@article{clarkson2024split,
  title = 	 {Split Conformal Prediction under Data Contamination},
  author =       {Clarkson, Jason and Xu, Wenkai and Cucuringu, Mihai and Reinert, Gesine},
  journal = 	 {Proceedings of the Thirteenth Symposium on Conformal and Probabilistic Prediction with Applications},
  pages = 	 {5--27},
  year = 	 {2024},
}

@article{barber2023conformal,
  title={Conformal prediction beyond exchangeability},
  author={Barber, Rina Foygel and Candes, Emmanuel J and Ramdas, Aaditya and Tibshirani, Ryan J},
  journal={The Annals of Statistics},
  volume={51},
  number={2},
  pages={816--845},
  year={2023}
}

@article{fukumizu2004dimensionality,
  title={Dimensionality reduction for supervised learning with reproducing kernel Hilbert spaces},
  author={Fukumizu, Kenji and Bach, Francis R and Jordan, Michael I},
  journal={Journal of Machine Learning Research},
  volume={5},
  pages={73--99},
  year={2004}
}

@article{sriperumbudur2010hilbert,
  title={Hilbert space embeddings and metrics on probability measures},
  author={Sriperumbudur, Bharath K and Gretton, Arthur and Fukumizu, Kenji and Sch{\"o}lkopf, Bernhard and Lanckriet, Gert RG},
  journal={The Journal of Machine Learning Research},
  volume={11},
  pages={1517--1561},
  year={2010}
}

@misc{alaoui2015fast,
      title={Fast Randomized Kernel Methods With Statistical Guarantees}, 
      author={Ahmed El Alaoui and Michael W. Mahoney},
      year={2015},
      eprint={1411.0306}
}

@article{feldman2022,
  author  = {Shai Feldman and Stephen Bates and Yaniv Romano},
  title   = {Calibrated Multiple-Output Quantile Regression with Representation Learning},
  journal = {Journal of Machine Learning Research},
  year    = {2023},
  volume  = {24},
  number  = {24},
  pages   = {1--48},
  url     = {http://jmlr.org/papers/v24/21-1280.html}
}

@misc{blog_data,
  author       = {Buza, Krisztian},
  title        = {{BlogFeedback}},
  year         = {2014},
  howpublished = {UCI Machine Learning Repository},
  note         = {{DOI}: https://doi.org/10.24432/C58S3F}
}

@misc{bio_data,
  author       = {Rana, Prashant},
  title        = {{Physicochemical Properties of Protein Tertiary Structure}},
  year         = {2013},
  howpublished = {UCI Machine Learning Repository},
  note         = {{DOI}: https://doi.org/10.24432/C5QW3H}
}

@misc{house_data,
	author={{Harlfoxem}},
	title = {House sales in King County, {USA}.},
	howpublished = {\url{https://www.kaggle.com/harlfoxem/housesalesprediction/metadata}},
	note = {Accessed: July, 2021.}
}

@inproceedings{
sun2024copula,
title={Copula Conformal prediction for multi-step time series prediction},
author={Sophia Huiwen Sun and Rose Yu},
booktitle={The Twelfth International Conference on Learning Representations},
year={2024},
url={https://openreview.net/forum?id=ojIJZDNIBj}
}

@book{vershynin2018high,
  title={High-dimensional probability: An introduction with applications in data science},
  author={Vershynin, Roman},
  volume={47},
  year={2018},
  publisher={Cambridge university press}
}

@article{cauchois2021knowing,
  title={Knowing what you know: valid and validated confidence sets in multiclass and multilabel prediction},
  author={Cauchois, Maxime and Gupta, Suyash and Duchi, John C},
  journal={Journal of machine learning research},
  volume={22},
  number={81},
  pages={1--42},
  year={2021}
}
\bibliographystyle{plain}


\appendix


\section{Derivations and Proofs}
\label{sec:proofs}

\subsection{Derivation of Proposition~\ref{prop:kmd_computable}  (Computable Form)}
\label{proof:kernel_md}

We show that the RKHS Mahalanobis distance in Definition~\ref{def:kmd} reduces to the finite-dimensional expression in Proposition~\ref{prop:kmd_computable} via the Woodbury matrix identity.

\begin{proof}
Let $\Phi = [\phi(\varepsilon_1), \ldots, \phi(\varepsilon_T)]$ denote the collection of feature vectors in $\mathcal{H}$, and let $H = I_T - \frac{1}{T}\mathbf{1}\mathbf{1}^\top \in \mathbb{R}^{T \times T}$ be the centering matrix. The centered feature vectors are $\tilde{\Phi} = \Phi H$, and the sample covariance operator is $\widehat{C}_\phi = \frac{1}{T-1}\tilde{\Phi}\tilde{\Phi}^\top$.

We apply the Woodbury matrix identity. For $A = \lambda I_{\mathcal{H}}$, $U = \frac{1}{\sqrt{T-1}}\tilde{\Phi}$, and $C = I_T$, the identity $(A + UU^\top)^{-1} = A^{-1} - A^{-1}U(I + U^\top A^{-1}U)^{-1}U^\top A^{-1}$ gives:
\begin{equation}
\left(\widehat{C}_\phi + \lambda I_{\mathcal{H}}\right)^{-1} = \frac{1}{\lambda}I_{\mathcal{H}} - \frac{1}{\lambda^2(T-1)} \tilde{\Phi}\left(I_T + \frac{1}{\lambda(T-1)}\tilde{\Phi}^\top\tilde{\Phi}\right)^{-1}\tilde{\Phi}^\top
\label{eq:woodbury_applied}
\end{equation}

The centered Gram matrix is $\tilde{K} = \tilde{\Phi}^\top\tilde{\Phi} = HKH \in \mathbb{R}^{T \times T}$, where $K_{ij} = k(\varepsilon_i, \varepsilon_j)$. Substituting into~\eqref{eq:woodbury_applied}:
\begin{equation}
\left(\widehat{C}_\phi + \lambda I_{\mathcal{H}}\right)^{-1} = \frac{1}{\lambda}I_{\mathcal{H}} - \frac{1}{\lambda^2(T-1)} \tilde{\Phi}\left(I_T + \frac{1}{\lambda(T-1)}\tilde{K}\right)^{-1}\tilde{\Phi}^\top
\label{eq:woodbury_gram}
\end{equation}

Let $\tilde{\phi}(\varepsilon) = \phi(\varepsilon) - \hat{\mu}_\phi$ be the centered feature map of a test residual. Applying Definition~\ref{def:kmd}:
\begin{align}
\hat{e}_k(\varepsilon) &= \left\langle \tilde{\phi}(\varepsilon), \left(\widehat{C}_\phi + \lambda I_{\mathcal{H}}\right)^{-1} \tilde{\phi}(\varepsilon) \right\rangle_{\mathcal{H}} \nonumber \\
&= \frac{1}{\lambda} \tilde{k}(\varepsilon, \varepsilon) - \frac{1}{\lambda^2(T-1)} \tilde{\mathbf{k}}_*(\varepsilon)^\top \left(I_T + \frac{1}{\lambda(T-1)}\tilde{K}\right)^{-1} \tilde{\mathbf{k}}_*(\varepsilon)
\label{eq:score_expanded}
\end{align}
where $\tilde{\mathbf{k}}_*(\varepsilon) = \tilde{\Phi}^\top \tilde{\phi}(\varepsilon) \in \mathbb{R}^T$ is the centered kernel vector with entries $[\tilde{\mathbf{k}}_*(\varepsilon)]_i = \langle \tilde{\phi}(\varepsilon), \tilde{\phi}(\varepsilon_i) \rangle_{\mathcal{H}}$.

Setting $\gamma = \lambda(T-1)$, we have $I_T + \frac{1}{\gamma}\tilde{K} = \frac{1}{\gamma}(\tilde{K} + \gamma I_T)$, so $(I_T + \frac{1}{\gamma}\tilde{K})^{-1} = \gamma(\tilde{K} + \gamma I_T)^{-1}$. Substituting into~\eqref{eq:score_expanded}:
\begin{align*}
\hat{e}_k(\varepsilon)
&= \frac{1}{\lambda} \tilde{k}(\varepsilon, \varepsilon) - \frac{\gamma}{\lambda^2(T-1)} \tilde{\mathbf{k}}_*(\varepsilon)^\top (\tilde{K} + \gamma I_T)^{-1} \tilde{\mathbf{k}}_*(\varepsilon) \\
&= \frac{1}{\lambda} \left[ \tilde{k}(\varepsilon, \varepsilon) - \tilde{\mathbf{k}}_*(\varepsilon)^\top (\tilde{K} + \gamma I_T)^{-1} \tilde{\mathbf{k}}_*(\varepsilon) \right]
\end{align*}
where the second line uses $\gamma / [\lambda(T-1)] = 1$. Since the factor $1/\lambda$ is a positive constant that does not affect the ranking of residuals, we define the \scorename as in 
Proposition~\ref{prop:kmd_computable}:
\begin{equation*}
\hat{e}_k(\varepsilon)
= \tilde{k}(\varepsilon, \varepsilon)
- \tilde{\mathbf{k}}_*(\varepsilon)^\top
(\tilde{K} + \gamma I_T)^{-1}
\tilde{\mathbf{k}}_*(\varepsilon)
\end{equation*}

For completeness, the centered quantities in terms of the raw kernel are:
\begin{align*}
\tilde{k}(\varepsilon,\varepsilon)
&= k(\varepsilon,\varepsilon) - \frac{2}{T}\sum_{i=1}^T k(\varepsilon, \varepsilon_i) + \frac{1}{T^2}\sum_{i,j=1}^T k(\varepsilon_i, \varepsilon_j), \\[4pt]
[\tilde{\mathbf{k}}_*(\varepsilon)]_i
&= k(\varepsilon, \varepsilon_i)
   - \frac{1}{T}\sum_{j=1}^T k(\varepsilon, \varepsilon_j)
   - \frac{1}{T}\sum_{j=1}^T k(\varepsilon_j, \varepsilon_i)
   + \frac{1}{T^2}\sum_{j,\ell=1}^T k(\varepsilon_j, \varepsilon_\ell)
\end{align*}
\end{proof}

\subsection{Proof of Proposition~\ref{prop:gp_connection} (Gaussian Process Identity)}
\label{proof:gp_connection}

\begin{proof}
Let $g \sim \mathcal{GP}(0, \tilde{k})$ and suppose we observe $g$ at locations $\varepsilon_1, \ldots, \varepsilon_T$ with i.i.d.\ Gaussian noise of variance $\gamma$. By standard GP conditioning~\cite[Eq.~2.26]{williams2006gaussian}, the posterior variance at a test point $\varepsilon$ is
\begin{equation*}
\mathrm{Var}_{\mathrm{GP}}\bigl[g(\varepsilon) \mid \{\varepsilon_i\}_{i=1}^T\bigr]
= \tilde{k}(\varepsilon, \varepsilon)
- \tilde{\mathbf{k}}_*(\varepsilon)^\top (\tilde{K} + \gamma I_T)^{-1} \tilde{\mathbf{k}}_*(\varepsilon)
\end{equation*}
where $\tilde{K}_{ij} = \tilde{k}(\varepsilon_i, \varepsilon_j)$ and $[\tilde{\mathbf{k}}_*(\varepsilon)]_i = \tilde{k}(\varepsilon, \varepsilon_i)$. This is identical to the expression in Proposition~\ref{prop:kmd_computable}. Note that the posterior variance depends only on the observation locations and the kernel, not on the observed function values. The identity $\hat{e}_k(\varepsilon) = \mathrm{Var}_{\mathrm{GP}}[g(\varepsilon) \mid \{\varepsilon_i\}_{i=1}^T]$ follows by direct comparison.
\end{proof}

\subsection{Proof of Proposition~\ref{prop:kmd_mmd} (MMD--KPCA Decomposition)}
\label{proof:kmd_mmd}

\begin{proof}
We establish two intermediate 
steps and combine them for the result.

\paragraph{The MMD term.}
The squared maximum mean discrepancy between two distributions $P$ and $Q$ with respect to a kernel $k$ is~\cite{gretton2012kernel}
\begin{equation*}
\mathrm{MMD}^2(P, Q) = \mathbb{E}_{x, x' \sim P}\bigl[k(x, x')\bigr] + \mathbb{E}_{y, y' \sim Q}\bigl[k(y, y')\bigr] - 2\,\mathbb{E}_{x \sim P,\, y \sim Q}\bigl[k(x, y)\bigr]
\end{equation*}
For $P = \delta_\varepsilon$ (the Dirac measure at $\varepsilon$) and $Q = Q_T = \frac{1}{T}\sum_{i=1}^T \delta_{\varepsilon_i}$ (the empirical measure of the calibration residuals), each expectation evaluates as follows. Since $x = x' = \varepsilon$ almost surely under $\delta_\varepsilon$: $\mathbb{E}_{x, x' \sim \delta_\varepsilon}[k(x,x')] = k(\varepsilon, \varepsilon)$. Since $y, y'$ are drawn independently from $Q_T$: $\mathbb{E}_{y, y' \sim Q_T}[k(y,y')] = \frac{1}{T^2}\sum_{i,j} k(\varepsilon_i, \varepsilon_j) = \bar{k}$. The cross term: $\mathbb{E}_{x \sim \delta_\varepsilon, y \sim Q_T}[k(x,y)] = \frac{1}{T}\sum_i k(\varepsilon, \varepsilon_i) = \hat{p}_k(\varepsilon)$. Therefore
\begin{equation*}
\mathrm{MMD}^2(\delta_\varepsilon, Q_T) = k(\varepsilon, \varepsilon) + \bar{k} - 2\hat{p}_k(\varepsilon)
\end{equation*}

Equivalently, $\mathrm{MMD}^2(P, Q) = \|\mu_P - \mu_Q\|_{\mathcal{H}}^2$ where $\mu_P = \mathbb{E}_{x \sim P}[\phi(x)]$ is the kernel mean embedding~\cite{gretton2012kernel}. Since $\mu_{\delta_\varepsilon} = \phi(\varepsilon)$ and $\mu_{Q_T} = \hat{\mu}_\phi$, we have
\begin{align*}
\mathrm{MMD}^2(\delta_\varepsilon, Q_T)
&= \|\phi(\varepsilon) - \hat{\mu}_\phi\|_{\mathcal{H}}^2
= \|\tilde{\phi}(\varepsilon)\|_{\mathcal{H}}^2
= \tilde{k}(\varepsilon, \varepsilon)
\end{align*}

\paragraph{The KPCA correction term.}
Let $\tilde{K} = V\Lambda V^\top$ be the eigendecomposition, with $V = [\mathbf{v}_1, \ldots, \mathbf{v}_T]$ orthogonal and $\Lambda = \mathrm{diag}(\lambda_1, \ldots, \lambda_T)$. Then
\begin{equation*}
(\tilde{K} + \gamma I_T)^{-1} = V(\Lambda + \gamma I_T)^{-1}V^\top = \sum_{j=1}^T \frac{1}{\lambda_j + \gamma}\,\mathbf{v}_j\mathbf{v}_j^\top
\end{equation*}
and therefore
\begin{equation*}
\tilde{\mathbf{k}}_*(\varepsilon)^\top (\tilde{K} + \gamma I_T)^{-1} \tilde{\mathbf{k}}_*(\varepsilon)
= \sum_{j=1}^T \frac{(\mathbf{v}_j^\top \tilde{\mathbf{k}}_*(\varepsilon))^2}{\lambda_j + \gamma}
= \sum_{j=1}^T \frac{\alpha_j(\varepsilon)^2}{\lambda_j + \gamma}
\end{equation*}
where $\alpha_j(\varepsilon) = \mathbf{v}_j^\top \tilde{\mathbf{k}}_*(\varepsilon)$.

\paragraph{Verification that $\alpha_j(\varepsilon)$ is a KPCA projection.}
Define the RKHS vectors $\mathbf{w}_j = \sum_{i=1}^T [\mathbf{v}_j]_i\,\tilde{\phi}(\varepsilon_i) \in \mathcal{H}$. These are eigenfunctions of the sample covariance operator $\widehat{C}_\phi$ with eigenvalues $\lambda_j/(T-1)$:
\begin{align*}
\widehat{C}_\phi\,\mathbf{w}_j
&= \frac{1}{T-1}\sum_{\ell=1}^T \tilde{\phi}(\varepsilon_\ell)\left\langle \tilde{\phi}(\varepsilon_\ell),\,\sum_{i=1}^T [\mathbf{v}_j]_i\,\tilde{\phi}(\varepsilon_i)\right\rangle_{\mathcal{H}} \\
&= \frac{1}{T-1}\sum_{\ell=1}^T \tilde{\phi}(\varepsilon_\ell)\sum_{i=1}^T [\mathbf{v}_j]_i\,\tilde{K}_{\ell i}
= \frac{1}{T-1}\sum_{\ell=1}^T \tilde{\phi}(\varepsilon_\ell)\,[\tilde{K}\mathbf{v}_j]_\ell \\
&= \frac{\lambda_j}{T-1}\sum_{\ell=1}^T [\mathbf{v}_j]_\ell\,\tilde{\phi}(\varepsilon_\ell)
= \frac{\lambda_j}{T-1}\,\mathbf{w}_j
\end{align*}
The projection of $\tilde{\phi}(\varepsilon)$ onto $\mathbf{w}_j$ is
\begin{equation*}
\langle \tilde{\phi}(\varepsilon),\,\mathbf{w}_j\rangle_{\mathcal{H}}
= \sum_{i=1}^T [\mathbf{v}_j]_i\,\langle \tilde{\phi}(\varepsilon),\,\tilde{\phi}(\varepsilon_i)\rangle_{\mathcal{H}}
= \mathbf{v}_j^\top\tilde{\mathbf{k}}_*(\varepsilon)
= \alpha_j(\varepsilon)
\end{equation*}
confirming that $\alpha_j(\varepsilon)$ is the projection of $\tilde{\phi}(\varepsilon)$ onto the $j$-th kernel principal component.

\paragraph{Combining.}
From Proposition~\ref{prop:kmd_computable}:
\begin{equation*}
\hat{e}_k(\varepsilon)
= \tilde{k}(\varepsilon, \varepsilon) - \tilde{\mathbf{k}}_*(\varepsilon)^\top(\tilde{K} + \gamma I_T)^{-1}\tilde{\mathbf{k}}_*(\varepsilon)
= \mathrm{MMD}^2(\delta_\varepsilon, Q_T) - \sum_{j=1}^T \frac{\alpha_j(\varepsilon)^2}{\lambda_j + \gamma}
\end{equation*}

{
Here, for $\lambda_j > 0$, $\beta_j(\varepsilon) = \lambda_j^{-1/2}\,\alpha_j(\varepsilon) = \langle \tilde{\phi}(\varepsilon), \mathbf{w}_j/\sqrt{\lambda_j} \rangle_{\mathcal{H}}$ is the projection of $\tilde{\phi}(\varepsilon)$ onto the normalized principal component $\mathbf{w}_j/\sqrt{\lambda_j}$, which has unit norm since $\|\mathbf{w}_j\|_{\mathcal{H}}^2 = \mathbf{v}_j^\top \tilde{K}\,\mathbf{v}_j = \lambda_j$.
For $\lambda_j = 0$, 
$\|\mathbf{w}_j\|_{\mathcal{H}}^2 = \mathbf{v}_j^\top \tilde{K}\, \mathbf{v}_j = \lambda_j = 0$,
so $\mathbf{w}_j = 0$ and $\alpha_j(\varepsilon) = \langle \tilde{\phi}(\varepsilon), \mathbf{w}_j \rangle_{\mathcal{H}} = 0$, and these terms vanish. For $\lambda_j > 0$, $\alpha_j(\varepsilon) = \sqrt{\lambda_j}\,\beta_j(\varepsilon)$, hence
\begin{equation*}
\frac{\alpha_j(\varepsilon)^2}{\lambda_j + \gamma} = \frac{\lambda_j}{\lambda_j + \gamma}\,\beta_j(\varepsilon)^2.
\end{equation*}
Substituting into the above gives
\begin{equation*}
\hat{e}_k(\varepsilon) = \mathrm{MMD}^2(\delta_\varepsilon, Q_T) - \sum_{j:\,\lambda_j > 0} \frac{\lambda_j}{\lambda_j + \gamma}\, \beta_j(\varepsilon)^2. \qedhere
\end{equation*}
}
\end{proof}



\subsection{Proof of Proposition~\ref{prop:linear_kernel} (Linear Kernel Recovery)}
\label{proof:linear_kernel}

\begin{proof}
For the linear kernel $k(\mathbf{x}, \mathbf{x}') = \mathbf{x}^\top\mathbf{x}'$, the feature map is $\phi(\mathbf{x}) = \mathbf{x}$, so $\hat{\mu}_\phi = \bar{\varepsilon}$, $\tilde{\phi}(\varepsilon_i) = \varepsilon_i - \bar{\varepsilon}$, and $\tilde{K}_{ij} = (\varepsilon_i - \bar{\varepsilon})^\top(\varepsilon_j - \bar{\varepsilon})$.

Let $\tilde{E} = [\varepsilon_1 - \bar{\varepsilon}, \ldots, \varepsilon_T - \bar{\varepsilon}] \in \mathbb{R}^{d \times T}$. Then $\tilde{K} = \tilde{E}^\top\tilde{E}$, $\tilde{\mathbf{k}}_*(\varepsilon) = \tilde{E}^\top(\varepsilon - \bar{\varepsilon})$, and $\tilde{k}(\varepsilon, \varepsilon) = \|\varepsilon - \bar{\varepsilon}\|^2$. The score in Proposition~\ref{prop:kmd_computable} becomes:
\begin{align}
\hat{e}_k(\varepsilon)
&= \|\varepsilon - \bar{\varepsilon}\|^2 - (\varepsilon - \bar{\varepsilon})^\top\tilde{E}\,(\tilde{E}^\top\tilde{E} + \gamma I_T)^{-1}\tilde{E}^\top(\varepsilon - \bar{\varepsilon}) \nonumber \\
&= (\varepsilon - \bar{\varepsilon})^\top \left[I_d - \tilde{E}\,(\tilde{E}^\top\tilde{E} + \gamma I_T)^{-1}\tilde{E}^\top\right](\varepsilon - \bar{\varepsilon})
\label{eq:linear_intermediate}
\end{align}

By the push-through identity, $\tilde{E}\,(\tilde{E}^\top\tilde{E} + \gamma I_T)^{-1}\tilde{E}^\top = (\tilde{E}\,\tilde{E}^\top + \gamma I_d)^{-1}\tilde{E}\,\tilde{E}^\top$. Substituting into~\eqref{eq:linear_intermediate}:
\begin{align*}
\hat{e}_k(\varepsilon)
&= (\varepsilon - \bar{\varepsilon})^\top\left[I_d - (\tilde{E}\,\tilde{E}^\top + \gamma I_d)^{-1}\tilde{E}\,\tilde{E}^\top\right](\varepsilon - \bar{\varepsilon}) \\
&= (\varepsilon - \bar{\varepsilon})^\top(\tilde{E}\,\tilde{E}^\top + \gamma I_d)^{-1}\left[(\tilde{E}\,\tilde{E}^\top + \gamma I_d) - \tilde{E}\,\tilde{E}^\top\right](\varepsilon - \bar{\varepsilon}) \\
&= \gamma \cdot (\varepsilon - \bar{\varepsilon})^\top(\tilde{E}\,\tilde{E}^\top + \gamma I_d)^{-1}(\varepsilon - \bar{\varepsilon})
\end{align*}

Since $\tilde{E}\,\tilde{E}^\top = (T-1)\widehat{\Sigma}$ and $\gamma = \lambda(T-1)$:
\begin{align*}
\hat{e}_k(\varepsilon)
&= \lambda(T-1) \cdot (\varepsilon - \bar{\varepsilon})^\top\bigl[(T-1)(\widehat{\Sigma} + \lambda I_d)\bigr]^{-1}(\varepsilon - \bar{\varepsilon}) \\
&= \lambda \cdot (\varepsilon - \bar{\varepsilon})^\top(\widehat{\Sigma} + \lambda I_d)^{-1}(\varepsilon - \bar{\varepsilon})
= \lambda \cdot \mathrm{MD}(\varepsilon)
\end{align*}

The constant factor $\lambda > 0$ does not affect the ranking of residuals, so the \scoreaccsp and the regularized Mahalanobis distance produce identical conformal prediction regions.
\end{proof}

\label{appendix_section:spectral}

\subsection{Proof of Proposition~\ref{prop:spectral}
(Spectral Decomposition)}
\label{proof:spectral}

\begin{proof}
Throughout, sums run over the indices with $\lambda_j > 0$, since the remaining terms vanish by the proof of Proposition~\ref{prop:kmd_mmd}.
From the proof of Proposition~\ref{prop:kmd_mmd}, the projection of $\tilde{\phi}(\varepsilon)$ onto $\mathbf{w}_j$ satisfies $\langle \tilde{\phi}(\varepsilon), \mathbf{w}_j \rangle_{\mathcal{H}} = \alpha_j(\varepsilon)$, so the projection onto the normalized direction is
\begin{equation*}
\langle \tilde{\phi}(\varepsilon), \mathbf{e}_j \rangle_{\mathcal{H}} = \frac{\alpha_j(\varepsilon)}{\sqrt{\lambda_j}}
\end{equation*}

Since the $\mathbf{e}_j$ are orthonormal and $\tilde{\phi}_\perp(\varepsilon)$ is orthogonal to all of them, Pythagoras' theorem gives
\begin{equation*}
\|\tilde{\phi}(\varepsilon)\|_{\mathcal{H}}^2
= \sum_{j=1}^T \langle \tilde{\phi}(\varepsilon), \mathbf{e}_j \rangle_{\mathcal{H}}^2 + \|\tilde{\phi}_\perp(\varepsilon)\|_{\mathcal{H}}^2
= \sum_{j=1}^T \frac{\alpha_j(\varepsilon)^2}{\lambda_j} + \|\tilde{\phi}_\perp(\varepsilon)\|_{\mathcal{H}}^2
\end{equation*}

Substituting into the MMD--KPCA decomposition (Proposition~\ref{prop:kmd_mmd}), with $\|\tilde{\phi}(\varepsilon)\|_{\mathcal{H}}^2 = \mathrm{MMD}^2(\delta_\varepsilon, Q_T)$:
\begin{align*}
\hat{e}_k(\varepsilon)
&= \|\tilde{\phi}(\varepsilon)\|_{\mathcal{H}}^2 - \sum_{j=1}^T \frac{\alpha_j(\varepsilon)^2}{\lambda_j + \gamma} \\
&= \sum_{j=1}^T \frac{\alpha_j(\varepsilon)^2}{\lambda_j} + \|\tilde{\phi}_\perp(\varepsilon)\|_{\mathcal{H}}^2 - \sum_{j=1}^T \frac{\alpha_j(\varepsilon)^2}{\lambda_j + \gamma} \\
&= \sum_{j=1}^T \alpha_j(\varepsilon)^2 \left[\frac{1}{\lambda_j} - \frac{1}{\lambda_j + \gamma}\right] + \|\tilde{\phi}_\perp(\varepsilon)\|_{\mathcal{H}}^2 \\
&= \sum_{j=1}^T \frac{\gamma \cdot \alpha_j(\varepsilon)^2}{\lambda_j(\lambda_j + \gamma)} + \|\tilde{\phi}_\perp(\varepsilon)\|_{\mathcal{H}}^2
\end{align*}

{
Since $\alpha_j(\varepsilon)^2/\lambda_j = \beta_j(\varepsilon)^2$ for $\lambda_j > 0$, this gives
\begin{equation*}
\hat{e}_k(\varepsilon) = \sum_{j:\,\lambda_j > 0} \frac{\gamma}{\lambda_j + \gamma}\, \beta_j(\varepsilon)^2 + \|\tilde{\phi}_\perp(\varepsilon)\|_{\mathcal{H}}^2. \qedhere
\end{equation*}
}
\end{proof}

\subsection{Proof of Proposition~\ref{prop:leverage}
(Leverage Score Identity)}
\label{proof:leverage}

\begin{proof}
For a calibration residual $\varepsilon_i$,
$\tilde{\phi}(\varepsilon_i)$ lies in the span of the
calibration data in $\mathcal{H}$ by construction, so
$\|\tilde{\phi}_\perp(\varepsilon_i)\|_{\mathcal{H}}^2
= 0$ and the spectral decomposition
(Proposition~\ref{prop:spectral}) reduces to
\begin{equation*}
\hat{e}_k(\varepsilon_i)
= \sum_{j=1}^T
  \frac{\gamma\,\alpha_j(\varepsilon_i)^2}
       {\lambda_j(\lambda_j + \gamma)}.
\end{equation*}

Since
$\tilde{\mathbf{k}}_*(\varepsilon_i) = \tilde{K}_{:,i}$
and $\tilde{K}\mathbf{v}_j = \lambda_j\mathbf{v}_j$,
the KPCA projection at a calibration point is
$\alpha_j(\varepsilon_i)
= \mathbf{v}_j^\top \tilde{K}_{:,i}
= \lambda_j [\mathbf{v}_j]_i$.
Substituting,
\begin{equation*}
\hat{e}_k(\varepsilon_i)
= \sum_{j=1}^T
  \frac{\gamma\,\lambda_j\,[\mathbf{v}_j]_i^2}
       {\lambda_j + \gamma}
= \gamma \sum_{j=1}^T
  \frac{\lambda_j}{\lambda_j + \gamma}\,
  [\mathbf{v}_j]_i^2
= \gamma \cdot h_{ii}
\end{equation*}
where
$h_{ii} = [V\Lambda(\Lambda + \gamma I_T)^{-1}V^\top]_{ii}$
is the $i$-th diagonal entry of the matrix
$H = \tilde{K}(\tilde{K} + \gamma I_T)^{-1}$.
\end{proof}

\section{Theoretical Analysis}
\label{sec:theory_proofs}

This section provides the full theoretical analysis for the coverage guarantees stated in Theorem~\ref{thm:coverage}. We first state the assumptions and auxiliary results, then prove asymptotic conditional coverage under i.i.d. residuals (Section~\ref{sec:iid_theory}), where the standard exogeneity condition makes the conditional guarantee valid, and finally establish marginal coverage under $\alpha$-mixing dependence (Section~\ref{sec:mixing}), where temporal correlation between $X_{T+1}$ and $\varepsilon_{T+1}$ precludes a conditional guarantee.

Throughout this section, we work at the RKHS operator level. The true nonconformity score at time $t$ is
\begin{equation*}
e_t = \langle \phi(\varepsilon_t) - \mu,\; \Sigma_\lambda^{-1}(\phi(\varepsilon_t) - \mu) \rangle_{\mathcal{H}}
\end{equation*}
and the estimated nonconformity score is
\begin{equation*}
\hat{e}_t = \langle \phi(\hat{\varepsilon}_t) - \hat{\mu},\; \widehat{\Sigma}_\lambda^{-1}(\phi(\hat{\varepsilon}_t) - \hat{\mu}) \rangle_{\mathcal{H}}
\end{equation*}
where $\Sigma_\lambda = \Sigma + \lambda I_{\mathcal{H}}$ and $\widehat{\Sigma}_\lambda = \widehat{\Sigma} + \lambda I_{\mathcal{H}}$ are the regularized true and empirical cross-covariance operators. {Here $\mu$ and $\Sigma$ are the population mean embedding and covariance operator (fixed), while $\widehat{\mu}$ and $\widehat{\Sigma}$ are estimated from the calibration set.}
The estimated score $\hat{e}_t$ corresponds to $\hat{e}_k(\hat{\varepsilon}_t)$ in Proposition~\ref{prop:kmd_computable} after the Woodbury reduction (Appendix~\ref{proof:kernel_md}).

We also define the empirical CDFs of the true and estimated scores:
\begin{equation*}
\widetilde{F}_{T+1}(x) = \frac{1}{T}\sum_{t=1}^T \mathbbm{1}_{\{e_t \leq x\}}
\qquad
\widehat{F}_{T+1}(x) = \frac{1}{T}\sum_{t=1}^T \mathbbm{1}_{\{\hat{e}_t \leq x\}}.
\end{equation*}

\subsection{Assumptions}
\label{sec:assumptions}

\begin{assumption}[i.i.d.\ Residuals and Exogeneity]
\label{assumption_iid}
The residuals $\{\varepsilon_t\}_{t=1}^{T+1}$ are independent and identically distributed. Furthermore, the CDF $F_e$ of the true nonconformity score $e_t$ is Lipschitz continuous with constant $L > 0$. Finally, the test residual $\varepsilon_{T+1}$ is independent of the test input $X_{T+1}$.
\end{assumption}

\paragraph{Remark.} The Lipschitz condition on $F_e$ ensures that small perturbations of the scores lead to small changes in coverage probability. This assumption is relaxed in Section~\ref{sec:mixing}, where we extend the results to stationary $\alpha$-mixing sequences. The exogeneity condition $\varepsilon_{T+1} \perp X_{T+1}$ is standard in regression and ensures that the true nonconformity score $e_{T+1}$ is independent of $X_{T+1}$, which allows the marginal bound in the proof to hold conditionally on any fixed $X_{T+1} = x$. Under $\alpha$-mixing dependence (Section~\ref{sec:mixing}), $X_{T+1}$ typically contains lags of $Y_t$ and is correlated with $\varepsilon_{T+1}$, so this condition fails and only marginal coverage is established there.

\medskip
\begin{assumption}[Estimation Quality in RKHS]
\label{assumption_estimation_quality}
There exists a sequence $\{\delta_T\}_{T \geq 1}$ {satisfying $\delta_T = O(T^{-\rho})$ for some $\rho > 0$} such that:
\begin{equation*}
\frac{1}{T}\sum_{t=1}^{T} \|\phi(\hat{\varepsilon}_t) - \phi(\varepsilon_t)\|_{\mathcal{H}}^2 \leq \delta_T^2
\qquad \text{and} \qquad
\|\phi(\hat{\varepsilon}_{T+1}) - \phi(\varepsilon_{T+1})\|_{\mathcal{H}} \leq \delta_T.
\end{equation*}
\end{assumption}

\paragraph{Remark.} 
This requires the prediction error in the RKHS feature space to be small on average over the calibration set and pointwise at the test point. A sufficient condition is that the kernel satisfies
$\|\phi(x) - \phi(y)\|_{\mathcal{H}} \leq L_\phi \|x - y\|$
for some $L_\phi > 0$, in which case Assumption~\ref{assumption_estimation_quality} holds whenever
\begin{equation*}
\frac{1}{T}\sum_{t=1}^T \|\hat{\varepsilon}_t - \varepsilon_t\|^2 \leq (\delta_T / L_\phi)^2.
\end{equation*}
For the RBF kernel, $\|\phi(x) - \phi(y)\|_{\mathcal{H}}^2 = 2 - 2k(x,y) \leq \|x - y\|^2/\ell^2$, so the condition holds with $L_\phi = 1/\ell$.
{The polynomial rate $\rho$ at which the error vanishes depends on the estimation method and the problem structure. Faster estimators (larger $\rho$) yield tighter coverage guarantees.}


\medskip
\begin{assumption}[Boundedness]
\label{bounded_kernel}
The kernel $k$ is bounded: there exists $B > 0$ such that $k(x, x) \leq B^2$ for all $x$ in the domain.
\end{assumption}

\paragraph{Remark.} 
This is equivalent to $\|\phi(x)\|_{\mathcal{H}} \leq B$ for all $x$. It ensures that all moments of $\phi(\varepsilon)$ exist ($\mathbb{E}[\|\phi(\varepsilon)\|^q] \leq B^q$), which is required for the concentration inequalities used below.

\medskip
\begin{assumption}[Trace-Class Covariance]
\label{assumption:trace-class}
The population covariance operator $\Sigma = \mathbb{E}[(\phi(\varepsilon) - \mu) \otimes (\phi(\varepsilon) - \mu)]$ is trace-class:
$\mathrm{tr}(\Sigma) < \infty$.
\end{assumption}

\subsection{Coverage Guarantee and Convergence Rate (i.i.d.)}
\label{sec:iid_theory}

We prove the coverage guarantee through a sequence of lemmas, each bounding one source of error. These are combined in the proof of Theorem~\ref{thm:kernel_coverage}.

\subsubsection{Auxiliary Results}

We collect standard results used throughout the proofs.

\begin{lemma}[Cauchy--Schwarz for Self-Adjoint Operators]
\label{lemma:cauchy_schwarz_op}
For any self-adjoint operator $A$ on $\mathcal{H}$ and any $x \in \mathcal{H}$,
\begin{equation*}
|\langle x, Ax \rangle_{\mathcal{H}}| \leq \|x\|_{\mathcal{H}}^2 \, \|A\|_{\mathrm{op}}.\end{equation*}
\end{lemma}

\begin{lemma}[Resolvent Identity]
\label{lemma:resolvent}
For invertible operators $A$ and $B$ on $\mathcal{H}$,
\begin{equation*}
A^{-1} - (A + B)^{-1} = A^{-1} B (A + B)^{-1}.
\end{equation*}
\end{lemma}

\begin{proof}
We have
$A^{-1} = A^{-1}(A+B)(A+B)^{-1} = (I + A^{-1}B)(A+B)^{-1} = (A+B)^{-1} + A^{-1}B(A+B)^{-1}$.
Rearranging gives the result.
\end{proof}

\begin{lemma}[McDiarmid's Inequality]
\label{lemma:mcdiarmid}
Let $g: \mathcal{X}^m \to \mathbb{R}$ satisfy the bounded differences condition: for all $i \in \{1, \ldots, m\}$, there exists $c_i < \infty$ such that
\begin{equation*}
\sup_{x_1, \ldots, x_m, \tilde{x}_i} \bigl|g(x_1, \ldots, x_m) - g(x_1, \ldots, x_{i-1}, \tilde{x}_i, x_{i+1}, \ldots, x_m)\bigr| \leq c_i.
\end{equation*}
Then for independent random variables $X_1, \ldots, X_m$ and any $\xi > 0$,
\begin{equation*}
\mathbb{P}\bigl(g(X_1, \ldots, X_m) - \mathbb{E}[g(X_1, \ldots, X_m)] > \xi\bigr) \leq \exp\biggl(-\frac{2\xi^2}{\sum_{i=1}^m c_i^2}\biggr).
\end{equation*}
\end{lemma}

\subsubsection{Empirical CDF Concentration}

\begin{lemma}[Convergence of Empirical CDF under i.i.d.]
\label{lemma_convergence_empirical_cdf}
Under Assumption~\ref{assumption_iid}, for any $T \geq 1$ and $\delta \in (0,1)$, with probability at least $1 - \delta$,
\begin{equation}
\sup_{x \in \mathbb{R}} \bigl|\widetilde{F}_{T+1}(x) - F_e(x)\bigr| \leq \sqrt{\frac{\log(2/\delta)}{2T}}
.\end{equation}
\end{lemma}

\begin{proof}
Under Assumption~\ref{assumption_iid}, the true scores $e_1, \ldots, e_T$ are i.i.d.\ with CDF $F_e$. By the Dvoretzky--Kiefer--Wolfowitz (DKW) inequality, for any $\xi > 0$,
\begin{equation*}
\mathbb{P}\Bigl(\sup_x \bigl|\widetilde{F}_{T+1}(x) - F_e(x)\bigr| > \xi\Bigr) \leq 2\exp(-2T\xi^2)
.\end{equation*}
Setting the right-hand side equal to $\delta$ and solving: $\delta = 2\exp(-2T\xi^2)$ gives $\xi = \sqrt{\log(2/\delta)/(2T)}$.
\end{proof}

\subsubsection{Score Approximation Error}

\begin{lemma}[Difference Between Estimated and True Nonconformity Scores]
\label{lemma_difference_emp_true}
Under Assumptions~\ref{assumption_iid}, \ref{assumption_estimation_quality}, \ref{bounded_kernel}, and \ref{assumption:trace-class}, with probability at least $1 - 2\delta$,
\begin{align}
\frac{1}{T}\sum_{t=1}^T |e_t - \hat{e}_t|
&\leq
\frac{4B\delta_T}{\lambda}
+ \frac{1}{\lambda}\bigl(4B\|\Delta_\mu\| + \|\Delta_\mu\|^2\bigr) \nonumber \\
&\quad + \frac{4B^2}{\lambda^2}\Biggl[
C_K\|\Sigma\|_{\mathrm{op}}\biggl(
\sqrt{\frac{\mathbf{r}(\Sigma)}{T}} \bigvee \frac{\mathbf{r}(\Sigma)}{T}
\bigvee \sqrt{\frac{\log(1/\delta)}{T}} \bigvee \frac{\log(1/\delta)}{T}
\biggr) \nonumber \\
&\hspace{6em}
+ \biggl(\frac{2B}{\sqrt{T}} + B\sqrt{\frac{2\log(1/\delta)}{T}}\biggr)^{\!2}
+ \frac{8B^2}{T-1}
+ \frac{T}{T-1}\bigl(8B\delta_T + 4\delta_T^2\bigr)
\Biggr]
\label{eq:lemma2_bound}
,\end{align}
where $\Delta_\mu = \mu - \hat{\mu}$ is the total mean estimation error, satisfying with probability at least $1 - \delta$:
\begin{equation}
\|\Delta_\mu\| \leq \delta_T + \frac{2B}{\sqrt{T}} + B\sqrt{\frac{2\log(1/\delta)}{T}}
\label{eq:delta_mu_bound}
\end{equation}
and $C_K > 0$ is the constant from Koltchinskii et al.~\cite[Corollary~2]{koltchinskii2017concentration}.
\end{lemma}

\begin{proof}
We decompose the score difference via a telescoping argument. For each $t = 1, \ldots, T$, define the intermediate quantities:
\begin{align*}
I_1^{(t)} &= \langle \hat{v}_t,\; \Sigma_\lambda^{-1}\hat{v}_t \rangle_{\mathcal{H}} \\
I_2^{(t)} &= \langle \hat{w}_t,\; \Sigma_\lambda^{-1}\hat{w}_t \rangle_{\mathcal{H}}
\end{align*}
where we use the notation:
\begin{align*}
v_t &= \phi(\varepsilon_t) - \mu &\text{(true features, population-centered)}, \\
\hat{v}_t &= \phi(\hat{\varepsilon}_t) - \mu &\text{(estimated features, population-centered)}, \\
\hat{w}_t &= \phi(\hat{\varepsilon}_t) - \hat{\mu} &\text{(estimated features, sample-centered)}.
\end{align*}
Note that $e_t = \langle v_t, \Sigma_\lambda^{-1} v_t \rangle$ and $\hat{e}_t = \langle \hat{w}_t, \widehat{\Sigma}_\lambda^{-1} \hat{w}_t \rangle$, and that $\hat{w}_t = \hat{v}_t + \Delta_\mu$ where $\Delta_\mu = \mu - \hat{\mu}$.

By the triangle inequality:
\begin{equation*}
|e_t - \hat{e}_t| \leq
\underbrace{|e_t - I_1^{(t)}|}_{T_1^{(t)}: \text{ residual error}}
+ \underbrace{|I_1^{(t)} - I_2^{(t)}|}_{T_2^{(t)}: \text{ mean error}}
+ \underbrace{|I_2^{(t)} - \hat{e}_t|}_{T_3^{(t)}: \text{ covariance error}}
.\end{equation*}

We bound each term separately.

\paragraph{Bounding $T_1^{(t)}$ (residual estimation error)}

\begin{align*}
T_1^{(t)}
&= \bigl|\langle v_t, \Sigma_\lambda^{-1} v_t \rangle - \langle \hat{v}_t, \Sigma_\lambda^{-1} \hat{v}_t \rangle\bigr| \\
&= \bigl|\langle v_t - \hat{v}_t,\; \Sigma_\lambda^{-1} v_t \rangle + \langle \hat{v}_t,\; \Sigma_\lambda^{-1}(v_t - \hat{v}_t) \rangle\bigr| \\
&= \bigl|\langle v_t - \hat{v}_t,\; \Sigma_\lambda^{-1}(v_t + \hat{v}_t) \rangle\bigr|
\qquad \text{(self-adjointness of $\Sigma_\lambda^{-1}$)} \\
&\leq \|v_t - \hat{v}_t\| \cdot \|v_t + \hat{v}_t\| \cdot \|\Sigma_\lambda^{-1}\|_{\mathrm{op}}
\qquad \text{(Lemma~\ref{lemma:cauchy_schwarz_op})}.
\end{align*}

Since $\|\Sigma_\lambda^{-1}\|_{\mathrm{op}} \leq 1/\lambda$ and, by the triangle inequality and Assumption~\ref{bounded_kernel},
\begin{equation*}
\|v_t + \hat{v}_t\|
= \|\phi(\varepsilon_t) + \phi(\hat{\varepsilon}_t) - 2\mu\|
\leq \|\phi(\varepsilon_t)\| + \|\phi(\hat{\varepsilon}_t)\| + 2\|\mu\|
\leq 4B,
\end{equation*}
we obtain
\begin{equation*}
T_1^{(t)} \leq \frac{4B}{\lambda}\|\phi(\varepsilon_t) - \phi(\hat{\varepsilon}_t)\|.
\end{equation*}

Summing over $t$ and applying the Cauchy--Schwarz inequality:
\begin{align*}
\frac{1}{T}\sum_{t=1}^T T_1^{(t)}
&\leq \frac{4B}{\lambda} \cdot \frac{1}{T}\sum_{t=1}^T \|\phi(\varepsilon_t) - \phi(\hat{\varepsilon}_t)\| \\
&\leq \frac{4B}{\lambda} \sqrt{\frac{1}{T}\sum_{t=1}^T \|\phi(\varepsilon_t) - \phi(\hat{\varepsilon}_t)\|^2} \\
&\leq \frac{4B\delta_T}{\lambda},
\end{align*}
where the last step uses Assumption~\ref{assumption_estimation_quality}.

\paragraph{Bounding $T_2^{(t)}$ (mean estimation error).}

Since $\hat{w}_t = \hat{v}_t + \Delta_\mu$:
\begin{align*}
T_2^{(t)}
&= \bigl|\langle \hat{v}_t, \Sigma_\lambda^{-1}\hat{v}_t \rangle
- \langle \hat{v}_t + \Delta_\mu, \Sigma_\lambda^{-1}(\hat{v}_t + \Delta_\mu) \rangle\bigr| \\
&= \bigl|\langle \hat{v}_t, \Sigma_\lambda^{-1}\hat{v}_t \rangle
- \langle \hat{v}_t, \Sigma_\lambda^{-1}\hat{v}_t \rangle
- 2\langle \hat{v}_t, \Sigma_\lambda^{-1}\Delta_\mu \rangle
- \langle \Delta_\mu, \Sigma_\lambda^{-1}\Delta_\mu \rangle\bigr| \\
&= \bigl|2\langle \hat{v}_t, \Sigma_\lambda^{-1}\Delta_\mu \rangle
+ \langle \Delta_\mu, \Sigma_\lambda^{-1}\Delta_\mu \rangle\bigr| \\
&\leq 2\bigl|\langle \hat{v}_t, \Sigma_\lambda^{-1}\Delta_\mu \rangle\bigr|
+ \bigl|\langle \Delta_\mu, \Sigma_\lambda^{-1}\Delta_\mu \rangle\bigr|.
\end{align*}

By Cauchy--Schwarz and $\|\hat{v}_t\| = \|\phi(\hat{\varepsilon}_t) - \mu\| \leq 2B$:
\begin{align*}
2\bigl|\langle \hat{v}_t, \Sigma_\lambda^{-1}\Delta_\mu \rangle\bigr|
&\leq 2\|\hat{v}_t\| \cdot \|\Sigma_\lambda^{-1}\|_{\mathrm{op}} \cdot \|\Delta_\mu\|
\leq \frac{4B}{\lambda}\|\Delta_\mu\|, \\
\bigl|\langle \Delta_\mu, \Sigma_\lambda^{-1}\Delta_\mu \rangle\bigr|
&\leq \|\Delta_\mu\|^2 \cdot \|\Sigma_\lambda^{-1}\|_{\mathrm{op}}
\leq \frac{1}{\lambda}\|\Delta_\mu\|^2
\end{align*}

Since this bound is independent of $t$:
\begin{equation*}
\frac{1}{T}\sum_{t=1}^T T_2^{(t)} \leq \frac{1}{\lambda}\bigl(4B\|\Delta_\mu\| + \|\Delta_\mu\|^2\bigr).
\end{equation*}

We now bound $\|\Delta_\mu\| = \|\mu - \hat{\mu}\|$. Introducing the sample mean of true features $\tilde{\mu} = (1/T)\sum_{t=1}^T \phi(\varepsilon_t)$:
\begin{equation*}
\|\Delta_\mu\| = \|\mu - \hat{\mu}\| \leq \|\mu - \tilde{\mu}\| + \|\tilde{\mu} - \hat{\mu}\|.
\end{equation*}

The second term is bounded by Assumption~\ref{assumption_estimation_quality}:
\begin{align*}
\|\tilde{\mu} - \hat{\mu}\|
&= \Bigl\|\frac{1}{T}\sum_{t=1}^T \bigl(\phi(\varepsilon_t) - \phi(\hat{\varepsilon}_t)\bigr)\Bigr\| \\
&\leq \sqrt{\frac{1}{T}\sum_{t=1}^T \|\phi(\varepsilon_t) - \phi(\hat{\varepsilon}_t)\|^2}
\leq \delta_T.
\end{align*}

For $\|\mu - \tilde{\mu}\|$, we apply McDiarmid's inequality (Lemma~\ref{lemma:mcdiarmid}). Define
\begin{equation*}
g(\varepsilon_1, \ldots, \varepsilon_T) = \|\tilde{\mu} - \mu\| = \Bigl\|\frac{1}{T}\sum_{t=1}^T \phi(\varepsilon_t) - \mu\Bigr\|.
\end{equation*}
Replacing any single $\varepsilon_i$ changes $g$ by at most
\begin{equation*}
|g(\ldots, \varepsilon_i, \ldots) - g(\ldots, \tilde{\varepsilon}_i, \ldots)|
\leq \frac{1}{T}\|\phi(\varepsilon_i) - \phi(\tilde{\varepsilon}_i)\|
\leq \frac{2B}{T}
\end{equation*}
so the bounded differences condition holds with $c_i = 2B/T$ for all $i$. By Lemma~\ref{lemma:mcdiarmid}:
\begin{equation*}
\mathbb{P}\bigl(\|\tilde{\mu} - \mu\| - \mathbb{E}\|\tilde{\mu} - \mu\| > \xi\bigr) \leq \exp\biggl(-\frac{2\xi^2}{T \cdot (2B/T)^2}\biggr)
= \exp\biggl(-\frac{T\xi^2}{2B^2}\biggr).
\end{equation*}

We bound the expectation using the i.i.d.\ assumption:
\begin{align*}
\mathbb{E}\|\tilde{\mu} - \mu\|^2
&= \frac{1}{T^2}\sum_{t_1, t_2} \mathbb{E}\bigl[\langle \phi(\varepsilon_{t_1}) - \mu, \phi(\varepsilon_{t_2}) - \mu \rangle\bigr] \\
&= \frac{1}{T}\,\mathbb{E}\|\phi(\varepsilon) - \mu\|^2
\qquad \text{(cross terms vanish by independence)} \\
&\leq \frac{4B^2}{T}
\end{align*}
where the last step uses $\|\phi(\varepsilon) - \mu\|^2 \leq 2(\|\phi(\varepsilon)\|^2 + \|\mu\|^2) \leq 4B^2$. By Jensen's inequality:
\begin{equation*}
\mathbb{E}\|\tilde{\mu} - \mu\| \leq \sqrt{\mathbb{E}\|\tilde{\mu} - \mu\|^2} \leq \frac{2B}{\sqrt{T}}
.\end{equation*}

Setting $\delta = \exp(-T\xi^2/(2B^2))$ and solving for $\xi$: with probability at least $1 - \delta$,
\begin{equation*}
\|\tilde{\mu} - \mu\| \leq \frac{2B}{\sqrt{T}} + B\sqrt{\frac{2\log(1/\delta)}{T}}
.\end{equation*}

Combining with $\|\tilde{\mu} - \hat{\mu}\| \leq \delta_T$: with probability at least $1 - \delta$,
\begin{equation}
\|\Delta_\mu\| \leq \delta_T + \frac{2B}{\sqrt{T}} + B\sqrt{\frac{2\log(1/\delta)}{T}}
\label{eq:delta_mu_bound_proof}
.\end{equation}

\paragraph{Bounding $T_3^{(t)}$ (covariance estimation error).}

Recall $\hat{w}_t = \phi(\hat{\varepsilon}_t) - \hat{\mu}$, so $\|\hat{w}_t\| \leq 2B$. By Lemma~\ref{lemma:cauchy_schwarz_op}:
\begin{align*}
T_3^{(t)}
&= \bigl|\langle \hat{w}_t, \Sigma_\lambda^{-1} \hat{w}_t \rangle - \langle \hat{w}_t, \widehat{\Sigma}_\lambda^{-1} \hat{w}_t \rangle\bigr| \\
&= \bigl|\langle \hat{w}_t, (\Sigma_\lambda^{-1} - \widehat{\Sigma}_\lambda^{-1})\hat{w}_t \rangle\bigr| \\
&\leq \|\hat{w}_t\|^2 \cdot \|\Sigma_\lambda^{-1} - \widehat{\Sigma}_\lambda^{-1}\|_{\mathrm{op}} \\
&\leq 4B^2 \cdot \|\Sigma_\lambda^{-1} - \widehat{\Sigma}_\lambda^{-1}\|_{\mathrm{op}}
.\end{align*}

By the resolvent identity (Lemma~\ref{lemma:resolvent}) with $A = \Sigma + \lambda I$ and $B = \widehat{\Sigma} - \Sigma$:
\begin{align*}
\|\Sigma_\lambda^{-1} - \widehat{\Sigma}_\lambda^{-1}\|_{\mathrm{op}}
&= \|(\Sigma + \lambda I)^{-1}(\widehat{\Sigma} - \Sigma)(\widehat{\Sigma} + \lambda I)^{-1}\|_{\mathrm{op}} \\
&\leq \|(\Sigma + \lambda I)^{-1}\|_{\mathrm{op}} \cdot \|\widehat{\Sigma} - \Sigma\|_{\mathrm{op}} \cdot \|(\widehat{\Sigma} + \lambda I)^{-1}\|_{\mathrm{op}} \\
&\leq \frac{1}{\lambda^2}\|\widehat{\Sigma} - \Sigma\|_{\mathrm{op}}
.\end{align*}

We decompose $\|\widehat{\Sigma} - \Sigma\|_{\mathrm{op}} \leq \|\widetilde{\Sigma} - \Sigma\|_{\mathrm{op}} + \|\widehat{\Sigma} - \widetilde{\Sigma}\|_{\mathrm{op}}$ and bound each part.

\medskip
\emph{Statistical error $\|\widetilde{\Sigma} - \Sigma\|_{\mathrm{op}}$.}
Define $v_t = \phi(\varepsilon_t) - \mu$, $\Delta = \tilde{\mu} - \mu = (1/T)\sum_{t=1}^T v_t$, and $C_T = (1/T)\sum_{t=1}^T v_t \otimes v_t$.

Since $\widetilde{\Sigma} = \frac{1}{T-1}\sum_{t=1}^T (\phi(\varepsilon_t) - \tilde{\mu}) \otimes (\phi(\varepsilon_t) - \tilde{\mu})$ and $\phi(\varepsilon_t) - \tilde{\mu} = v_t - \Delta$, a direct computation gives:
\begin{align*}
\widetilde{\Sigma}
&= \frac{T}{T-1} \cdot \frac{1}{T}\sum_{t=1}^T (v_t - \Delta) \otimes (v_t - \Delta) \\
&= \frac{T}{T-1}\biggl[\frac{1}{T}\sum_{t=1}^T v_t \otimes v_t - \Delta \otimes \Delta\biggr]
= \frac{T}{T-1}\bigl[C_T - \Delta \otimes \Delta\bigr]
.\end{align*}

Therefore:
\begin{align*}
\widetilde{\Sigma} - \Sigma
&= \frac{T}{T-1}(C_T - \Delta \otimes \Delta) - \Sigma \\
&= (C_T - \Sigma) + \frac{1}{T-1}C_T - \frac{T}{T-1}(\Delta \otimes \Delta)
.\end{align*}

Taking operator norms and using $\frac{T}{T-1}\|\Delta\|^2 = \|\Delta\|^2 + \frac{1}{T-1}\|\Delta\|^2$:
\begin{equation*}
\|\widetilde{\Sigma} - \Sigma\|_{\mathrm{op}}
\leq \|C_T - \Sigma\|_{\mathrm{op}} + \|\Delta\|^2 + \frac{1}{T-1}\|C_T\|_{\mathrm{op}} + \frac{1}{T-1}\|\Delta\|^2
.\end{equation*}

Since $\|v_t\| \leq 2B$ (Assumption~\ref{bounded_kernel}):
\begin{equation*}
\|C_T\|_{\mathrm{op}} \leq \frac{1}{T}\sum_{t=1}^T \|v_t\|^2 \leq 4B^2
\qquad
\|\Delta\| \leq \frac{1}{T}\sum_{t=1}^T \|v_t\| \leq 2B
\end{equation*}
giving $\frac{1}{T-1}\|C_T\|_{\mathrm{op}} \leq \frac{4B^2}{T-1}$ and $\frac{1}{T-1}\|\Delta\|^2 \leq \frac{4B^2}{T-1}$.

For $\|C_T - \Sigma\|_{\mathrm{op}}$: since $\|v_t\| \leq 2B$, the random vectors $v_t$ are bounded, hence subgaussian. Combined with $\mathrm{tr}(\Sigma) < \infty$ (Assumption~\ref{assumption:trace-class}), the $v_t$ are pregaussian, satisfying the conditions of \cite[Corollary~2]{koltchinskii2017concentration}. For any $\delta \in (0,1)$, with probability at least $1 - \delta$:
\begin{equation}
\|C_T - \Sigma\|_{\mathrm{op}}
\leq C_K\|\Sigma\|_{\mathrm{op}}\biggl(
\sqrt{\frac{\mathbf{r}(\Sigma)}{T}} \bigvee \frac{\mathbf{r}(\Sigma)}{T}
\bigvee \sqrt{\frac{\log(1/\delta)}{T}} \bigvee \frac{\log(1/\delta)}{T}
\biggr)
\label{eq:koltchinskii_bound}
\end{equation}
where $\mathbf{r}(\Sigma) = \mathbb{E}\|v_t\|^2 / \|\Sigma\|_{\mathrm{op}} \leq 4B^2 / \|\Sigma\|_{\mathrm{op}}$.

Using the bound on $\|\Delta\|^2$ from the mean estimation analysis (which holds with probability at least $1 - \delta$, a union bound over the covariance and mean events gives: with probability at least $1 - 2\delta$,
\begin{equation}
\begin{aligned}
\|\widetilde{\Sigma} - \Sigma\|_{\mathrm{op}}
&\leq C_K\|\Sigma\|_{\mathrm{op}}\biggl(
\sqrt{\frac{\mathbf{r}(\Sigma)}{T}} \bigvee \frac{\mathbf{r}(\Sigma)}{T}
\bigvee \sqrt{\frac{\log(1/\delta)}{T}} \bigvee \frac{\log(1/\delta)}{T}\biggr) \\
&\quad + \biggl(\frac{2B}{\sqrt{T}} + B\sqrt{\frac{2\log(1/\delta)}{T}}\biggr)^{\!2}
+ \frac{8B^2}{T-1}.
\end{aligned}
\label{eq:sigma_tilde_bound}
\end{equation}

\medskip
\emph{Prediction error $\|\widehat{\Sigma} - \widetilde{\Sigma}\|_{\mathrm{op}}$.}
Define $\tilde{w}_t = \phi(\varepsilon_t) - \tilde{\mu}$ (true features, sample-centered), $\Delta_t = \phi(\hat{\varepsilon}_t) - \phi(\varepsilon_t)$ (a pointwise prediction error), and $\bar{\Delta} = (1/T)\sum_{t=1}^T \Delta_t = \hat{\mu} - \tilde{\mu}$ (the mean prediction error).

Then $\hat{w}_t = \phi(\hat{\varepsilon}_t) - \hat{\mu} = (\phi(\varepsilon_t) + \Delta_t) - (\tilde{\mu} + \bar{\Delta}) = \tilde{w}_t + \theta_t$, where $\theta_t = \Delta_t - \bar{\Delta}$ is the demeaned prediction error. Expanding:
\begin{align*}
\|\widehat{\Sigma} - \widetilde{\Sigma}\|_{\mathrm{op}}
&= \Bigl\|\frac{1}{T-1}\sum_{t=1}^T \bigl(\hat{w}_t \otimes \hat{w}_t - \tilde{w}_t \otimes \tilde{w}_t\bigr)\Bigr\|_{\mathrm{op}} \\
&= \Bigl\|\frac{1}{T-1}\sum_{t=1}^T \bigl(\tilde{w}_t \otimes \theta_t + \theta_t \otimes \tilde{w}_t + \theta_t \otimes \theta_t\bigr)\Bigr\|_{\mathrm{op}} \\
&\leq \frac{1}{T-1}\sum_{t=1}^T \bigl(2\|\tilde{w}_t\|\,\|\theta_t\| + \|\theta_t\|^2\bigr)
.\end{align*}

We have $\|\tilde{w}_t\| = \|\phi(\varepsilon_t) - \tilde{\mu}\| \leq 2B$ and $\|\theta_t\| = \|\Delta_t - \bar{\Delta}\| \leq \|\Delta_t\| + \|\bar{\Delta}\|$. By Assumption~\ref{assumption_estimation_quality}:
\begin{equation*}
\|\bar{\Delta}\| = \Bigl\|\frac{1}{T}\sum_{t=1}^T \Delta_t\Bigr\| \leq \sqrt{\frac{1}{T}\sum_{t=1}^T \|\Delta_t\|^2} \leq \delta_T
.\end{equation*}

Using $(1/T)\sum_t \|\Delta_t\| \leq \delta_T$ and $(1/T)\sum_t \|\Delta_t\|^2 \leq \delta_T^2$:
\begin{align*}
\|\widehat{\Sigma} - \widetilde{\Sigma}\|_{\mathrm{op}}
&\leq \frac{T}{T-1} \cdot \frac{1}{T}\sum_{t=1}^T \bigl(4B(\|\Delta_t\| + \delta_T) + (\|\Delta_t\| + \delta_T)^2\bigr) \\
&\leq \frac{T}{T-1}\bigl(8B\delta_T + 4\delta_T^2\bigr)
.\end{align*}

\medskip
\emph{Combining.} Summing and dividing by $T$:
\begin{equation*}
\frac{1}{T}\sum_{t=1}^T T_3^{(t)}
\leq \frac{4B^2}{\lambda^2}\Bigl[\|\widetilde{\Sigma} - \Sigma\|_{\mathrm{op}} + \|\widehat{\Sigma} - \widetilde{\Sigma}\|_{\mathrm{op}}\Bigr]
.\end{equation*}

The bounds on $\|\widetilde{\Sigma} - \Sigma\|_{\mathrm{op}}$ (Equation~\eqref{eq:sigma_tilde_bound}) and $\|\widehat{\Sigma} - \widetilde{\Sigma}\|_{\mathrm{op}}$ hold simultaneously with probability at least $1 - 2\delta$.
Adding $T_1 + T_2 + T_3$ gives~\eqref{eq:lemma2_bound}.
\end{proof}

\subsubsection{ECDF Coupling}

\begin{lemma}[Distance Between Empirical CDFs of True and Estimated Scores]
\label{lemma:distance_ecdfs}
Under Assumptions~\ref{assumption_iid}, \ref{assumption_estimation_quality}, \ref{bounded_kernel}, and \ref{assumption:trace-class}, with probability at least $1 - 2\delta$,
\begin{equation}
\sup_{x \in \mathbb{R}} \bigl|\widehat{F}_{T+1}(x) - \widetilde{F}_{T+1}(x)\bigr|
\leq (L+1)\,C_S + 2\sup_{x \in \mathbb{R}} \bigl|\widetilde{F}_{T+1}(x) - F_e(x)\bigr|
\end{equation}
where $C_S = \bigl[(1/T)\sum_{t=1}^T |e_t - \hat{e}_t|\bigr]^{1/2}$ and $L$ is the Lipschitz constant of $F_e$.
\end{lemma}

\begin{proof}
The argument follows \cite[Lemma~4.11]{xu2024conformal}. For any threshold $x$:
\begin{equation*}
\bigl|\widehat{F}_{T+1}(x) - \widetilde{F}_{T+1}(x)\bigr|
= \frac{1}{T}\Bigl|\sum_{t=1}^T \bigl(\mathbbm{1}_{\{\hat{e}_t \leq x\}} - \mathbbm{1}_{\{e_t \leq x\}}\bigr)\Bigr|
\leq \frac{1}{T}\sum_{t=1}^T \mathbbm{1}_{\{|e_t - x| \leq |e_t - \hat{e}_t|\}}
.\end{equation*}

For any $\eta > 0$, we split each indicator:
\begin{equation*}
\mathbbm{1}_{\{|e_t - x| \leq |e_t - \hat{e}_t|\}}
\leq \mathbbm{1}_{\{|e_t - x| \leq \eta\}} + \mathbbm{1}_{\{|e_t - \hat{e}_t| > \eta\}}
.\end{equation*}

For the second term, by Markov's inequality:
\begin{equation*}
\frac{1}{T}\sum_{t=1}^T \mathbbm{1}_{\{|e_t - \hat{e}_t| > \eta\}}
\leq \frac{1}{\eta} \cdot \frac{1}{T}\sum_{t=1}^T |e_t - \hat{e}_t|
= \frac{C_S^2}{\eta}.
\end{equation*}

For the first term, using the Lipschitz property of $F_e$:
\begin{align*}
\frac{1}{T}\sum_{t=1}^T \mathbbm{1}_{\{|e_t - x| \leq \eta\}}
&\leq \widetilde{F}_{T+1}(x + \eta) - \widetilde{F}_{T+1}(x - \eta) \\
&\leq F_e(x + \eta) - F_e(x - \eta) + 2\sup_x |\widetilde{F}_{T+1}(x) - F_e(x)| \\
&\leq 2L\eta + 2\sup_x |\widetilde{F}_{T+1}(x) - F_e(x)|.
\end{align*}

Combining, $|\widehat{F}_{T+1}(x) - \widetilde{F}_{T+1}(x)| \leq 2L\eta + C_S^2/\eta + 2\sup_x |\widetilde{F}_{T+1}(x) - F_e(x)|$. Treating the two signs of $\widehat{F}_{T+1}(x) - \widetilde{F}_{T+1}(x)$ separately replaces $2L\eta$ by $L\eta$, and $\eta = C_S$ gives the stated bound \cite[Lemma~4.11]{xu2024conformal}.
\end{proof}

\subsubsection{Test Point Score Error}

\begin{lemma}[Test Point Nonconformity Score Error]
\label{lemma:test_point_error}
Under Assumptions~\ref{assumption_iid}, \ref{assumption_estimation_quality}, \ref{bounded_kernel}, and \ref{assumption:trace-class}, with probability at least $1 - 2\delta$,
\begin{equation}
|e_{T+1} - \hat{e}_{T+1}| \leq \zeta_T
\end{equation}
where $\zeta_T$ has the same form as the right-hand side of~\eqref{eq:lemma2_bound}.
\end{lemma}

\begin{proof}
By the same telescoping decomposition:
\begin{equation*}
|e_{T+1} - \hat{e}_{T+1}| \leq T_1^{(T+1)} + T_2^{(T+1)} + T_3^{(T+1)}
.\end{equation*}

The bounds on $T_2^{(T+1)}$ and $T_3^{(T+1)}$ are identical to those in Lemma~\ref{lemma_difference_emp_true}, since they depend only on $\hat{\mu}$, $\widehat{\Sigma}$, and $\Sigma$ (shared between calibration and test point). For $T_1^{(T+1)}$, the same derivation gives:
\begin{equation*}
T_1^{(T+1)} \leq \frac{4B}{\lambda}\|\phi(\varepsilon_{T+1}) - \phi(\hat{\varepsilon}_{T+1})\| \leq \frac{4B\delta_T}{\lambda}
\end{equation*}
using the test-point bound in Assumption~\ref{assumption_estimation_quality}. Summing all three terms gives $|e_{T+1} - \hat{e}_{T+1}| \leq \zeta_T$, where $\zeta_T$ has the same expression as $(1/T)\sum_t |e_t - \hat{e}_t| = C_S^2$ from Lemma~\ref{lemma_difference_emp_true}.
\end{proof}

\subsubsection{Coverage Guarantee}

\begin{theorem}[Asymptotic Conditional Coverage Guarantee under i.i.d.]
\label{thm:kernel_coverage}
Under Assumptions~\ref{assumption_iid}, \ref{assumption_estimation_quality}, \ref{bounded_kernel}, and \ref{assumption:trace-class}, for any $T \geq 1$, $\alpha \in (0,1)$, and $\delta \in (0,1)$, with probability at least $1 - 4\delta$:
\begin{equation*}
\bigl|\mathbb{P}(Y_{T+1} \in \widehat{C}_{T+1}^\alpha \mid X_{T+1}) - (1-\alpha)\bigr|
\leq 12\sqrt{\frac{\log(2/\delta)}{2T}} + 4(L+1)(C_S + \zeta_T) + 2\delta
.\end{equation*}
\end{theorem}

\begin{proof}
Under Assumption~\ref{assumption_iid}, $\varepsilon_{T+1} \perp X_{T+1}$, so $e_{T+1}$ is a function of $\varepsilon_{T+1}$ alone and satisfies $F_e(e_{T+1}) \mid X_{T+1} \sim \mathrm{Uniform}(0,1)$. Furthermore, conditioned on $X_{T+1} = x$, the estimated score $\hat{e}_{T+1} = \hat{e}_k(f(x) - \hat{f}(x) + \varepsilon_{T+1})$ depends on $x$ only through the deterministic shift $f(x) - \hat{f}(x)$, with all remaining randomness from $\varepsilon_{T+1} \perp x$. All subsequent bounds therefore hold conditionally on any fixed $X_{T+1} = x$, and the $\mid X_{T+1}$ notation is justified throughout this proof.

For any $\beta \in [0, \alpha]$, by construction of $\widehat{C}_{T+1}^\alpha$:
\begin{equation*}
Y_{T+1} \in \widehat{C}_{T+1}^\alpha
\iff \beta \leq \widehat{F}_{T+1}(\hat{e}_{T+1}) \leq 1 - \alpha + \beta
\end{equation*}

Since $F_e(e_{T+1}) \sim \mathrm{Uniform}(0,1)$ under Assumption~\ref{assumption_iid}, by the elementary bound $|\mathbbm{1}\{a \leq x \leq b\} - \mathbbm{1}\{a \leq y \leq b\}| \leq |\mathbbm{1}\{a \leq x\} - \mathbbm{1}\{a \leq y\}| + |\mathbbm{1}\{x \leq b\} - \mathbbm{1}\{y \leq b\}|$:
\begin{align*}
&\bigl|\mathbb{P}(Y_{T+1} \in \widehat{C}_{T+1}^\alpha \mid X_{T+1}) - (1-\alpha)\bigr| \\
&\leq \mathbb{P}\bigl(|F_e(e_{T+1}) - \beta| \leq |\widehat{F}_{T+1}(\hat{e}_{T+1}) - F_e(e_{T+1})|\bigr) \\
&\quad + \mathbb{P}\bigl(|F_e(e_{T+1}) - (1-\alpha+\beta)| \leq |\widehat{F}_{T+1}(\hat{e}_{T+1}) - F_e(e_{T+1})|\bigr)
\end{align*}

Let $A_T$ denote the event in Lemma~\ref{lemma_convergence_empirical_cdf}: $\sup_x |\widetilde{F}_{T+1}(x) - F_e(x)| \leq \sqrt{\log(2/\delta)/(2T)}$, with $\mathbb{P}(A_T) \geq 1 - \delta$. For any $\gamma \in [0,1]$:
\begin{align*}
&\mathbb{P}\bigl(|F_e(e_{T+1}) - \gamma| \leq |\widehat{F}_{T+1}(\hat{e}_{T+1}) - F_e(e_{T+1})|\bigr) \\
&\leq \mathbb{P}\bigl(|F_e(e_{T+1}) - \gamma| \leq |\widehat{F}_{T+1}(\hat{e}_{T+1}) - F_e(\hat{e}_{T+1})|
+ |F_e(\hat{e}_{T+1}) - F_e(e_{T+1})| \;\big|\; A_T\bigr) + \delta
\end{align*}

Conditioning on $A_T$, we bound $|\widehat{F}_{T+1}(\hat{e}_{T+1}) - F_e(\hat{e}_{T+1})|$ by the triangle inequality:
\begin{align*}
|\widehat{F}_{T+1}(\hat{e}_{T+1}) - F_e(\hat{e}_{T+1})|
&\leq \sup_x |\widehat{F}_{T+1}(x) - \widetilde{F}_{T+1}(x)|
+ \sup_x |\widetilde{F}_{T+1}(x) - F_e(x)|
\end{align*}

By Lemma~\ref{lemma:distance_ecdfs} and the definition of $A_T$:
\begin{align*}
\sup_x |\widehat{F}_{T+1}(x) - \widetilde{F}_{T+1}(x)|
&\leq (L+1)C_S + 2\sqrt{\frac{\log(2/\delta)}{2T}}
\end{align*}

Adding $\sup_x |\widetilde{F}_{T+1}(x) - F_e(x)| \leq \sqrt{\log(2/\delta)/(2T)}$ on $A_T$:
\begin{equation*}
|\widehat{F}_{T+1}(\hat{e}_{T+1}) - F_e(\hat{e}_{T+1})|
\leq (L+1)C_S + 3\sqrt{\frac{\log(2/\delta)}{2T}}
\end{equation*}

By the Lipschitz continuity of $F_e$ and Lemma~\ref{lemma:test_point_error}:
\begin{equation*}
|F_e(\hat{e}_{T+1}) - F_e(e_{T+1})| \leq L|\hat{e}_{T+1} - e_{T+1}| \leq L\zeta_T
\end{equation*}

Combining:
\begin{align*}
|\widehat{F}_{T+1}(\hat{e}_{T+1}) - F_e(e_{T+1})|
&\leq (L+1)C_S + 3\sqrt{\frac{\log(2/\delta)}{2T}} + L\zeta_T \\
&\leq 3\sqrt{\frac{\log(2/\delta)}{2T}} + (L+1)(C_S + \zeta_T)
\end{align*}

Since $F_e(e_{T+1}) \sim \mathrm{Uniform}(0,1)$, $\mathbb{P}(|F_e(e_{T+1}) - \gamma| \leq \eta \mid A_T) \leq 2\eta$. Therefore:
\begin{align*}
&\mathbb{P}\bigl(|F_e(e_{T+1}) - \gamma| \leq |\widehat{F}_{T+1}(\hat{e}_{T+1}) - F_e(e_{T+1})| \;\big|\; A_T\bigr) \\
&\leq 6\sqrt{\frac{\log(2/\delta)}{2T}} + 2(L+1)(C_S + \zeta_T)
\end{align*}

Adding $\mathbb{P}(A_T^c) \leq \delta$ and applying with $\gamma = \beta$ and $\gamma = 1-\alpha+\beta$:
\begin{align*}
&\bigl|\mathbb{P}(Y_{T+1} \in \widehat{C}_{T+1}^\alpha \mid X_{T+1}) - (1-\alpha)\bigr| \\
&\leq 2\Bigl[6\sqrt{\frac{\log(2/\delta)}{2T}} + 2(L+1)(C_S + \zeta_T) + \delta\Bigr] \\
&= 12\sqrt{\frac{\log(2/\delta)}{2T}} + 4(L+1)(C_S + \zeta_T) + 2\delta
\end{align*}

The probability is at least $1 - 4\delta$ by a union bound over: DKW ($1 - \delta$), mean concentration ($1 - \delta$), covariance concentration ($1 - \delta$), and test-point estimation ($1 - \delta$).
\end{proof}

\subsubsection{Convergence Rate}

\begin{theorem}[Conditional Coverage Convergence Rate under i.i.d.]
\label{thm:convergence_iid}
Under Assumptions~\ref{assumption_iid}--\ref{assumption:trace-class}, there exists a constant $C > 0$ depending only on $B$, $C_K$, $\|\Sigma\|_{\mathrm{op}}$, $L$ such that with probability at least $1 - 4T^{-1}$:
\begin{equation*}
\bigl|\mathbb{P}(Y_{T+1} \in \widehat{C}_{T+1}^\alpha \mid X_{T+1}) - (1-\alpha)\bigr|
\leq
\begin{cases}
C \cdot T^{\beta - 1/4}(\log T)^{1/4}
& \text{if } \rho \geq 1/2, \; \lambda = T^{-\beta}, \; \beta \in (0, 1/4) \\[4pt]
C \cdot T^{\beta - \rho/2}
& \text{if } \rho < 1/2, \; \lambda = T^{-\beta}, \; \beta \in (0, \rho/2)
\end{cases}
\end{equation*}
\end{theorem}

\begin{proof}
Set $\delta = T^{-1}$ (so $\log(1/\delta) = \log T$) and $\lambda = T^{-\beta}$. We determine the asymptotic rate of each term in Lemma~\ref{lemma_difference_emp_true}.

\paragraph{Residual error term.}
\begin{equation}
\frac{4B\delta_T}{\lambda} = O(T^{\beta - \rho})
\label{eq:t1_rate}
\end{equation}

\paragraph{Mean error term.}
From~\eqref{eq:delta_mu_bound_proof}:
\begin{equation*}
\|\Delta_\mu\| \leq \delta_T + \frac{2B}{\sqrt{T}} + B\sqrt{\frac{2\log T}{T}}
= O\bigl(\delta_T + T^{-1/2}\sqrt{\log T}\bigr)
\end{equation*}

If $\rho \geq 1/2$, the $T^{-1/2}\sqrt{\log T}$ term dominates $\delta_T = O(T^{-\rho})$, so $\|\Delta_\mu\| = O(T^{-1/2}\sqrt{\log T})$. Since $\|\Delta_\mu\|^2 = O(T^{-1}\log T)$ is dominated by $\|\Delta_\mu\|$ for large $T$:
\begin{equation*}
\frac{1}{\lambda}\bigl(4B\|\Delta_\mu\| + \|\Delta_\mu\|^2\bigr) = O\bigl(T^{\beta - 1/2}\sqrt{\log T}\bigr)
\end{equation*}

If $\rho < 1/2$, then $\delta_T$ dominates and:
\begin{equation*}
\frac{1}{\lambda}\bigl(4B\|\Delta_\mu\| + \|\Delta_\mu\|^2\bigr) = O(T^{\beta - \rho})
\end{equation*}

In both cases, this term vanishes for $\beta < \min(\rho, 1/2)$. We write:
\begin{equation}
\frac{1}{\lambda}\bigl(4B\|\Delta_\mu\| + \|\Delta_\mu\|^2\bigr) = O\bigl(T^{\beta - \min(\rho, 1/2)}\sqrt{\log T}\bigr)
\label{eq:t2_rate}
\end{equation}

\paragraph{Covariance error term.}
Define the effective rank bound $r_0 = 4B^2/\|\Sigma\|_{\mathrm{op}} \geq \mathbf{r}(\Sigma)$. For $T \geq e^{r_0}$, we have $\log T \geq r_0 \geq \mathbf{r}(\Sigma)$, so:
\begin{equation*}
\sqrt{\frac{\mathbf{r}(\Sigma)}{T}} \leq \sqrt{\frac{\log T}{T}},
\qquad
\frac{\mathbf{r}(\Sigma)}{T} \leq \frac{\log T}{T}
\end{equation*}

Since $\sqrt{\log T / T}$ dominates $\log T / T$ for all $T$, the Koltchinskii bound~\eqref{eq:koltchinskii_bound} simplifies to:
\begin{equation}
\|C_T - \Sigma\|_{\mathrm{op}} = O\biggl(\sqrt{\frac{\log T}{T}}\biggr)
\label{eq:t3_kolt_rate}
\end{equation}

The remaining terms in~\eqref{eq:sigma_tilde_bound}:
\begin{equation}
\biggl(\frac{2B}{\sqrt{T}} + B\sqrt{\frac{2\log T}{T}}\biggr)^{\!2}
= O\biggl(\frac{\log T}{T}\biggr)
\label{eq:t3_mean_sq_rate}
\end{equation}
\begin{equation}
\frac{8B^2}{T-1} = O(T^{-1}),
\qquad
\frac{T}{T-1}\bigl(8B\delta_T + 4\delta_T^2\bigr) = O(\delta_T) = O(T^{-\rho})
\label{eq:t3_remaining_rate}
.\end{equation}

Since $O(\sqrt{\log T/T})$ dominates both $O(\log T/T)$ and $O(T^{-1})$, and $O(T^{-\rho})$ dominates $O(T^{-1})$ for $\rho < 1$, we denote the full covariance error bracket by $A$:
\begin{equation*}
A = \|\widetilde{\Sigma} - \Sigma\|_{\mathrm{op}} + \|\widehat{\Sigma} - \widetilde{\Sigma}\|_{\mathrm{op}}
= O\biggl(\sqrt{\frac{\log T}{T}}\biggr) + O(T^{-\rho})
.\end{equation*}

Multiplying by $4B^2/\lambda^2 = O(T^{2\beta})$:
\begin{equation}
\frac{4B^2}{\lambda^2} \cdot A = O(T^{2\beta}) \cdot \biggl[O\biggl(T^{-1/2}\sqrt{\log T}\biggr) + O(T^{-\rho})\biggr].
\label{eq:t3_combined_rate}
\end{equation}

For $\rho \geq 1/2$, the first term dominates:
\begin{equation*}
\frac{4B^2}{\lambda^2} \cdot A = O(T^{2\beta - 1/2}\sqrt{\log T})
\end{equation*}
which vanishes for $\beta < 1/4$. For $\rho < 1/2$, the second term dominates:
\begin{equation*}
\frac{4B^2}{\lambda^2} \cdot A = O(T^{2\beta - \rho})
\end{equation*}
which vanishes for $\beta < \rho/2$.

\paragraph{Dominant term.}
Comparing the exponents from~\eqref{eq:t1_rate}, \eqref{eq:t2_rate}, and~\eqref{eq:t3_combined_rate}:
\begin{equation*}
\beta - \rho \;\leq\; \beta - \min(\rho, 1/2) \;<\; 2\beta - \min(\rho, 1/2)
\end{equation*}
so the covariance term~\eqref{eq:t3_combined_rate} dominates. Thus:
\begin{equation}
C_S^2 = \zeta_T =
\begin{cases}
O(T^{2\beta - 1/2}\sqrt{\log T}) & \text{if } \rho \geq 1/2, \\
O(T^{2\beta - \rho}) & \text{if } \rho < 1/2
\end{cases}
\label{eq:cs_squared_rate}
\end{equation}
where we use that $C_S^2$ is the calibration average bound from Lemma~\ref{lemma_difference_emp_true} and $\zeta_T$ is the test-point bound from Lemma~\ref{lemma:test_point_error}.

Taking square roots: $C_S = O(T^{\beta - 1/4}(\log T)^{1/4})$ for $\rho \geq 1/2$, and $C_S = O(T^{\beta - \rho/2})$ for $\rho < 1/2$.

\paragraph{Final bound.}
From Theorem~\ref{thm:kernel_coverage} with $\delta = T^{-1}$, with probability at least $1 - 4T^{-1}$:
\begin{equation*}
\bigl|\mathbb{P}(Y_{T+1} \in \widehat{C}_{T+1}^\alpha \mid X_{T+1}) - (1-\alpha)\bigr|
\leq \underbrace{12\sqrt{\frac{\log(2T)}{2T}}}_{O(T^{-1/2}\sqrt{\log T})}
+ \underbrace{4(L+1)(C_S + \zeta_T)}_{\text{score error}}
+ \underbrace{2T^{-1}}_{O(T^{-1})}.
\end{equation*}

Since $\zeta_T = C_S^2$ and $C_S \to 0$ (for $\beta < \min(\rho/2, 1/4)$), we have $C_S + \zeta_T \leq 2C_S$ for large $T$. The $C_S$ term dominates the DKW and $T^{-1}$ terms because $\beta - \min(\rho/2, 1/4) > -1/2 > -1$ for $\beta, \rho > 0$, giving:
\begin{equation*}
\bigl|\mathbb{P}(Y_{T+1} \in \widehat{C}_{T+1}^\alpha \mid X_{T+1}) - (1-\alpha)\bigr|
\leq
\begin{cases}
C \cdot T^{\beta - 1/4}(\log T)^{1/4}
& \text{if } \rho \geq 1/2 \\[4pt]
C \cdot T^{\beta - \rho/2}
& \text{if } \rho < 1/2
\end{cases}
\end{equation*}
where $C$ depends on $B$, $C_K$, $\|\Sigma\|_{\mathrm{op}}$, $L$.
\end{proof}

{\paragraph{Remark.} The constant $C$ depends on $B$, $C_K$, $\|\Sigma\|_{\mathrm{op}}$ and $L$, and its dependence on the kernel can be characterised for common choices. Any stationary kernel with $k(0) = 1$, such as the RBF, Mat\'ern and Laplace kernels, has $B = 1$. Since $\Sigma$ is built from centered features, $\mathrm{tr}(\Sigma) = 1 - \|\mu\|_{\mathcal{H}}^2 \le 1$ in every dimension. The spectral factor of the covariance concentration bound then satisfies $\|\Sigma\|_{\mathrm{op}}\sqrt{\mathbf{r}(\Sigma)} = \sqrt{\mathrm{tr}(\Sigma)\,\|\Sigma\|_{\mathrm{op}}} \le 1$. For these kernels, the part of the constant that depends on the kernel is therefore bounded by universal numbers that do not depend on $d$. The only remaining quantity is the Lipschitz constant $L$, which is a property of the residual distribution. For comparison, the linear kernel on standardized and bounded residuals gives $\mathrm{tr}(\Sigma) = d$, which recovers the dimension dependence of the Mahalanobis score. The kernel also enters the exponent through $\rho$ in Assumption~\ref{assumption_estimation_quality}, since $\delta_T$ measures the error of the base model in the RKHS norm of the kernel.}

\subsection{Coverage under $\alpha$-Mixing Dependence}
\label{sec:mixing}

We extend the coverage analysis to sequences with temporal dependence. We assume that the mean and covariance operator are known (Assumption~\ref{assumption:known_mean_cov}), eliminating the mean and covariance estimation error and isolating the effect of mixing on conformal calibration.

\subsubsection{Additional Assumptions}

\begin{assumption}[Stationary Strong Mixing]
\label{assumption:stationary_mixing}
The process $\{\varepsilon_t\}_{t \geq 1}$ is stationary and $\alpha$-mixing with mixing coefficients $\{\alpha_k\}_{k \geq 0}$, satisfying $\sum_{k=0}^\infty \alpha_k \leq M < \infty$. Furthermore, the CDF $F_e$ of the true nonconformity score is $L$-Lipschitz.
\end{assumption}

\begin{assumption}[Known Mean and Covariance]
\label{assumption:known_mean_cov}
The population mean $\mu = \mathbb{E}[\phi(\varepsilon)]$ and covariance operator $\Sigma = \mathbb{E}[(\phi(\varepsilon) - \mu) \otimes (\phi(\varepsilon) - \mu)]$ are known.
\end{assumption}

\paragraph{Remark.} 
Under Assumption~\ref{assumption:known_mean_cov}, we set $\hat{\mu} = \mu$ and $\widehat{\Sigma} = \Sigma$ in the construction of the nonconformity scores for analysis. This eliminates the mean estimation error (term $T_2$) and the covariance estimation error (term $T_3$) from the decomposition in Lemma~\ref{lemma_difference_emp_true}, leaving only the residual estimation error (term $T_1$).

This assumption isolates the effect of temporal dependence on conformal calibration. Analyzing covariance operator estimation under $\alpha$-mixing in infinite-dimensional RKHSs would require concentration inequalities considerably more involved than their i.i.d.\ counterparts and is beyond the scope of this work. In practice, $\Sigma$ can be estimated from an independent sample or from a sufficiently long stationary pre-training period.

\subsubsection{Inheritance of Mixing}

\begin{lemma}[Inheritance of Mixing by Nonconformity Scores]
\label{lemma:score_mixing}
Let $\{\varepsilon_t\}_{t \geq 1}$ be a stationary $\alpha$-mixing sequence with mixing coefficients $\{\alpha_k\}_{k \geq 0}$. Define
\begin{equation*}
e_t = \langle \phi(\varepsilon_t) - \mu,\; \Sigma_\lambda^{-1}(\phi(\varepsilon_t) - \mu) \rangle_{\mathcal{H}}.
\end{equation*}
Then $\{e_t\}_{t \geq 1}$ is stationary and $\alpha$-mixing with $\alpha_k(\{e_t\}) \leq \alpha_k(\{\varepsilon_t\})$ for all $k \geq 0$.
\end{lemma}

\begin{proof}
The map $\varepsilon \mapsto e(\varepsilon) = \langle \phi(\varepsilon) - \mu, \Sigma_\lambda^{-1}(\phi(\varepsilon) - \mu) \rangle_{\mathcal{H}}$ is measurable as a composition of continuous functions: $\phi$ is continuous (for RBF, Mat\'ern, polynomial kernels) and $v \mapsto \langle v, \Sigma_\lambda^{-1}v \rangle$ is continuous on $\mathcal{H}$ since $\|\Sigma_\lambda^{-1}\|_{\mathrm{op}} \leq 1/\lambda < \infty$. Therefore:
\begin{equation*}
\sigma(e_t : t \in I) \subseteq \sigma(\varepsilon_t : t \in I)
\end{equation*}
for any index set $I \subseteq \mathbb{Z}$. Since $\alpha$-mixing coefficients are defined as suprema over sub-$\sigma$-algebras, the containment gives $\alpha_k(\{e_t\}) \leq \alpha_k(\{\varepsilon_t\})$ for all $k \geq 0$.
\end{proof}

\subsubsection{ECDF Concentration under Mixing}

\begin{lemma}[Empirical CDF Concentration under Mixing]
\label{lemma:ecdf_mixing}
Under Assumption~\ref{assumption:stationary_mixing}, for the true nonconformity scores $\{e_t\}_{t=1}^T$, with probability at least $1 - (M(\log T)^2/(2T))^{1/3}$:
\begin{equation}
\sup_{x \in \mathbb{R}} \bigl|\widetilde{F}_{T+1}(x) - F_e(x)\bigr|
\leq \frac{(M/2)^{1/3}(\log T)^{2/3}}{T^{1/3}}.
\end{equation}
\end{lemma}

\begin{proof}
By Lemma~\ref{lemma:score_mixing}, the scores $\{e_t\}_{t=1}^T$ are stationary and $\alpha$-mixing with $\sum_{k=0}^\infty \alpha_k \leq M$. The result follows from Rio's inequality \cite[Proposition~7.1]{rio2017asymptotic} applied to the indicator processes $\{\mathbbm{1}_{\{e_t \leq x\}}\}_{t=1}^T$, following the argument of Xu et al.~\cite[Lemma~B.11]{xu2024conformal}.
\end{proof}

\subsubsection{Test Point Error under Known Covariance}

\begin{lemma}[Test Point Nonconformity Score Error under Known Mean and Covariance]
\label{lemma:test_point_error_mixing}
Under Assumptions~\ref{assumption:stationary_mixing}, \ref{assumption:known_mean_cov}, \ref{assumption_estimation_quality}, and \ref{bounded_kernel}:
\begin{equation}
|e_{T+1} - \hat{e}_{T+1}| \leq \tilde{\zeta}_T = \frac{4B\delta_T}{\lambda}.
\end{equation}
\end{lemma}

\begin{proof}
Under Assumption~\ref{assumption:known_mean_cov}, $\hat{\mu} = \mu$ and $\widehat{\Sigma} = \Sigma$. In the telescoping decomposition, $T_2^{(T+1)} = 0$ (no mean error) and $T_3^{(T+1)} = 0$ (no covariance error). Only the residual error remains:
\begin{equation*}
|e_{T+1} - \hat{e}_{T+1}| = T_1^{(T+1)}
\leq \frac{4B}{\lambda}\|\phi(\varepsilon_{T+1}) - \phi(\hat{\varepsilon}_{T+1})\|
\leq \frac{4B\delta_T}{\lambda}. \qedhere
\end{equation*}
\end{proof}

\subsubsection{Coverage under Mixing}

\begin{theorem}[Marginal Coverage Guarantee under $\alpha$-Mixing]
\label{thm:coverage_mixing}
Under Assumptions~\ref{assumption:stationary_mixing}, \ref{assumption:known_mean_cov}, \ref{assumption_estimation_quality}, and \ref{bounded_kernel}, for any $T \geq 2$, $\alpha \in (0,1)$, and $\lambda > 0$, with probability at least $1 - (M(\log T)^2/(2T))^{1/3}$:
\begin{equation*}
\bigl|\mathbb{P}(Y_{T+1} \in \widehat{C}_{T+1}^\alpha) - (1-\alpha)\bigr|
\leq 16 \cdot \frac{(M/2)^{1/3}(\log T)^{2/3}}{T^{1/3}} + 4(L+1)(\tilde{C}_S + \tilde{\zeta}_T),
\end{equation*}
where $\tilde{C}_S = \sqrt{4B\delta_T/\lambda}$ and $\tilde{\zeta}_T = 4B\delta_T/\lambda = \tilde{C}_S^2$.
\end{theorem}

Under $\alpha$-mixing dependence, $X_{T+1}$ typically contains lags of $Y_t$ and is therefore correlated with $\varepsilon_{T+1}$. The exogeneity condition $\varepsilon_{T+1} \perp X_{T+1}$ required for conditional coverage (Assumption~\ref{assumption_iid}) fails in this setting, and the bound is strictly marginal. Exact conditional coverage is known to be unachievable distribution-free~\cite{vovk2012conditional, foygel2021limits}. Approximate conditional coverage is assessed empirically via worst-slab coverage in Section~\ref{sec:exp:2d}.

\begin{proof}
Under Assumption~\ref{assumption:known_mean_cov}, the same argument as Lemma~\ref{lemma:test_point_error_mixing} applies to each calibration point: for $t = 1, \ldots, T$,
\begin{equation*}
|e_t - \hat{e}_t| \leq \frac{4B\delta_T}{\lambda} = \tilde{C}_S^2,
\end{equation*}
so $(1/T)\sum_{t=1}^T |e_t - \hat{e}_t| \leq \tilde{C}_S^2$.

By the same argument as Lemma~\ref{lemma:distance_ecdfs} (which does not rely on independence of the sequence):
\begin{equation*}
\sup_x |\widehat{F}_{T+1}(x) - \widetilde{F}_{T+1}(x)|
\leq (L+1)\tilde{C}_S + 2\sup_x |\widetilde{F}_{T+1}(x) - F_e(x)|.
\end{equation*}

Define $A_T = \{\sup_x |\widetilde{F}_{T+1}(x) - F_e(x)| \leq (M/2)^{1/3}(\log T)^{2/3}/T^{1/3}\}$. By Lemma~\ref{lemma:ecdf_mixing}:
\begin{equation*}
\mathbb{P}(A_T) \geq 1 - \biggl(\frac{M(\log T)^2}{2T}\biggr)^{1/3}.
\end{equation*}

Following the framework of Theorem~\ref{thm:kernel_coverage}, conditioning on $A_T$:
\begin{align*}
|\widehat{F}_{T+1}(\hat{e}_{T+1}) - F_e(e_{T+1})|
&\leq \sup_x |\widehat{F}_{T+1}(x) - \widetilde{F}_{T+1}(x)|
+ \sup_x |\widetilde{F}_{T+1}(x) - F_e(x)|
+ L\tilde{\zeta}_T \\
&\leq (L+1)\tilde{C}_S + 3 \cdot \frac{(M/2)^{1/3}(\log T)^{2/3}}{T^{1/3}}
+ L\tilde{\zeta}_T \\
&\leq (L+1)(\tilde{C}_S + \tilde{\zeta}_T) + 3 \cdot \frac{(M/2)^{1/3}(\log T)^{2/3}}{T^{1/3}}.
\end{align*}

Since $F_e(e_{T+1}) \sim \mathrm{Uniform}(0,1)$ under stationarity, $\mathbb{P}(|F_e(e_{T+1}) - \gamma| \leq \eta \mid A_T) \leq 2\eta$. Note also that
\begin{equation*}
\mathbb{P}(A_T^c) \leq \biggl(\frac{M(\log T)^2}{2T}\biggr)^{1/3}
< 2 \cdot \frac{(M/2)^{1/3}(\log T)^{2/3}}{T^{1/3}}
\qquad \text{for } T \geq 2.
\end{equation*}

Therefore, for any $\gamma \in [0,1]$:
\begin{align*}
&\mathbb{P}\bigl(|F_e(e_{T+1}) - \gamma| \leq |\widehat{F}_{T+1}(\hat{e}_{T+1}) - F_e(e_{T+1})|\bigr) \\
&\leq 2(L+1)(\tilde{C}_S + \tilde{\zeta}_T) + 8 \cdot \frac{(M/2)^{1/3}(\log T)^{2/3}}{T^{1/3}}.
\end{align*}

Applying with $\gamma = \beta$ and $\gamma = 1-\alpha+\beta$ and summing gives the stated bound.
\end{proof}

\subsubsection{Convergence Rate under Mixing}

\begin{theorem}[Marginal Coverage Convergence Rate under $\alpha$-Mixing]
\label{thm:convergence_mixing}
Under Assumptions~\ref{assumption:stationary_mixing}, \ref{assumption:known_mean_cov}, \ref{assumption_estimation_quality}, and \ref{bounded_kernel}, there exists $C > 0$ depending only on $B$, $M$, $L$ such that with probability at least $1 - (M(\log T)^2/(2T))^{1/3}$:
\begin{equation*}
\bigl|\mathbb{P}(Y_{T+1} \in \widehat{C}_{T+1}^\alpha) - (1-\alpha)\bigr|
\leq
\begin{cases}
C \cdot T^{-1/3}(\log T)^{2/3}
& \text{if } \rho > 2/3, \; \lambda = T^{-\beta}, \; \beta \in (0, \rho - 2/3) \\[4pt]
C \cdot T^{(\beta-\rho)/2}
& \text{if } \rho \leq 2/3, \; \lambda = T^{-\beta}, \; \beta \in (0, \rho)
\end{cases}
\end{equation*}
\end{theorem}

\begin{proof}
With $\delta_T = O(T^{-\rho})$ and $\lambda = T^{-\beta}$:
\begin{equation}
\tilde{C}_S = \sqrt{\frac{4B\delta_T}{\lambda}} = O(T^{(\beta-\rho)/2}),
\qquad
\tilde{\zeta}_T = \frac{4B\delta_T}{\lambda} = O(T^{\beta-\rho})
\label{eq:mixing_cs_rate}
\end{equation}

Since $\beta < \rho$, both vanish as $T \to \infty$. For large $T$ where $\tilde{C}_S < 1$:
\begin{equation*}
\tilde{C}_S + \tilde{\zeta}_T = \tilde{C}_S(1 + \tilde{C}_S) \leq 2\tilde{C}_S
\end{equation*}
so $4(L+1)(\tilde{C}_S + \tilde{\zeta}_T) = O(T^{(\beta-\rho)/2})$.

From Theorem~\ref{thm:coverage_mixing}:
\begin{equation*}
\bigl|\mathbb{P}(Y_{T+1} \in \widehat{C}_{T+1}^\alpha) - (1-\alpha)\bigr|
\leq \underbrace{O(T^{-1/3}(\log T)^{2/3})}_{\text{mixing ECDF term}}
+ \underbrace{O(T^{(\beta-\rho)/2})}_{\text{score approximation term}}.
\end{equation*}

The score term decays faster than the mixing term when:
\begin{equation*}
\frac{\beta - \rho}{2} < -\frac{1}{3}
\quad \iff \quad
\beta < \rho - \frac{2}{3}.
\end{equation*}

\emph{Case $\rho > 2/3$.} We can choose $\beta \in (0, \rho - 2/3)$. Then:
\begin{equation*}
\frac{\beta - \rho}{2} < \frac{(\rho - 2/3) - \rho}{2} = -\frac{1}{3},
\end{equation*}
so the mixing ECDF term dominates, giving rate $O(T^{-1/3}(\log T)^{2/3})$.

\emph{Case $\rho \leq 2/3$.} We choose $\beta \in (0, \rho)$. Since $\rho \leq 2/3$, for any $\beta > 0$:
\begin{equation*}
\frac{\beta - \rho}{2} \geq \frac{-\rho}{2} \geq -\frac{1}{3},
\end{equation*}
so the score term dominates, giving rate $O(T^{(\beta-\rho)/2})$.
\end{proof}

\section{Evaluation Details and Additional Experimental Results}
\label{sec:additional_experiments}

\subsection{Volume Estimation}
\label{sec:volume_estimation}

For the Mahalanobis score, the prediction region is an ellipsoid and its volume admits a closed-form expression
\begin{equation*}
\mathrm{V} = \frac{\pi^{d/2}}{\Gamma(d/2 + 1)}
\cdot \hat{q}^{d/2} \cdot \det(\widehat{\Sigma})^{1/2}
\end{equation*}
where $\hat{q}$ is the conformal threshold and $\widehat{\Sigma}$ is the sample covariance. For all other methods, the prediction region has no closed form and we estimate its volume by Monte Carlo. We sample uniformly from a bounding box constructed by padding the range of the calibration residuals by 1\% in each coordinate, score each sample, and compute the fraction falling inside the prediction region. We use $50{,}000$ MC samples. We use the same random samples (controlled by a fixed seed) for all scorers within each experimental seed. In the real data experiments, we instead sample from an enlarged Mahalanobis ellipsoid.

\subsection{Worst-Slab Coverage}
\label{sec:wsc}

Worst-slab coverage (WSC)~\cite{cauchois2021knowing} approximates conditional coverage by searching for subgroups of the test data with low empirical coverage. Let $c_i = \mathbf{1}\{Y_i \in C(X_i)\}$ denote the coverage indicator for test point $i$. We consider slabs of the form $\{x : a \le v^\top x \le b\}$ where $v$ is a unit direction. For each direction $v$, we project the test features onto $v$ and compute the empirical quantiles of the projections $\{v^\top X_i\}$. We construct slabs that contain approximately $20\%$ of the data. Worst-slab coverage is then defined as
\begin{equation*}
\mathrm{WSC} =
\min_{v,q}
\frac{
\sum_i c_i \,
\mathbf{1}\{Q_v(q) \le v^\top X_i \le Q_v(q+0.2)\}
}{
\sum_i
\mathbf{1}\{Q_v(q) \le v^\top X_i \le Q_v(q+0.2)\}
}.
\end{equation*}

In practice, we approximate this search by sampling 200 random unit directions and evaluating 20 quantile positions per direction. Each slab is required to contain at least 30 test points. The same set of random directions is used for all methods and random seeds.

\subsection{Additional Details and Results on Synthetic Data}
\subsubsection{Data Generating Process}
\label{appendix:synthetic_dgp}
We consider a bivariate regression task with a single input $X \sim \mathrm{Uniform}(0,1)$ and response $Y = f(X) + \varepsilon \in \mathbb{R}^2$, where the true regression function is
\begin{align*}
f_1(x) &= 3x^2 - 1.5x + \sin(4\pi x), \\
f_2(x) &= 2x^3 - x + 0.5\cos(3\pi x).
\end{align*} 
The noise is constructed as $\varepsilon = (z_1,\; z_2 + 0.5\,z_1^2 - 0.5\sigma_1^2)$ where $z_1 \sim \mathcal{N}(0, 0.3^2)$ and $z_2 \sim \mathcal{N}(0, 0.1^2)$ are independent, and $\sigma_1=0.3$. The noise is also centered by subtracting its sample mean. We generate $n = 10{,}000$ samples and use a 50/25/25 train/calibration/test split.

\subsubsection{Models}

\label{appendix:synthetic_models}

We use two fitted models to compare the conformal prediction results: a simple ordinary least squares (\texttt{LinearRegression}) and a more complex fitting via a two-layer MLP with hidden sizes $(64, 64)$ trained for up to 1000 iterations.


\subsection{Additional Details and Results on Real Data}
\label{appendix:real_data}
\subsubsection{Datasets}
\label{appendix:datasets}

We use three datasets previously used in Feldman et al.~\cite{feldman2022} and Romano et al.~\cite{romano2019conformalized}: House (King County house sales, Kaggle)~\cite{house_data}, Bio (physicochemical properties of protein tertiary structure, UCI \#265)~\cite{bio_data}, and Blog (BlogFeedback, UCI \#304)~\cite{blog_data}. Following Feldman et al.~\cite{feldman2022}, we augment the original univariate response with one additional variable to obtain a bivariate target ($d=2$). We further extend each dataset to $d=3$ and $d=4$ by including additional response variables from the same source, listed in Table~\ref{tab:datasets}. On House we also consider $d = 6, 8$ and $10$, described below.

\begin{table}[H]
\centering
\begin{tabular}{l c l l l l}
\toprule
Dataset & $n$ & Response & Additional response 1 & Add. response 2 & Add. response 3 \\
\midrule
House & 21{,}613 & price & lat & sqft\_living15 & sqft\_lot15 \\
Bio & 45{,}730 & RMSD & F7 & F8 & F9 \\
Blog & 52{,}397 & col.~281 & col.~61 & col.~62 & col.~280\\
\bottomrule
\end{tabular}
\vspace{0.3cm}
\caption{Dataset characteristics and target variables.}
\vspace{-0.5cm}
\label{tab:datasets}
\end{table}

{
The House settings with $d = 6, 8$ and $10$ use nested sets of continuous response variables, listed in Table~\ref{tab:house_highdim}. The variable sqft\_above is removed, because sqft\_living $=$ sqft\_above $+$ sqft\_basement holds exactly and would make the residual covariance singular.
}
\begin{table}[H]
\centering
\begin{tabular}{c l}
\toprule
$d$ & Response variables \\
\midrule
6 & price, lat, long, sqft\_living, sqft\_living15, sqft\_lot \\
8 & those at $d = 6$, sqft\_lot15, grade \\
10 & those at $d = 8$, bathrooms, bedrooms \\
\bottomrule
\end{tabular}
\vspace{0.3cm}
\caption{Response variables for House at $d = 6, 8, 10$.}
\label{tab:house_highdim}
\end{table}

All datasets are split $50/25/25$ into training, calibration, and test sets, and the calibration set is further divided into two equal folds: a fit fold, on which the score is constructed, and a quantile fold, on which the threshold is computed. Results are averaged over 50 random seeds controlling the split permutation. {Full results including the comparison among all methods, coverages, volume and WSC are provided in Section~\ref{appendix:full_results}.}


\subsubsection{Models}

We also use two models for our real data experiment, with a simpler and a more complex model choice. Ridge regression (regularization parameter chosen to be $1.0$) and a two-layer MLP with hidden sizes $(128, 64)$ trained for up to 500 iterations. For multivariate targets, Ridge is wrapped in \texttt{MultiOutputRegressor}, while the MLP handles multi-output natively. All models are implemented with SciKit-Learn. 
{The two models represent different levels of misspecification: Ridge regression cannot capture the nonlinear structure in the data and produces residuals with pronounced non-elliptical geometry. This allows us to evaluate the \scoreaccsp under different fitting regimes.}

\subsubsection{Full Results}
\label{appendix:full_results}

In {Tables~\ref{tab:bio_2d}--\ref{tab:blog_4d}, we report coverage, volume, and 
worst-slab-coverage (WSC) for all datasets, dimensions, coverage levels, and models as described in Appendix~\ref{appendix:datasets}. Under Ridge, the \scoreaccsp achieves the smallest volumes across all configurations. Under the MLP it does so in all but seven configurations, all at $\alpha \le 0.02$ and six of them on Blog, where the density-only score is smaller by up to $12$\%. The Bonferroni method produces the largest regions in all cases. The density-only score and Mahalanobis ellipsoid are closer in performance, with the relative ordering depending on the dataset and dimension. Coverage is maintained at the nominal level across all methods, consistent with the distribution-free guarantee.} 
{WSC values are comparable across methods at the tighter coverage levels. At $\alpha = 0.1$ and $0.05$ the \scoreaccsp has the lowest WSC in most configurations, consistent with the non-linear, anisotropic nature of the score as mentioned in Section~\ref{sec:exp:2d}.}

\subsubsection*{Bio $d=2$}
\begin{table}[H]
  \centering
  \resizebox{\textwidth}{!}{%
\begin{tabular}{ll ccc ccc}
    \toprule
    & & \multicolumn{3}{c}{Ridge} & \multicolumn{3}{c}{MLP} \\
    \cmidrule(lr){3-5} \cmidrule(lr){6-8}
    $\alpha$ & Method & Coverage & Volume & WSC & Coverage & Volume & WSC \\
    \midrule
    \multirow{5}{*}{0.1} & Bonferroni & 0.90024\tiny{$\pm$0.00442} & 6.311\tiny{$\pm$2.211} & 0.80853\tiny{$\pm$0.01461} & 0.89990\tiny{$\pm$0.00378} & 5.612\tiny{$\pm$2.048} & 0.84163\tiny{$\pm$0.01323} \\
    & Mahalanobis & 0.89963\tiny{$\pm$0.00468} & 5.258\tiny{$\pm$1.704} & 0.79697\tiny{$\pm$0.01796} & 0.89997\tiny{$\pm$0.00390} & 4.621\tiny{$\pm$1.594} & 0.83398\tiny{$\pm$0.01552} \\
    & Density (med$/2$) & 0.89975\tiny{$\pm$0.00500} & 3.461\tiny{$\pm$0.191} & 0.77721\tiny{$\pm$0.01251} & 0.89962\tiny{$\pm$0.00446} & 2.773\tiny{$\pm$0.114} & 0.81109\tiny{$\pm$0.00927} \\
    & OT-CP & 0.90047\tiny{$\pm$0.00542} & 3.277\tiny{$\pm$0.218} & 0.77165\tiny{$\pm$0.01183} & 0.89943\tiny{$\pm$0.00468} & 2.612\tiny{$\pm$0.134} & 0.80339\tiny{$\pm$0.01055} \\
    & \textbf{MKS (med$/2$)} & 0.89959\tiny{$\pm$0.00639} & \textbf{2.706\tiny{$\pm$0.217}} & 0.72916\tiny{$\pm$0.01516} & 0.90004\tiny{$\pm$0.00481} & \textbf{2.094\tiny{$\pm$0.138}} & 0.77356\tiny{$\pm$0.01256} \\
    \midrule
    \multirow{5}{*}{0.05} & Bonferroni & 0.95035\tiny{$\pm$0.00363} & 8.954\tiny{$\pm$3.127} & 0.88657\tiny{$\pm$0.01092} & 0.95047\tiny{$\pm$0.00282} & 8.639\tiny{$\pm$3.111} & 0.90546\tiny{$\pm$0.01137} \\
\bottomrule
  \end{tabular}}
  \vspace{0.5em}
  \caption{Bio dataset ($d=2$), Ridge and MLP. Mean $\pm$ std over 50 seeds.}
  \label{tab:bio_2d}
\end{table}

\subsubsection*{Bio $d=3$}

\begin{table}[H]
  \centering
  \resizebox{\textwidth}{!}{%
  \begin{tabular}{ll ccc ccc}
    \toprule
    & & \multicolumn{3}{c}{Ridge} & \multicolumn{3}{c}{MLP} \\
    \cmidrule(lr){3-5} \cmidrule(lr){6-8}
    $\alpha$ & Method & Coverage & Volume & WSC & Coverage & Volume & WSC \\
    \midrule
    \multirow{5}{*}{0.1} & Bonferroni & 0.90062\tiny{$\pm$0.00514} & 28.30\tiny{$\pm$9.91} & 0.67673\tiny{$\pm$0.01274} & 0.90014\tiny{$\pm$0.00559} & 20.45\tiny{$\pm$7.87} & 0.76367\tiny{$\pm$0.01208} \\
    & Mahalanobis & 0.90055\tiny{$\pm$0.00541} & 20.55\tiny{$\pm$6.65} & 0.68561\tiny{$\pm$0.01514} & 0.90052\tiny{$\pm$0.00541} & 14.86\tiny{$\pm$5.34} & 0.76042\tiny{$\pm$0.01385} \\
    & Density (med$/2$) & 0.90039\tiny{$\pm$0.00550} & 15.55\tiny{$\pm$0.70} & 0.68509\tiny{$\pm$0.01414} & 0.90055\tiny{$\pm$0.00533} & 10.50\tiny{$\pm$0.62} & 0.75755\tiny{$\pm$0.01071} \\
    & OT-CP & 0.90061\tiny{$\pm$0.00539} & 13.48\tiny{$\pm$0.75} & 0.68072\tiny{$\pm$0.01380} & 0.90007\tiny{$\pm$0.00568} & 9.187\tiny{$\pm$0.649} & 0.74698\tiny{$\pm$0.01133} \\
    & \textbf{MKS (med$/2$)} & 0.89971\tiny{$\pm$0.00564} & \textbf{9.164\tiny{$\pm$0.689}} & 0.66672\tiny{$\pm$0.01675} & 0.90030\tiny{$\pm$0.00523} & \textbf{6.266\tiny{$\pm$0.542}} & 0.72609\tiny{$\pm$0.01280} \\
    \midrule
    \multirow{5}{*}{0.05} & Bonferroni & 0.95011\tiny{$\pm$0.00364} & 47.19\tiny{$\pm$16.40} & 0.82223\tiny{$\pm$0.01145} & 0.95053\tiny{$\pm$0.00311} & 36.90\tiny{$\pm$13.44} & 0.85860\tiny{$\pm$0.00850} \\
\bottomrule
  \end{tabular}}
  \vspace{0.5em}
  \caption{Bio dataset ($d=3$), Ridge and MLP. Mean $\pm$ std over 50 seeds.}
  \label{tab:bio_3d}
\end{table}

\subsubsection*{Bio $d=4$}
\begin{table}[H]
  \centering
  \resizebox{\textwidth}{!}{%
  \begin{tabular}{ll ccc ccc}
    \toprule
    & & \multicolumn{3}{c}{Ridge} & \multicolumn{3}{c}{MLP} \\
    \cmidrule(lr){3-5} \cmidrule(lr){6-8}
    $\alpha$ & Method & Coverage & Volume & WSC & Coverage & Volume & WSC \\
    \midrule
    \multirow{5}{*}{0.1} & Bonferroni & 0.90077\tiny{$\pm$0.00526} & 65.18\tiny{$\pm$22.63} & 0.69891\tiny{$\pm$0.01348} & 0.90028\tiny{$\pm$0.00559} & 46.96\tiny{$\pm$16.59} & 0.76188\tiny{$\pm$0.01405} \\
    & Mahalanobis & 0.89917\tiny{$\pm$0.00567} & 38.20\tiny{$\pm$11.95} & 0.70354\tiny{$\pm$0.01472} & 0.89971\tiny{$\pm$0.00571} & 28.22\tiny{$\pm$9.44} & 0.75492\tiny{$\pm$0.01611} \\
    & Density (med$/2$) & 0.89970\tiny{$\pm$0.00534} & 34.91\tiny{$\pm$1.55} & 0.69797\tiny{$\pm$0.01306} & 0.89980\tiny{$\pm$0.00582} & 24.23\tiny{$\pm$1.51} & 0.75685\tiny{$\pm$0.01399} \\
    & OT-CP & 0.90060\tiny{$\pm$0.00521} & 28.66\tiny{$\pm$2.01} & 0.68825\tiny{$\pm$0.01320} & 0.90014\tiny{$\pm$0.00539} & 20.85\tiny{$\pm$1.55} & 0.74028\tiny{$\pm$0.01362} \\
    & \textbf{MKS (med$/2$)} & 0.89978\tiny{$\pm$0.00618} & \textbf{14.47\tiny{$\pm$1.21}} & 0.70254\tiny{$\pm$0.01783} & 0.90004\tiny{$\pm$0.00520} & \textbf{11.31\tiny{$\pm$0.96}} & 0.72206\tiny{$\pm$0.01351} \\
    \midrule
    \multirow{5}{*}{0.05} & Bonferroni & 0.95007\tiny{$\pm$0.00388} & 128.5\tiny{$\pm$41.8} & 0.84640\tiny{$\pm$0.01118} & 0.95002\tiny{$\pm$0.00338} & 106.7\tiny{$\pm$35.2} & 0.87083\tiny{$\pm$0.00925} \\
\bottomrule
  \end{tabular}}
  \vspace{0.5em}
  \caption{Bio dataset ($d=4$), Ridge and MLP. Mean $\pm$ std over 50 seeds.}
  \label{tab:bio_4d}
\end{table}

\subsubsection*{House $d=2$}
\begin{table}[H]
  \centering
  \resizebox{\textwidth}{!}{%
  \begin{tabular}{ll ccc ccc}
    \toprule
    & & \multicolumn{3}{c}{Ridge} & \multicolumn{3}{c}{MLP} \\
    \cmidrule(lr){3-5} \cmidrule(lr){6-8}
    $\alpha$ & Method & Coverage & Volume & WSC & Coverage & Volume & WSC \\
    \midrule
    \multirow{5}{*}{0.1} & Bonferroni & 0.90006\tiny{$\pm$0.00668} & 6.937\tiny{$\pm$0.293} & 0.77850\tiny{$\pm$0.01488} & 0.89912\tiny{$\pm$0.00650} & 4.788\tiny{$\pm$0.217} & 0.77956\tiny{$\pm$0.01492} \\
    & Mahalanobis & 0.89887\tiny{$\pm$0.00633} & 6.084\tiny{$\pm$0.248} & 0.75310\tiny{$\pm$0.01587} & 0.89935\tiny{$\pm$0.00720} & 4.198\tiny{$\pm$0.189} & 0.76938\tiny{$\pm$0.01503} \\
    & Density (med$/2$) & 0.89988\tiny{$\pm$0.00704} & 5.856\tiny{$\pm$0.164} & 0.75724\tiny{$\pm$0.01467} & 0.89885\tiny{$\pm$0.00697} & 4.291\tiny{$\pm$0.199} & 0.77813\tiny{$\pm$0.01507} \\
    & OT-CP & 0.90015\tiny{$\pm$0.00651} & 5.831\tiny{$\pm$0.165} & 0.76022\tiny{$\pm$0.01329} & 0.89897\tiny{$\pm$0.00702} & 4.219\tiny{$\pm$0.202} & 0.77495\tiny{$\pm$0.01534} \\
    & \textbf{MKS (med$/2$)} & 0.90038\tiny{$\pm$0.00632} & \textbf{5.314\tiny{$\pm$0.175}} & 0.73602\tiny{$\pm$0.01514} & 0.89945\tiny{$\pm$0.00764} & \textbf{3.897\tiny{$\pm$0.188}} & 0.75428\tiny{$\pm$0.01582} \\
    \midrule
    \multirow{5}{*}{0.05} & Bonferroni & 0.95039\tiny{$\pm$0.00420} & 9.621\tiny{$\pm$0.406} & 0.85574\tiny{$\pm$0.01102} & 0.94965\tiny{$\pm$0.00488} & 7.606\tiny{$\pm$0.403} & 0.86303\tiny{$\pm$0.01333} \\
\bottomrule
  \end{tabular}}
  \vspace{0.5em}
  \caption{House dataset ($d=2$), Ridge and MLP. Mean $\pm$ std over 50 seeds.}
  \label{tab:house_2d}
\end{table}

\subsubsection*{House $d=3$}

\begin{table}[H]
  \centering
  \resizebox{\textwidth}{!}{%
  \begin{tabular}{ll ccc ccc}
    \toprule
    & & \multicolumn{3}{c}{Ridge} & \multicolumn{3}{c}{MLP} \\
    \cmidrule(lr){3-5} \cmidrule(lr){6-8}
    $\alpha$ & Method & Coverage & Volume & WSC & Coverage & Volume & WSC \\
    \midrule
    \multirow{5}{*}{0.1} & Bonferroni & 0.90065\tiny{$\pm$0.00657} & 20.85\tiny{$\pm$1.16} & 0.73362\tiny{$\pm$0.01642} & 0.90169\tiny{$\pm$0.00744} & 14.93\tiny{$\pm$0.89} & 0.75628\tiny{$\pm$0.01503} \\
    & Mahalanobis & 0.90021\tiny{$\pm$0.00731} & 16.75\tiny{$\pm$0.83} & 0.72011\tiny{$\pm$0.01762} & 0.90104\tiny{$\pm$0.00726} & 11.21\tiny{$\pm$0.69} & 0.73932\tiny{$\pm$0.01643} \\
    & Density (med$/2$) & 0.89993\tiny{$\pm$0.00691} & 16.79\tiny{$\pm$0.68} & 0.73210\tiny{$\pm$0.01571} & 0.90020\tiny{$\pm$0.00691} & 11.97\tiny{$\pm$0.68} & 0.75362\tiny{$\pm$0.01476} \\
    & OT-CP & 0.90088\tiny{$\pm$0.00682} & 17.08\tiny{$\pm$0.80} & 0.73737\tiny{$\pm$0.01567} & 0.90081\tiny{$\pm$0.00701} & 11.67\tiny{$\pm$0.69} & 0.74997\tiny{$\pm$0.01453} \\
    & \textbf{MKS (med$/2$)} & 0.90041\tiny{$\pm$0.00700} & \textbf{14.79\tiny{$\pm$0.70}} & 0.71360\tiny{$\pm$0.01933} & 0.90137\tiny{$\pm$0.00722} & \textbf{10.32\tiny{$\pm$0.68}} & 0.72964\tiny{$\pm$0.01702} \\
    \midrule
    \multirow{5}{*}{0.05} & Bonferroni & 0.95053\tiny{$\pm$0.00502} & 38.09\tiny{$\pm$2.92} & 0.83451\tiny{$\pm$0.01292} & 0.95058\tiny{$\pm$0.00527} & 28.93\tiny{$\pm$2.40} & 0.85234\tiny{$\pm$0.01524} \\
\bottomrule
  \end{tabular}}
  \vspace{0.5em}
  \caption{House dataset ($d=3$), Ridge and MLP. Mean $\pm$ std over 50 seeds.}
  \label{tab:house_3d}
\end{table}

\subsubsection*{House $d=4$}

\begin{table}[H]
  \centering
  \resizebox{\textwidth}{!}{%
  \begin{tabular}{ll ccc ccc}
    \toprule
    & & \multicolumn{3}{c}{Ridge} & \multicolumn{3}{c}{MLP} \\
    \cmidrule(lr){3-5} \cmidrule(lr){6-8}
    $\alpha$ & Method & Coverage & Volume & WSC & Coverage & Volume & WSC \\
    \midrule
    \multirow{5}{*}{0.1} & Bonferroni & 0.90121\tiny{$\pm$0.00770} & 71.19\tiny{$\pm$12.50} & 0.69704\tiny{$\pm$0.02018} & 0.90116\tiny{$\pm$0.00761} & 53.00\tiny{$\pm$7.85} & 0.71708\tiny{$\pm$0.01802} \\
    & Mahalanobis & 0.90012\tiny{$\pm$0.00785} & 40.15\tiny{$\pm$6.51} & 0.67636\tiny{$\pm$0.02025} & 0.90088\tiny{$\pm$0.00793} & 28.44\tiny{$\pm$4.16} & 0.69745\tiny{$\pm$0.01927} \\
    & Density (med$/2$) & 0.90085\tiny{$\pm$0.00803} & 33.34\tiny{$\pm$2.90} & 0.68651\tiny{$\pm$0.01932} & 0.90112\tiny{$\pm$0.00860} & 24.81\tiny{$\pm$2.49} & 0.70875\tiny{$\pm$0.01903} \\
    & OT-CP & 0.90028\tiny{$\pm$0.00742} & 35.53\tiny{$\pm$3.19} & 0.69105\tiny{$\pm$0.01780} & 0.90085\tiny{$\pm$0.00817} & 25.66\tiny{$\pm$2.79} & 0.70857\tiny{$\pm$0.01814} \\
    & \textbf{MKS (med$/2$)} & 0.90203\tiny{$\pm$0.00793} & \textbf{18.94\tiny{$\pm$2.18}} & 0.66531\tiny{$\pm$0.02129} & 0.90166\tiny{$\pm$0.00689} & \textbf{15.32\tiny{$\pm$1.72}} & 0.67872\tiny{$\pm$0.02019} \\
    \midrule
    \multirow{5}{*}{0.05} & Bonferroni & 0.95047\tiny{$\pm$0.00579} & 201.5\tiny{$\pm$45.5} & 0.80905\tiny{$\pm$0.02005} & 0.95021\tiny{$\pm$0.00520} & 144.3\tiny{$\pm$25.0} & 0.82313\tiny{$\pm$0.01833} \\
\bottomrule
  \end{tabular}}
  \vspace{0.5em}
  \caption{House dataset ($d=4$), Ridge and MLP. Mean $\pm$ std over 50 seeds.}
  \label{tab:house_4d}
\end{table}

\subsubsection*{House $d=6$}
\begin{table}[H]
  \centering
  \resizebox{\textwidth}{!}{%
  \begin{tabular}{ll ccc ccc}
    \toprule
    & & \multicolumn{3}{c}{Ridge} & \multicolumn{3}{c}{MLP} \\
    \cmidrule(lr){3-5} \cmidrule(lr){6-8}
    $\alpha$ & Method & Coverage & Volume & WSC & Coverage & Volume & WSC \\
    \midrule
    \multirow{5}{*}{0.1} & Bonferroni & 0.89973\tiny{$\pm$0.00880} & 4.416\tiny{$\pm$0.910}$\times10^{3}$ & 0.78980\tiny{$\pm$0.01780} & 0.90122\tiny{$\pm$0.00787} & 2.783\tiny{$\pm$0.641}$\times10^{3}$ & 0.79730\tiny{$\pm$0.01600} \\
    & Mahalanobis & 0.90020\tiny{$\pm$0.00830} & 990.2\tiny{$\pm$210.9} & 0.79480\tiny{$\pm$0.01620} & 0.90081\tiny{$\pm$0.00775} & 762.1\tiny{$\pm$159.5} & 0.78644\tiny{$\pm$0.01597} \\
    & Density (med$/2$) & 0.90028\tiny{$\pm$0.00864} & 1.126\tiny{$\pm$0.159}$\times10^{3}$ & 0.80352\tiny{$\pm$0.01575} & 0.90132\tiny{$\pm$0.00767} & 775.6\tiny{$\pm$104.7} & 0.78758\tiny{$\pm$0.01803} \\
    & OT-CP & 0.90070\tiny{$\pm$0.00782} & 1.306\tiny{$\pm$0.148}$\times10^{3}$ & 0.79594\tiny{$\pm$0.01783} & 0.90172\tiny{$\pm$0.00698} & 962.4\tiny{$\pm$125.7} & 0.77695\tiny{$\pm$0.01757} \\
    & \textbf{MKS (med$/2$)} & 0.90138\tiny{$\pm$0.00790} & \textbf{215.8\tiny{$\pm$29.7}} & 0.78398\tiny{$\pm$0.01725} & 0.90179\tiny{$\pm$0.00715} & \textbf{271.6\tiny{$\pm$49.2}} & 0.76608\tiny{$\pm$0.01613} \\
    \midrule
    \multirow{5}{*}{0.05} & Bonferroni & 0.95024\tiny{$\pm$0.00584} & 2.411\tiny{$\pm$0.695}$\times10^{4}$ & 0.89319\tiny{$\pm$0.01146} & 0.95013\tiny{$\pm$0.00611} & 1.593\tiny{$\pm$0.470}$\times10^{4}$ & 0.89397\tiny{$\pm$0.01148} \\
\bottomrule
  \end{tabular}}
  \vspace{0.5em}
  \caption{House dataset ($d=6$), Ridge and MLP. Mean $\pm$ std over 50 seeds.}
  \label{tab:house_6d}
\end{table}

\subsubsection*{House $d=8$}
\begin{table}[H]
  \centering
  \resizebox{\textwidth}{!}{%
  \begin{tabular}{ll ccc ccc}
    \toprule
    & & \multicolumn{3}{c}{Ridge} & \multicolumn{3}{c}{MLP} \\
    \cmidrule(lr){3-5} \cmidrule(lr){6-8}
    $\alpha$ & Method & Coverage & Volume & WSC & Coverage & Volume & WSC \\
    \midrule
    \multirow{5}{*}{0.1} & Bonferroni & 0.89947\tiny{$\pm$0.00916} & 1.033\tiny{$\pm$0.310}$\times10^{5}$ & 0.80330\tiny{$\pm$0.01771} & 0.90100\tiny{$\pm$0.00850} & 6.542\tiny{$\pm$2.157}$\times10^{4}$ & 0.80289\tiny{$\pm$0.01687} \\
    & Mahalanobis & 0.90055\tiny{$\pm$0.00815} & 6.866\tiny{$\pm$2.303}$\times10^{3}$ & 0.80401\tiny{$\pm$0.01466} & 0.90152\tiny{$\pm$0.00739} & 5.045\tiny{$\pm$1.553}$\times10^{3}$ & 0.79315\tiny{$\pm$0.01462} \\
    & Density (med$/2$) & 0.90068\tiny{$\pm$0.00868} & 1.172\tiny{$\pm$0.228}$\times10^{4}$ & 0.80822\tiny{$\pm$0.01615} & 0.90077\tiny{$\pm$0.00717} & 7.121\tiny{$\pm$1.355}$\times10^{3}$ & 0.78692\tiny{$\pm$0.01511} \\
    & OT-CP & 0.89972\tiny{$\pm$0.00832} & 2.215\tiny{$\pm$0.415}$\times10^{4}$ & 0.79501\tiny{$\pm$0.01731} & 0.90134\tiny{$\pm$0.00787} & 1.518\tiny{$\pm$0.318}$\times10^{4}$ & 0.78045\tiny{$\pm$0.01779} \\
    & \textbf{MKS (med$/2$)} & 0.90165\tiny{$\pm$0.00743} & \textbf{895.3\tiny{$\pm$165.0}} & 0.79750\tiny{$\pm$0.01543} & 0.90161\tiny{$\pm$0.00809} & \textbf{1.300\tiny{$\pm$0.322}$\times10^{3}$} & 0.77334\tiny{$\pm$0.01806} \\
    \midrule
    \multirow{5}{*}{0.05} & Bonferroni & 0.95048\tiny{$\pm$0.00625} & 1.161\tiny{$\pm$0.452}$\times10^{6}$ & 0.89366\tiny{$\pm$0.01335} & 0.95102\tiny{$\pm$0.00573} & 7.116\tiny{$\pm$2.783}$\times10^{5}$ & 0.89522\tiny{$\pm$0.01194} \\
\bottomrule
  \end{tabular}}
  \vspace{0.5em}
  \caption{House dataset ($d=8$), Ridge and MLP. Mean $\pm$ std over 50 seeds.}
  \label{tab:house_8d}
\end{table}

\subsubsection*{House $d=10$}

\begin{table}[H]
  \centering
  \resizebox{\textwidth}{!}{%
  \begin{tabular}{ll ccc ccc}
    \toprule
    & & \multicolumn{3}{c}{Ridge} & \multicolumn{3}{c}{MLP} \\
    \cmidrule(lr){3-5} \cmidrule(lr){6-8}
    $\alpha$ & Method & Coverage & Volume & WSC & Coverage & Volume & WSC \\
    \midrule
    \multirow{5}{*}{0.1} & Bonferroni & 0.90031\tiny{$\pm$0.00889} & 3.655\tiny{$\pm$1.386}$\times10^{6}$ & 0.81452\tiny{$\pm$0.01774} & 0.90040\tiny{$\pm$0.00893} & 2.020\tiny{$\pm$0.835}$\times10^{6}$ & 0.81630\tiny{$\pm$0.01537} \\
    & Mahalanobis & 0.90081\tiny{$\pm$0.00825} & 7.574\tiny{$\pm$2.510}$\times10^{4}$ & 0.79989\tiny{$\pm$0.01476} & 0.90113\tiny{$\pm$0.00791} & 4.971\tiny{$\pm$1.612}$\times10^{4}$ & 0.79086\tiny{$\pm$0.01549} \\
    & Density (med$/2$) & 0.90033\tiny{$\pm$0.00809} & 1.936\tiny{$\pm$0.391}$\times10^{5}$ & 0.80688\tiny{$\pm$0.01583} & 0.90084\tiny{$\pm$0.00781} & 1.028\tiny{$\pm$0.238}$\times10^{5}$ & 0.79270\tiny{$\pm$0.01650} \\
    & OT-CP & 0.89941\tiny{$\pm$0.00773} & 4.959\tiny{$\pm$1.200}$\times10^{5}$ & 0.80037\tiny{$\pm$0.01535} & 0.90105\tiny{$\pm$0.00786} & 3.446\tiny{$\pm$0.901}$\times10^{5}$ & 0.78951\tiny{$\pm$0.01602} \\
    & \textbf{MKS (med$/2$)} & 0.90120\tiny{$\pm$0.00741} & \textbf{1.490\tiny{$\pm$0.312}$\times10^{4}$} & 0.79483\tiny{$\pm$0.01595} & 0.90138\tiny{$\pm$0.00776} & \textbf{1.687\tiny{$\pm$0.433}$\times10^{4}$} & 0.77516\tiny{$\pm$0.01757} \\
    \midrule
    \multirow{5}{*}{0.05} & Bonferroni & 0.95013\tiny{$\pm$0.00615} & 5.299\tiny{$\pm$2.349}$\times10^{7}$ & 0.89619\tiny{$\pm$0.01215} & 0.95058\tiny{$\pm$0.00596} & 3.167\tiny{$\pm$1.450}$\times10^{7}$ & 0.89899\tiny{$\pm$0.01076} \\
\bottomrule
  \end{tabular}}
  \vspace{0.5em}
  \caption{House dataset ($d=10$), Ridge and MLP. Mean $\pm$ std over 50 seeds.}
  \label{tab:house_10d}
\end{table}

\subsubsection*{Blog $d=2$}

\begin{table}[H]
  \centering
  \resizebox{\textwidth}{!}{%
  \begin{tabular}{ll ccc ccc}
    \toprule
    & & \multicolumn{3}{c}{Ridge} & \multicolumn{3}{c}{MLP} \\
    \cmidrule(lr){3-5} \cmidrule(lr){6-8}
    $\alpha$ & Method & Coverage & Volume & WSC & Coverage & Volume & WSC \\
    \midrule
    \multirow{5}{*}{0.1} & Bonferroni & 0.89928\tiny{$\pm$0.00431} & 8.331\tiny{$\pm$0.869} & 0.79801\tiny{$\pm$0.00845} & 0.89926\tiny{$\pm$0.00413} & 3.449\tiny{$\pm$0.980} & 0.77705\tiny{$\pm$0.01435} \\
    & Mahalanobis & 0.89944\tiny{$\pm$0.00418} & 6.752\tiny{$\pm$0.684} & 0.79486\tiny{$\pm$0.00830} & 0.89943\tiny{$\pm$0.00407} & 2.894\tiny{$\pm$0.803} & 0.77036\tiny{$\pm$0.01404} \\
    & Density (med$/2$) & 0.89934\tiny{$\pm$0.00416} & 4.163\tiny{$\pm$0.110} & 0.76854\tiny{$\pm$0.00866} & 0.89914\tiny{$\pm$0.00456} & 1.776\tiny{$\pm$0.514} & 0.72959\tiny{$\pm$0.01327} \\
    & OT-CP & 0.89948\tiny{$\pm$0.00438} & 4.537\tiny{$\pm$0.159} & 0.78036\tiny{$\pm$0.00814} & 0.89913\tiny{$\pm$0.00433} & 1.852\tiny{$\pm$0.551} & 0.73659\tiny{$\pm$0.01204} \\
    & \textbf{MKS (med$/2$)} & 0.90024\tiny{$\pm$0.00457} & \textbf{3.049\tiny{$\pm$0.116}} & 0.72292\tiny{$\pm$0.01147} & 0.89905\tiny{$\pm$0.00491} & \textbf{1.722\tiny{$\pm$0.432}} & 0.72824\tiny{$\pm$0.01344} \\
    \midrule
    \multirow{5}{*}{0.05} & Bonferroni & 0.94954\tiny{$\pm$0.00301} & 11.27\tiny{$\pm$1.08} & 0.86834\tiny{$\pm$0.00760} & 0.94967\tiny{$\pm$0.00270} & 6.104\tiny{$\pm$1.450} & 0.86324\tiny{$\pm$0.00943} \\
\bottomrule
  \end{tabular}}
  \vspace{0.5em}
  \caption{Blog dataset ($d=2$), Ridge and MLP. Mean $\pm$ std over 50 seeds.}
  \label{tab:blog_2d}
\end{table}

\subsubsection*{Blog $d=3$}

\begin{table}[H]
  \centering
  \resizebox{\textwidth}{!}{%
  \begin{tabular}{ll ccc ccc}
    \toprule
    & & \multicolumn{3}{c}{Ridge} & \multicolumn{3}{c}{MLP} \\
    \cmidrule(lr){3-5} \cmidrule(lr){6-8}
    $\alpha$ & Method & Coverage & Volume & WSC & Coverage & Volume & WSC \\
    \midrule
    \multirow{5}{*}{0.1} & Bonferroni & 0.89990\tiny{$\pm$0.00409} & 19.66\tiny{$\pm$1.93} & 0.76751\tiny{$\pm$0.00897} & 0.90011\tiny{$\pm$0.00354} & 8.126\tiny{$\pm$2.579} & 0.76696\tiny{$\pm$0.00859} \\
    & Mahalanobis & 0.90011\tiny{$\pm$0.00479} & 13.75\tiny{$\pm$1.27} & 0.75441\tiny{$\pm$0.00935} & 0.90021\tiny{$\pm$0.00399} & 6.115\tiny{$\pm$2.416} & 0.76096\tiny{$\pm$0.00828} \\
    & Density (med$/2$) & 0.90003\tiny{$\pm$0.00449} & 9.797\tiny{$\pm$0.413} & 0.74079\tiny{$\pm$0.00958} & 0.89965\tiny{$\pm$0.00395} & 3.912\tiny{$\pm$1.369} & 0.74125\tiny{$\pm$0.00940} \\
    & OT-CP & 0.90014\tiny{$\pm$0.00465} & 11.01\tiny{$\pm$0.52} & 0.75375\tiny{$\pm$0.00804} & 0.89976\tiny{$\pm$0.00407} & 4.148\tiny{$\pm$1.538} & 0.74406\tiny{$\pm$0.00921} \\
    & \textbf{MKS (med$/2$)} & 0.89961\tiny{$\pm$0.00454} & \textbf{7.226\tiny{$\pm$0.400}} & 0.71731\tiny{$\pm$0.01019} & 0.89981\tiny{$\pm$0.00454} & \textbf{3.588\tiny{$\pm$1.229}} & 0.73545\tiny{$\pm$0.01101} \\
    \midrule
    \multirow{5}{*}{0.05} & Bonferroni & 0.95017\tiny{$\pm$0.00333} & 39.27\tiny{$\pm$3.55} & 0.84628\tiny{$\pm$0.00935} & 0.95035\tiny{$\pm$0.00302} & 20.32\tiny{$\pm$5.47} & 0.86750\tiny{$\pm$0.00822} \\
\bottomrule
  \end{tabular}}
  \vspace{0.5em}
  \caption{Blog dataset ($d=3$), Ridge and MLP. Mean $\pm$ std over 50 seeds.}
  \label{tab:blog_3d}
\end{table}

\subsubsection*{Blog $d=4$}

\begin{table}[H]
  \centering
  \resizebox{\textwidth}{!}{%
  \begin{tabular}{ll ccc ccc}
    \toprule
    & & \multicolumn{3}{c}{Ridge} & \multicolumn{3}{c}{MLP} \\
    \cmidrule(lr){3-5} \cmidrule(lr){6-8}
    $\alpha$ & Method & Coverage & Volume & WSC & Coverage & Volume & WSC \\
    \midrule
    \multirow{5}{*}{0.1} & Bonferroni & 0.89974\tiny{$\pm$0.00401} & 34.62\tiny{$\pm$11.46} & 0.76656\tiny{$\pm$0.00844} & 0.89874\tiny{$\pm$0.00387} & 13.80\tiny{$\pm$8.96} & 0.76012\tiny{$\pm$0.01132} \\
    & Mahalanobis & 0.89960\tiny{$\pm$0.00493} & 16.16\tiny{$\pm$5.20} & 0.75140\tiny{$\pm$0.00960} & 0.89900\tiny{$\pm$0.00386} & 7.633\tiny{$\pm$5.564} & 0.75081\tiny{$\pm$0.01099} \\
    & Density (med$/2$) & 0.89959\tiny{$\pm$0.00453} & 12.27\tiny{$\pm$0.74} & 0.74199\tiny{$\pm$0.01030} & 0.89940\tiny{$\pm$0.00417} & 4.406\tiny{$\pm$3.027} & 0.73946\tiny{$\pm$0.01176} \\
    & OT-CP & 0.89979\tiny{$\pm$0.00446} & 14.70\tiny{$\pm$1.07} & 0.75743\tiny{$\pm$0.00923} & 0.89965\tiny{$\pm$0.00385} & 4.785\tiny{$\pm$3.323} & 0.74234\tiny{$\pm$0.01139} \\
    & \textbf{MKS (med$/2$)} & 0.89926\tiny{$\pm$0.00418} & \textbf{3.472\tiny{$\pm$0.394}} & 0.72208\tiny{$\pm$0.01068} & 0.89952\tiny{$\pm$0.00436} & \textbf{2.975\tiny{$\pm$2.373}} & 0.73056\tiny{$\pm$0.01239} \\
    \midrule
    \multirow{5}{*}{0.05} & Bonferroni & 0.95005\tiny{$\pm$0.00326} & 93.53\tiny{$\pm$31.98} & 0.84903\tiny{$\pm$0.00828} & 0.94945\tiny{$\pm$0.00368} & 49.45\tiny{$\pm$27.95} & 0.86673\tiny{$\pm$0.01019} \\
\bottomrule
  \end{tabular}}
  \vspace{0.5em}
  \caption{Blog dataset ($d=4$), Ridge and MLP. Mean $\pm$ std over 50 seeds.}
  \label{tab:blog_4d}
\end{table}

\subsubsection{Hyperparameter ablation}
\label{appendix:hyperparameter_grid}

{
We assess sensitivity to both hyperparameters jointly, sweeping the lengthscale $\ell$ over nine values spanning a $256\times$ range and the regularization $\gamma$ over seven percentiles of the centred Gram spectrum, on three datasets. Coverage is unaffected throughout, which is the empirical counterpart of the guarantee in Theorem~\ref{thm:coverage}, valid for any configuration. Volumes vary with both parameters but are stable across a wide interior region and the only configurations that lose to the Mahalanobis ellipsoid combine a very wide kernel with a very small $\gamma$. A wide kernel collapses the centred spectrum onto a few smooth directions, and a low-percentile $\gamma$ then lands among the near-zero eigenvalues below them, so the score approaches the unregularized limit in which the calibration scores $\gamma h_{ii}$ degenerate (Proposition~\ref{prop:leverage}).}
\begin{table}[h!]
\small
\centering
\resizebox{\textwidth}{!}{%
\begin{tabular}[t]{l c c c c c c c}
    \toprule
    & \multicolumn{7}{c}{Volume ratio to Mahalanobis} \\
    \cmidrule(lr){2-8}
    $\ell \,/\, (\mathrm{med}/2)$ & $5$ & $10$ & $30$ & $50$ & $70$ & $90$ & $95$ \\
    \midrule
    $16$ & 4.139 & 1.953 & 0.665 & 0.700 & 0.782 & 0.904 & 0.891 \\
    $8$ & 1.537 & 0.907 & 0.594 & 0.623 & 0.669 & 0.773 & 0.819 \\
    $4$ & 0.648 & 0.530 & 0.533 & 0.554 & 0.587 & 0.652 & 0.683 \\
    $2$ & 0.490 & 0.490 & 0.492 & 0.503 & 0.522 & 0.562 & 0.585 \\
    $1$ & 0.499 & 0.498 & 0.490 & 0.487 & 0.487 & 0.505 & 0.518 \\
    $1/2$ & 0.521 & 0.522 & 0.517 & 0.511 & 0.513 & 0.496 & 0.501 \\
    $1/4$ & 0.629 & 0.629 & 0.636 & 0.641 & 0.642 & 0.637 & 0.632 \\
    $1/8$ & 0.702 & 0.702 & 0.701 & 0.701 & 0.700 & 0.699 & 0.698 \\
    $1/16$ & 0.712 & 0.712 & 0.712 & 0.712 & 0.712 & 0.712 & 0.712 \\
    \bottomrule
\end{tabular}%
\hspace{1.5em}%
\begin{tabular}[t]{l c c c c c c c}
    \toprule
    & \multicolumn{7}{c}{Coverage} \\
    \cmidrule(lr){2-8}
    $\ell \,/\, (\mathrm{med}/2)$ & $5$ & $10$ & $30$ & $50$ & $70$ & $90$ & $95$ \\
    \midrule
    $16$ & 0.9016 & 0.8994 & 0.9005 & 0.9008 & 0.9005 & 0.9001 & 0.9006 \\
    $8$ & 0.9009 & 0.8990 & 0.9013 & 0.9012 & 0.9008 & 0.9005 & 0.9004 \\
    $4$ & 0.9017 & 0.9021 & 0.9020 & 0.9016 & 0.9013 & 0.9010 & 0.9010 \\
    $2$ & 0.9019 & 0.9018 & 0.9018 & 0.9022 & 0.9024 & 0.9020 & 0.9021 \\
    $1$ & 0.9019 & 0.9018 & 0.9023 & 0.9022 & 0.9026 & 0.9025 & 0.9021 \\
    $1/2$ & 0.9021 & 0.9023 & 0.9024 & 0.9023 & 0.9020 & 0.9021 & 0.9023 \\
    $1/4$ & 0.9020 & 0.9019 & 0.9021 & 0.9022 & 0.9023 & 0.9023 & 0.9024 \\
    $1/8$ & 0.9023 & 0.9023 & 0.9023 & 0.9023 & 0.9023 & 0.9022 & 0.9022 \\
    $1/16$ & 0.9024 & 0.9024 & 0.9024 & 0.9024 & 0.9024 & 0.9024 & 0.9024 \\
    \bottomrule
\end{tabular}}
\vspace{0.5em}
\caption{House $d{=}4$, $1{-}\alpha{=}0.9$, 30 seeds.
Rows: kernel lengthscale as a multiple of the median heuristic
$\mathrm{med}/2$. Columns: $\gamma$ at the given percentile of the
centred-Gram spectrum. Left: volume ratio to Mahalanobis. Right: empirical
coverage, target $0.9$, which is unaffected by either parameter.}
\label{tab:hpgrid_house_4d}
\end{table}
\begin{table}[h!]
\small
\centering
\resizebox{\textwidth}{!}{%
\begin{tabular}[t]{l c c c c c c c}
    \toprule
    & \multicolumn{7}{c}{Volume ratio to Mahalanobis} \\
    \cmidrule(lr){2-8}
    $\ell \,/\, (\mathrm{med}/2)$ & $5$ & $10$ & $30$ & $50$ & $70$ & $90$ & $95$ \\
    \midrule
    $16$ & 4.078 & 1.548 & 0.570 & 0.610 & 0.658 & 0.741 & 0.818 \\
    $8$ & 2.349 & 0.948 & 0.521 & 0.537 & 0.572 & 0.644 & 0.687 \\
    $4$ & 1.052 & 0.551 & 0.484 & 0.492 & 0.510 & 0.535 & 0.558 \\
    $2$ & 0.635 & 0.458 & 0.459 & 0.463 & 0.470 & 0.482 & 0.494 \\
    $1$ & 0.469 & 0.447 & 0.446 & 0.446 & 0.449 & 0.453 & 0.458 \\
    $1/2$ & 0.448 & 0.447 & 0.444 & 0.442 & 0.442 & 0.443 & 0.444 \\
    $1/4$ & 0.442 & 0.443 & 0.452 & 0.449 & 0.445 & 0.443 & 0.441 \\
    $1/8$ & 0.440 & 0.441 & 0.453 & 0.473 & 0.462 & 0.455 & 0.451 \\
    $1/16$ & 0.520 & 0.521 & 0.526 & 0.545 & 0.551 & 0.533 & 0.522 \\
    \bottomrule
\end{tabular}%
\hspace{1.5em}%
\begin{tabular}[t]{l c c c c c c c}
    \toprule
    & \multicolumn{7}{c}{Coverage} \\
    \cmidrule(lr){2-8}
    $\ell \,/\, (\mathrm{med}/2)$ & $5$ & $10$ & $30$ & $50$ & $70$ & $90$ & $95$ \\
    \midrule
    $16$ & 0.9008 & 0.8998 & 0.8990 & 0.8990 & 0.8990 & 0.8989 & 0.8990 \\
    $8$ & 0.8998 & 0.8987 & 0.8991 & 0.8987 & 0.8988 & 0.8988 & 0.8987 \\
    $4$ & 0.8989 & 0.8994 & 0.8994 & 0.8989 & 0.8990 & 0.8985 & 0.8986 \\
    $2$ & 0.8993 & 0.8995 & 0.8997 & 0.8996 & 0.8996 & 0.8991 & 0.8990 \\
    $1$ & 0.9000 & 0.9001 & 0.8998 & 0.8998 & 0.9000 & 0.8999 & 0.8998 \\
    $1/2$ & 0.9001 & 0.9002 & 0.9001 & 0.9001 & 0.9003 & 0.9001 & 0.9000 \\
    $1/4$ & 0.9000 & 0.8999 & 0.9001 & 0.9002 & 0.9001 & 0.9001 & 0.9000 \\
    $1/8$ & 0.8999 & 0.8997 & 0.8999 & 0.9001 & 0.8997 & 0.9000 & 0.8998 \\
    $1/16$ & 0.8993 & 0.8993 & 0.8994 & 0.8996 & 0.8996 & 0.8995 & 0.8995 \\
    \bottomrule
\end{tabular}}
\vspace{0.5em}
\caption{Blog $d{=}2$, $1{-}\alpha{=}0.9$, 30 seeds.
Rows: kernel lengthscale as a multiple of the median heuristic
$\mathrm{med}/2$. Columns: $\gamma$ at the given percentile of the
centred-Gram spectrum. Left: volume ratio to Mahalanobis. Right: empirical
coverage, target $0.9$, which is unaffected by either parameter.}
\label{tab:hpgrid_blog_2d}
\end{table}
\begin{table}[h!]
\small
\centering
\resizebox{\textwidth}{!}{%
\begin{tabular}[t]{l c c c c c c c}
    \toprule
    & \multicolumn{7}{c}{Volume ratio to Mahalanobis} \\
    \cmidrule(lr){2-8}
    $\ell \,/\, (\mathrm{med}/2)$ & $5$ & $10$ & $30$ & $50$ & $70$ & $90$ & $95$ \\
    \midrule
    $16$ & 2.805 & 2.254 & 0.559 & 0.573 & 0.601 & 0.645 & 0.745 \\
    $8$ & 1.556 & 0.963 & 0.531 & 0.543 & 0.566 & 0.626 & 0.641 \\
    $4$ & 0.937 & 0.652 & 0.503 & 0.513 & 0.529 & 0.566 & 0.590 \\
    $2$ & 0.604 & 0.498 & 0.483 & 0.489 & 0.496 & 0.516 & 0.532 \\
    $1$ & 0.463 & 0.461 & 0.462 & 0.464 & 0.474 & 0.483 & 0.490 \\
    $1/2$ & 0.450 & 0.449 & 0.446 & 0.448 & 0.447 & 0.451 & 0.458 \\
    $1/4$ & 0.450 & 0.449 & 0.450 & 0.453 & 0.459 & 0.442 & 0.440 \\
    $1/8$ & 0.518 & 0.518 & 0.521 & 0.528 & 0.527 & 0.521 & 0.518 \\
    $1/16$ & 0.559 & 0.559 & 0.559 & 0.559 & 0.559 & 0.559 & 0.559 \\
    \bottomrule
\end{tabular}%
\hspace{1.5em}%
\begin{tabular}[t]{l c c c c c c c}
    \toprule
    & \multicolumn{7}{c}{Coverage} \\
    \cmidrule(lr){2-8}
    $\ell \,/\, (\mathrm{med}/2)$ & $5$ & $10$ & $30$ & $50$ & $70$ & $90$ & $95$ \\
    \midrule
    $16$ & 0.8998 & 0.8992 & 0.8992 & 0.8995 & 0.8995 & 0.8992 & 0.8996 \\
    $8$ & 0.9006 & 0.9000 & 0.8991 & 0.8993 & 0.8993 & 0.8992 & 0.8990 \\
    $4$ & 0.8989 & 0.8992 & 0.8988 & 0.8988 & 0.8989 & 0.8990 & 0.8992 \\
    $2$ & 0.8995 & 0.8983 & 0.8989 & 0.8990 & 0.8990 & 0.8989 & 0.8986 \\
    $1$ & 0.8984 & 0.8984 & 0.8983 & 0.8983 & 0.8988 & 0.8989 & 0.8988 \\
    $1/2$ & 0.8989 & 0.8987 & 0.8987 & 0.8987 & 0.8985 & 0.8988 & 0.8988 \\
    $1/4$ & 0.8983 & 0.8984 & 0.8989 & 0.8990 & 0.8989 & 0.8988 & 0.8987 \\
    $1/8$ & 0.8997 & 0.8998 & 0.8999 & 0.8998 & 0.8998 & 0.8998 & 0.8997 \\
    $1/16$ & 0.8999 & 0.8999 & 0.9000 & 0.9000 & 0.9001 & 0.9000 & 0.9000 \\
    \bottomrule
\end{tabular}}
\vspace{0.5em}
\caption{Bio $d{=}3$, $1{-}\alpha{=}0.9$, 30 seeds.
Rows: kernel lengthscale as a multiple of the median heuristic
$\mathrm{med}/2$. Columns: $\gamma$ at the given percentile of the
centred-Gram spectrum. Left: volume ratio to Mahalanobis. Right: empirical
coverage, target $0.9$, which is unaffected by either parameter.}
\label{tab:hpgrid_bio_3d}
\end{table}



\end{document}